\documentclass[oneside,11pt]{article}

\usepackage{url,mathrsfs,algorithm}
\usepackage{amsmath,wrapfig,color}
\usepackage{amsfonts}
\usepackage{graphicx}
\usepackage{subfigure}

\begin{document}
\title{\Huge Tunable GMM Kernels}

\author{
 \bf{Ping Li} \\
         Department of Statistics and Biostatistics\\
         Department of Computer Science\\
       Rutgers University\\
          Piscataway, NJ 08854, USA\\
       \texttt{pingli@stat.rutgers.edu}\\
}

\date{}

\maketitle

\begin{abstract}
\noindent The recently proposed ``generalized min-max'' (GMM) kernel~\cite{Report:Li_GMM16} can be efficiently linearized, with direct applications in  large-scale statistical  learning and fast  near neighbor search. The linearized GMM kernel was extensively compared in~\cite{Report:Li_GMM16} with linearized radial basis function (RBF) kernel. On a large number of classification tasks,  the tuning-free GMM kernel performs (surprisingly) well compared to the best-tuned RBF kernel. Nevertheless, one would naturally expect that the GMM kernel ought to be further improved if we introduce tuning parameters.
\vspace{0.08in}

\noindent In this paper, we study three simple constructions of tunable GMM kernels: (i) the exponentiated-GMM (or eGMM) kernel,  (ii) the powered-GMM (or pGMM) kernel, and (iii) the exponentiated-powered-GMM (epGMM) kernel. The pGMM kernel can still be efficiently linearized by modifying the original hashing procedure for the GMM kernel. On about 60  publicly available classification datasets, we verify that the proposed tunable GMM kernels typically improve over the original GMM kernel. On some datasets, the improvements can be astonishingly significant.

\vspace{0.08in}

\noindent For example, on 11 popular datasets which were  used for testing deep learning algorithms and tree methods, our experiments show that the proposed tunable GMM kernels are  strong competitors to trees and deep nets. The previous studies~\cite{Article:Li_ABC_arXiv08,Proc:ABC_UAI10} developed tree methods including ``abc-robust-logitboost'' and demonstrated the excellent performance on those 11 datasets (and other datasets), by establishing the second-order tree-split formula and new derivatives for multi-class logistic loss. Compared to tree methods like ``abc-robust-logitboost'' (which are slow and need substantial model sizes), the  tunable GMM kernels produce largely comparable  results.

\vspace{0.08in}

\noindent We hope our introduction of tunable kernels would offer practitioners the flexibility of choosing  appropriate kernels and methods for large-scale search and learning for their specific applications.

\end{abstract}

\section{Introduction}

The ``generalized min-max (GMM)'' kernel~\cite{Report:Li_GMM16} was introduced for large-scale search and machine learning, owing to its efficient linearization  via either hashing or  the Nystrom method~\cite{Report:Li_GMM_Nys16}. For defining the GMM kernel, the first step is a simple transformation on the original data. Consider, for example,   the original data vector $u_i$, $i=1$ to $D$. We define the following transformation, depending on whether an entry $u_i$ is positive or negative:
\begin{align}\label{eqn_transform}
 \left\{\begin{array}{cc}
\tilde{u}_{2i-1} = u_i,\hspace{0.1in} \tilde{u}_{2i} = 0&\text{if } \ u_i >0\\
\tilde{u}_{2i-1} = 0,\hspace{0.1in} \tilde{u}_{2i} =  -u_i &\text{if } \ u_i \leq 0
\end{array}\right.
\end{align}
For example, when $D=2$ and $u = [-4\ \ 6]$, the transformed data vector becomes $\tilde{u} = [0\ \ 4\ \ 6\ \ 0]$. The GMM kernel is defined~\cite{Report:Li_GMM16} as  follows:
\begin{align}
GMM(u,v) = \frac{\sum_{i=1}^{2D}\min\{\tilde{u}_i,\tilde{v}_i\}}{\sum_{i=1}^{2D} \max\{\tilde{u}_i,\tilde{v}_i\}}
\end{align}

Even though the GMM kernel has no tuning parameter, it performs surprisingly well for classification tasks as empirically demonstrated in~\cite{Report:Li_GMM16} (also see Table~\ref{tab_data} and Table~\ref{tab_data2}), when compared to the best-tuned radial basis function (RBF) kernel:
\begin{align}
RBF(u,v;\gamma) = e^{-\gamma\left(1-\frac{\sum_{i=1}^{D}u_iv_i}{\sqrt{\sum_{i=1}^{D} u_i^2\sum_{i=1}^{D}v_i^2}}\right)}
\end{align}
where $\gamma>0$ is a crucial tuning parameter.

 Furthermore, the (nonlinear) GMM kernel can be efficiently linearized via  hashing~\cite{Report:Manasse_CWS10,Proc:Ioffe_ICDM10,Proc:Li_KDD15} (or the Nystrom method~\cite{Report:Li_GMM_Nys16}). This means we can use the linearized GMM kernel for large-scale machine learning tasks essentially at the cost of linear learning. \\

 Naturally, one would ask whether we can improve this (tuning-free) GMM kernel by introducing tuning parameters. For example, we can define the following ``exponentiated-GMM''  (eGMM) kernel:
\begin{align}
eGMM(u,v;\gamma) = e^{-\gamma\left(1-\frac{\sum_{i=1}^{2D}\min\{\tilde{u}_i,\tilde{v}_i\}}{\sum_{i=1}^{2D} \max\{\tilde{u}_i,\tilde{v}_i\}}\right)}
\end{align}
and the ``powered-GMM'' (pGMM) kernel:
\begin{align}
pGMM(u,v;\gamma) =  \frac{\sum_{i=1}^{2D}\left(\min\{\tilde{u}_i,\tilde{v}_i\}\right)^\gamma}{\sum_{i=1}^{2D}\left( \max\{\tilde{u}_i,\tilde{v}_i\}\right)^\gamma}
\end{align}
Of course, we can also combine these two kernels:
\begin{align}
epGMM(u,v;\gamma_1,\gamma_2) =  e^{-\gamma_2\left(1-\frac{\sum_{i=1}^{2D}\left(\min\{\tilde{u}_i,\tilde{v}_i\}\right)^{\gamma_1}}{\sum_{i=1}^{2D}\left( \max\{\tilde{u}_i,\tilde{v}_i\}\right)^{\gamma_1}}\right)}
\end{align}

In this study, we will provide an empirical study on kernel SVMs based on the above three tunable GMM kernels. Perhaps not surprisingly, the improvements can be substantial on some datasets. In particular, we will also compare them with deep nets and trees on 11 datasets~\cite{Proc:Larochelle_ICML07}. In their previous studies, \cite{Article:Li_ABC_arXiv08,Proc:ABC_ICML09,Proc:ABC_UAI10} developed tree methods including ``abc-mart'', ``robust logitboost'', and ``abc-robust-logitboost'' and demonstrated their excellent performance on those 11 datasets (and other datasets), by establishing the second-order tree-split formula and new derivatives for multi-class logistic loss. Compared to tree methods like ``abc-robust-logitboost'' (which are slow and need substantial model sizes), the proposed tunable GMM kernels produced largely comparable  classification results.

\newpage\clearpage

\section{An Experimental Study on Kernel SVMs}\label{sec_kernel}

We essentially use similar datasets as in~\cite{Report:Li_GMM16}. Table~\ref{tab_data} lists a large number of publicly available  datasets from the UCI repository and Table~\ref{tab_data2} presents datasets from the LIBSVM website and the 11 datasets for testing deep learning methods and trees~\cite{Proc:Larochelle_ICML07,Proc:ABC_UAI10}. In both tables, we report
the kernel SVM test classification results for the linear kernel, the best-tuned RBF kernel, the original (tuning-free) GMM kernel, the best-tuned eGMM kernel, and the pGMM kernel.  For the epGMM kernel, the experimental results are reported in Section~\ref{sec_epGMM}, e.g., Table~\ref{tab_epGMM}.\\

\begin{figure}[h!]
\begin{center}
\mbox{
\includegraphics[width=2.3in]{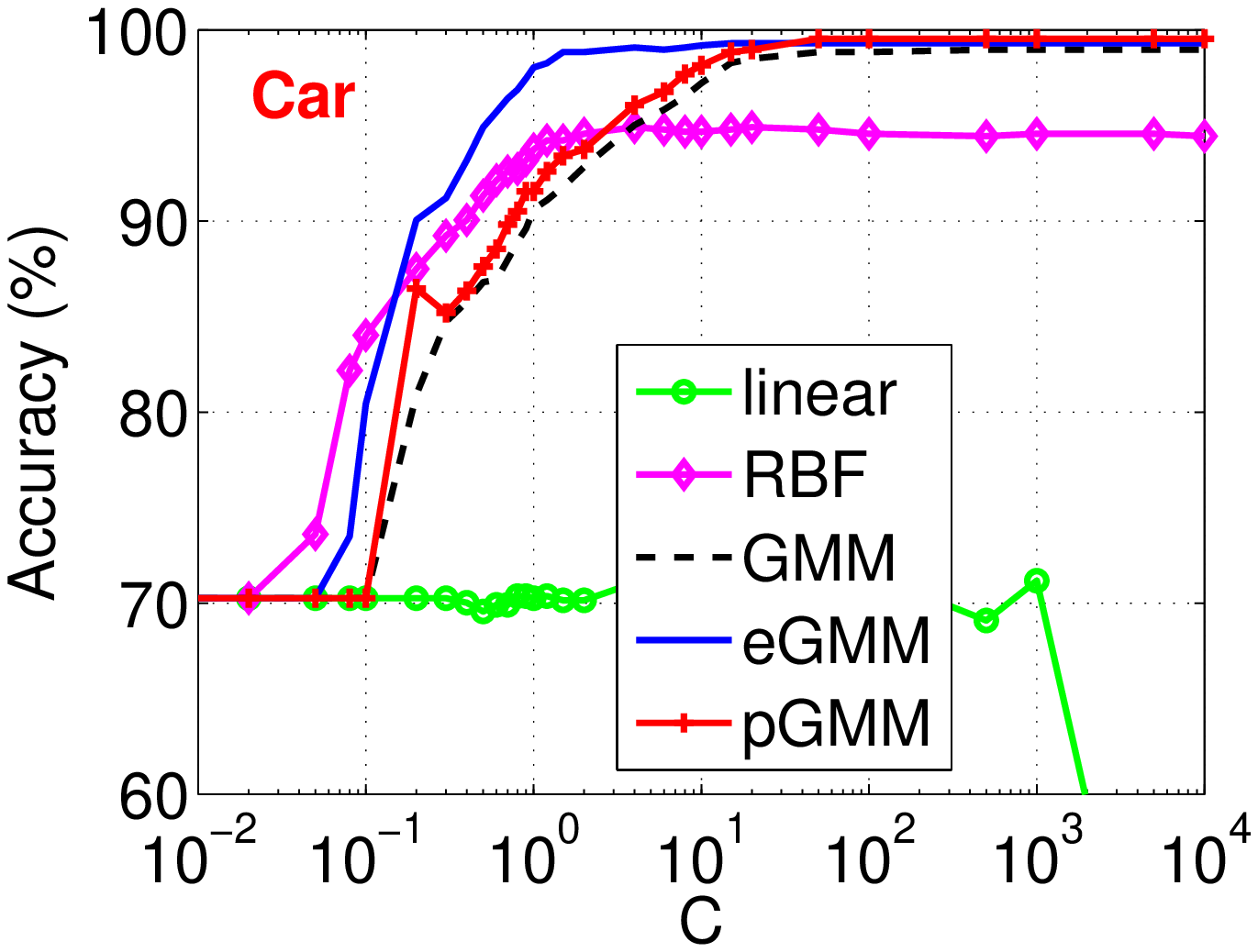}\hspace{-0.14in}
\includegraphics[width=2.3in]{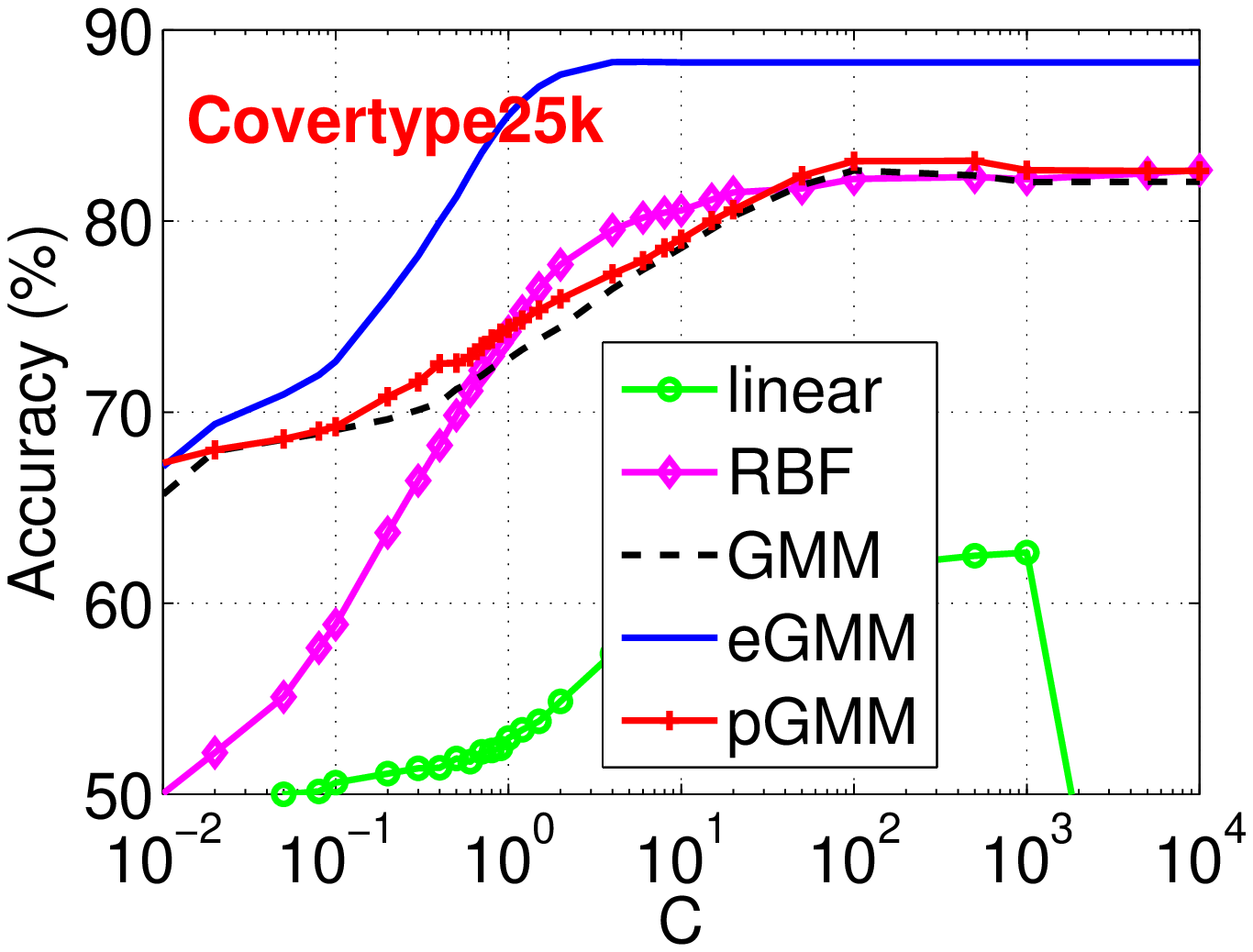}\hspace{-0.14in}
\includegraphics[width=2.3in]{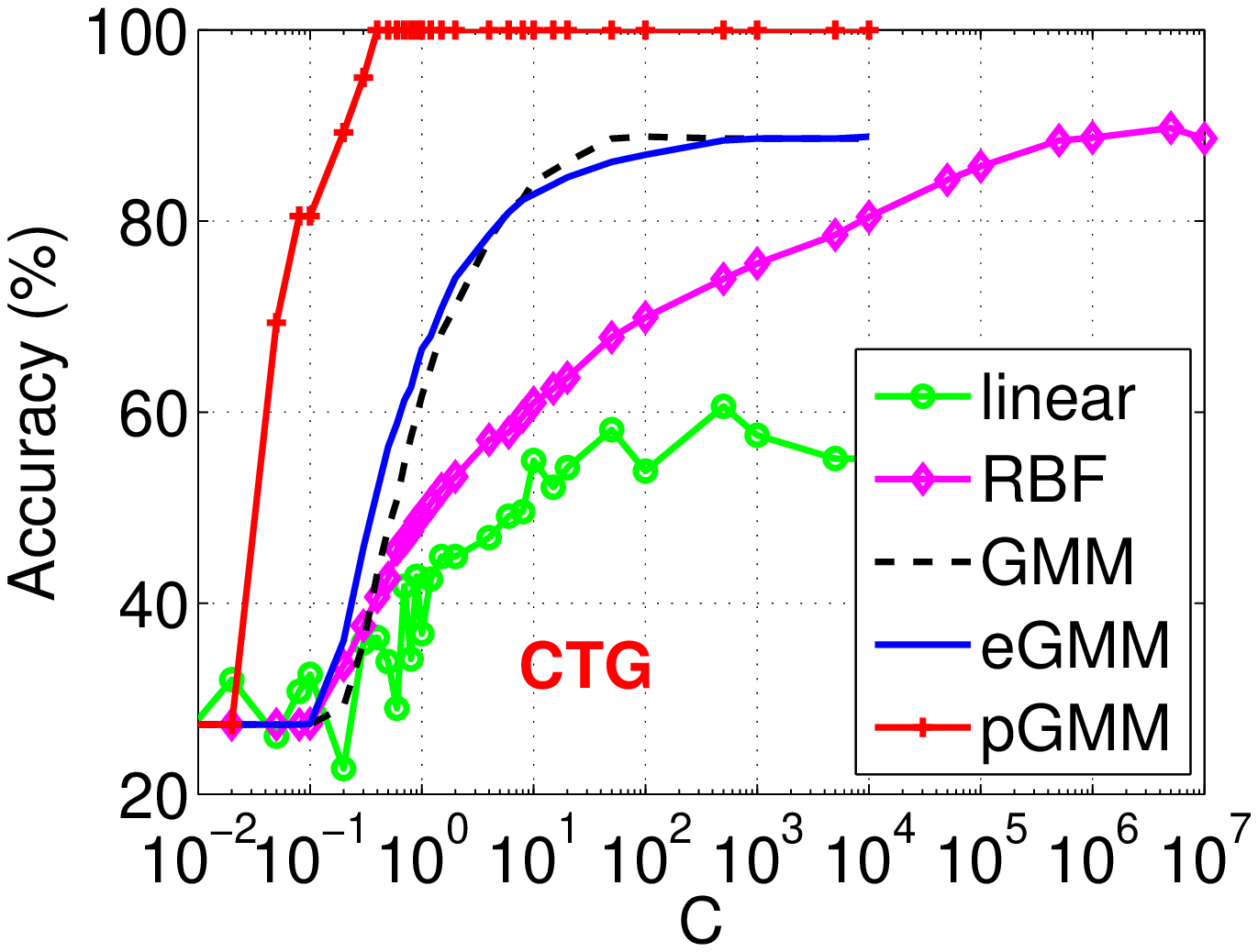}
}

\mbox{
\includegraphics[width=2.3in]{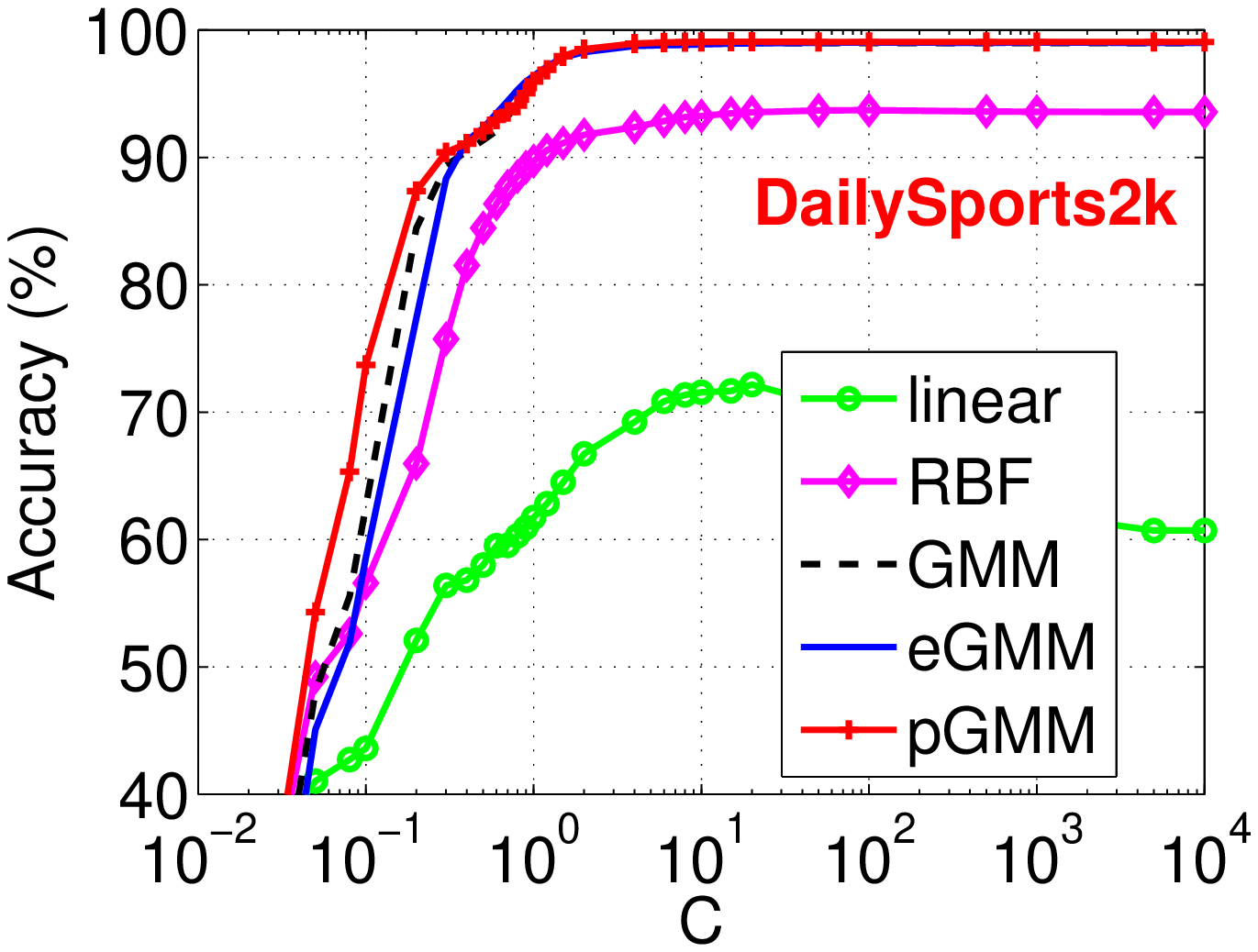}\hspace{-0.14in}
\includegraphics[width=2.3in]{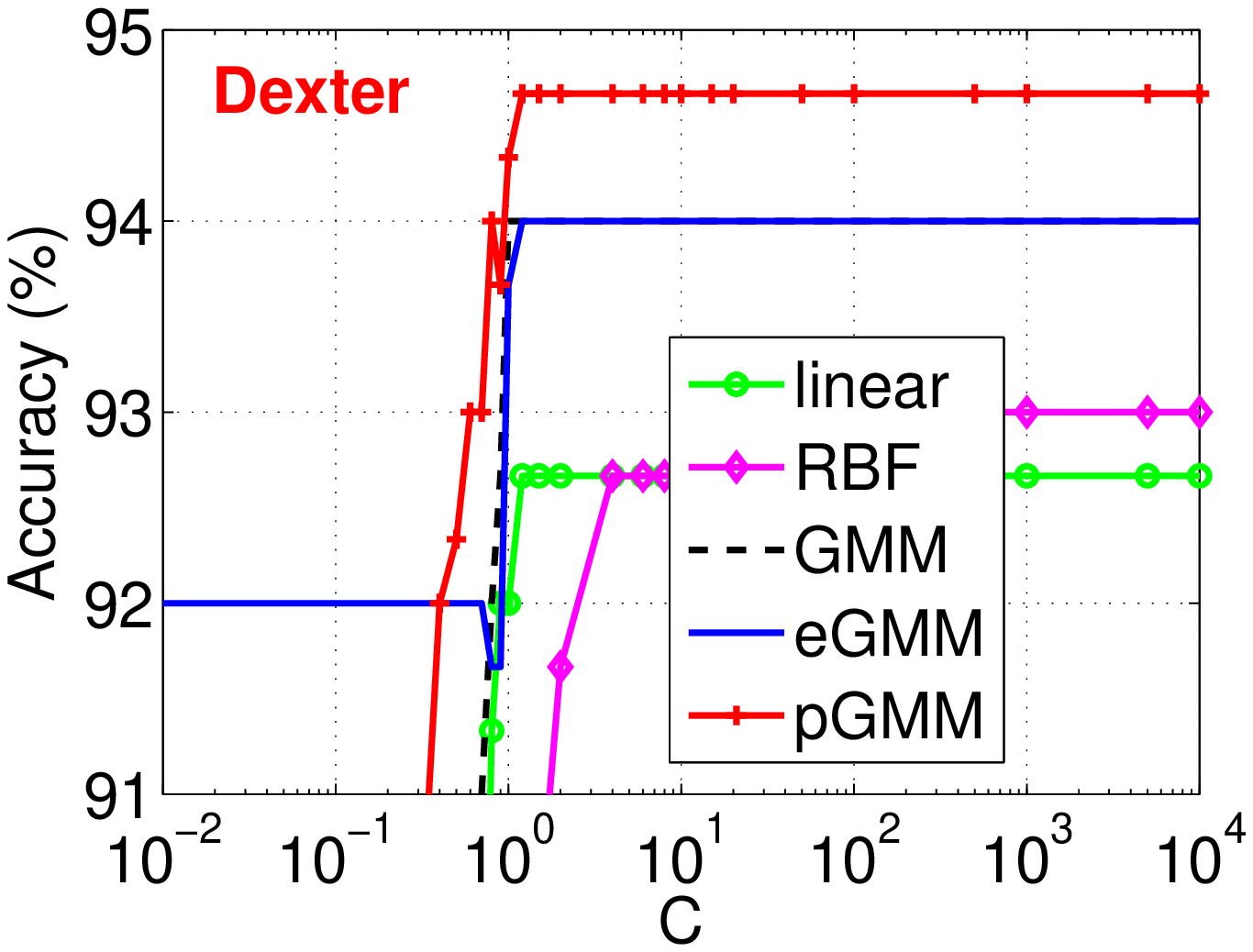}\hspace{-0.14in}
\includegraphics[width=2.3in]{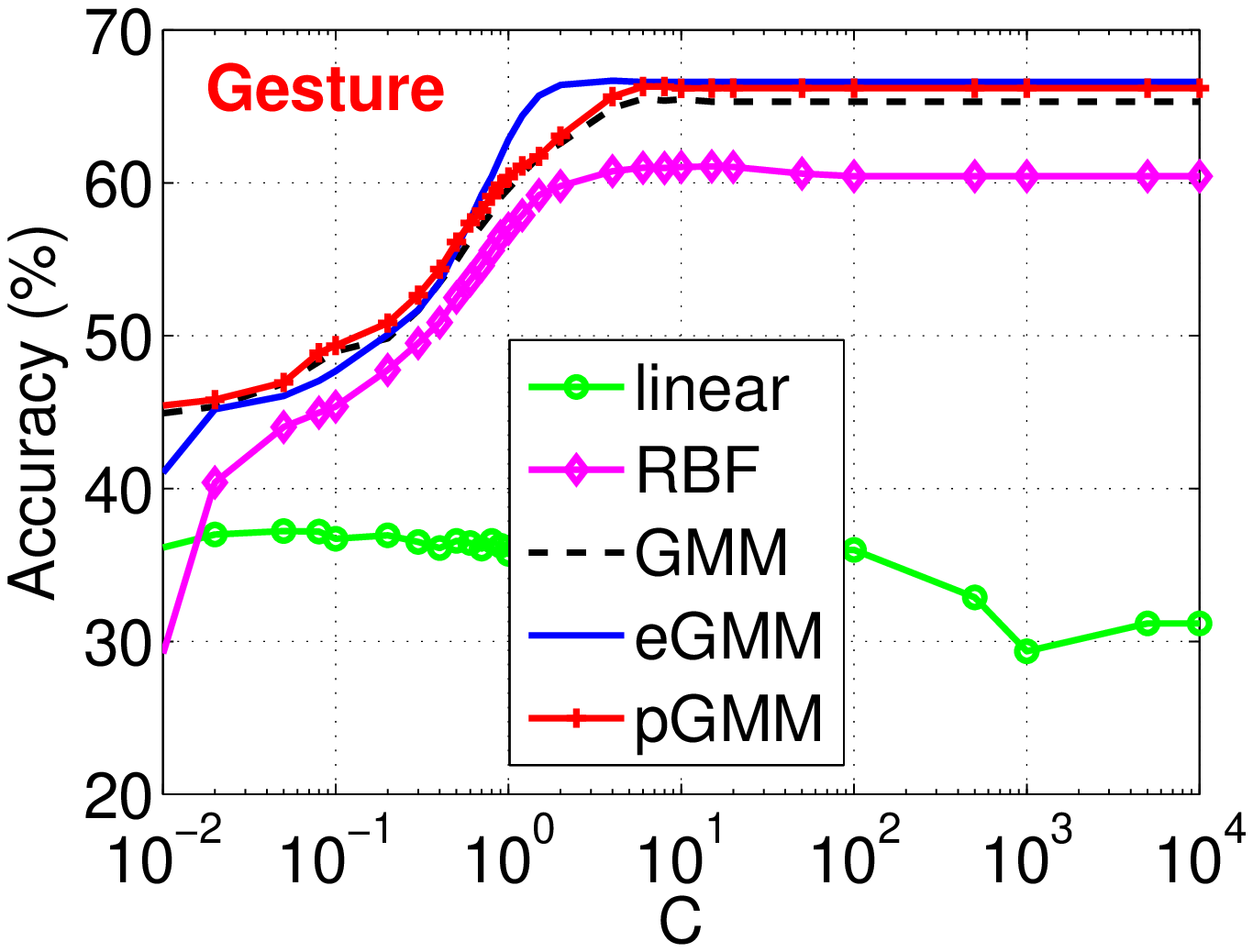}
}

\mbox{

\includegraphics[width=2.3in]{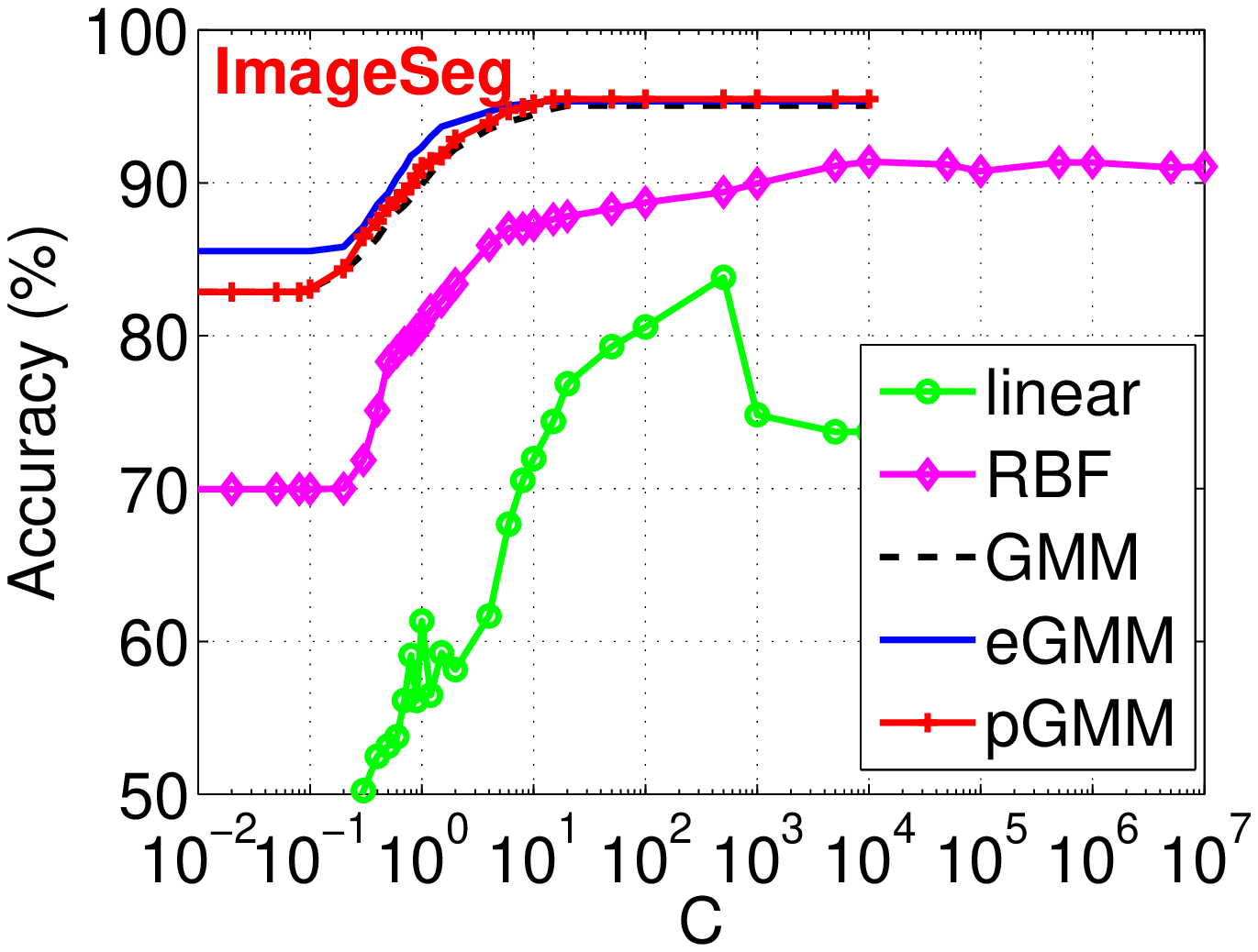}\hspace{-0.14in}
\includegraphics[width=2.3in]{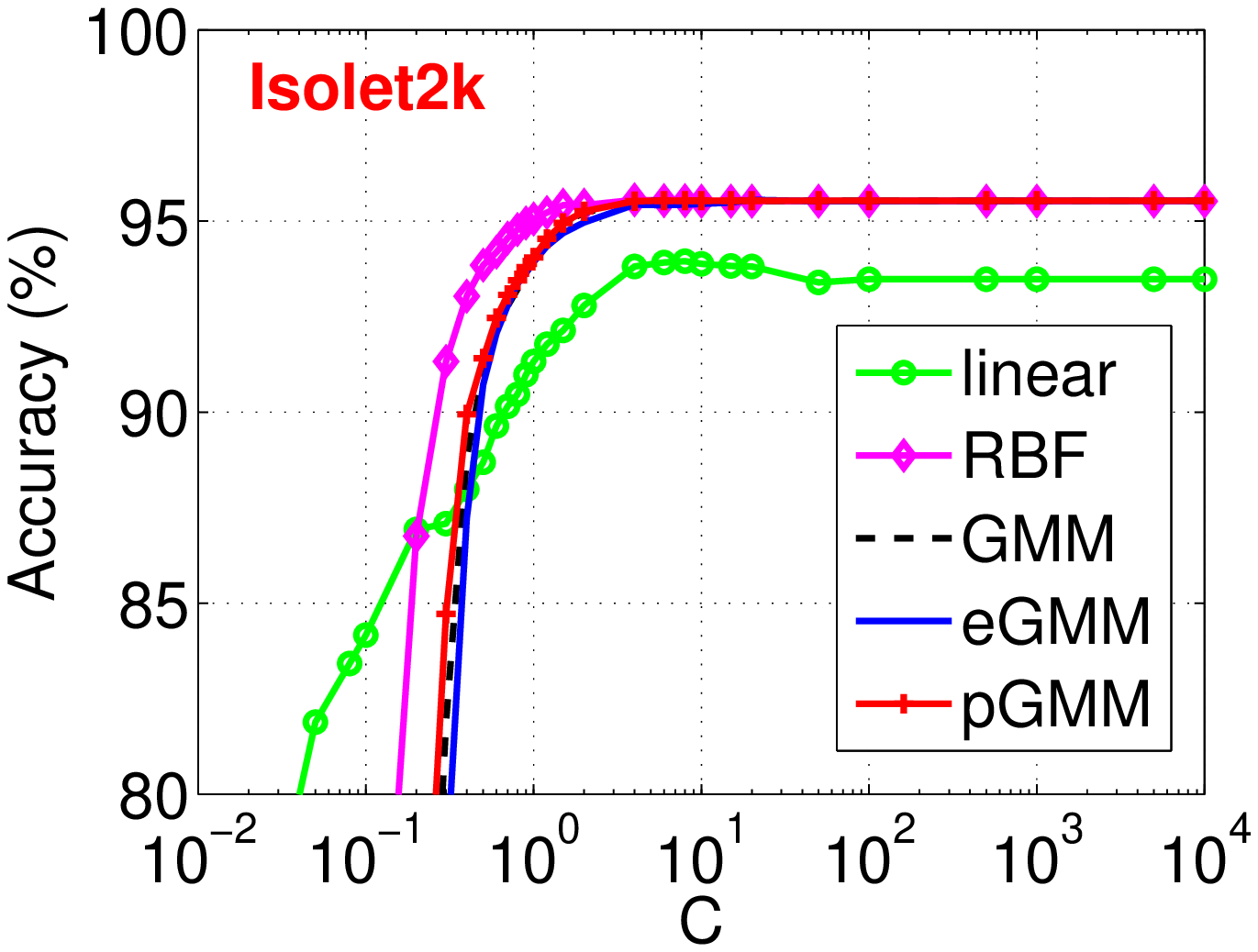}\hspace{-0.14in}
\includegraphics[width=2.3in]{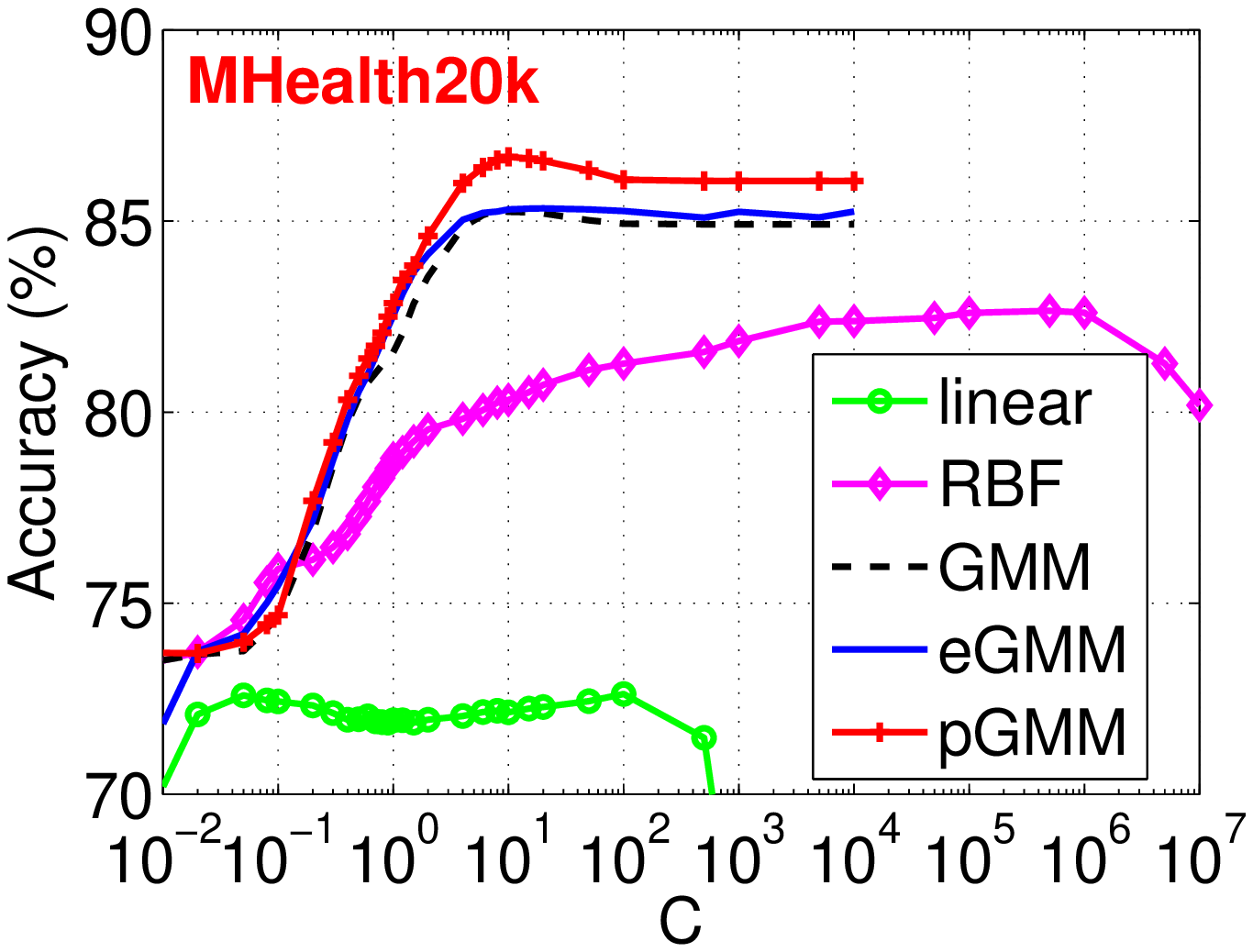}
}

\mbox{
\includegraphics[width=2.3in]{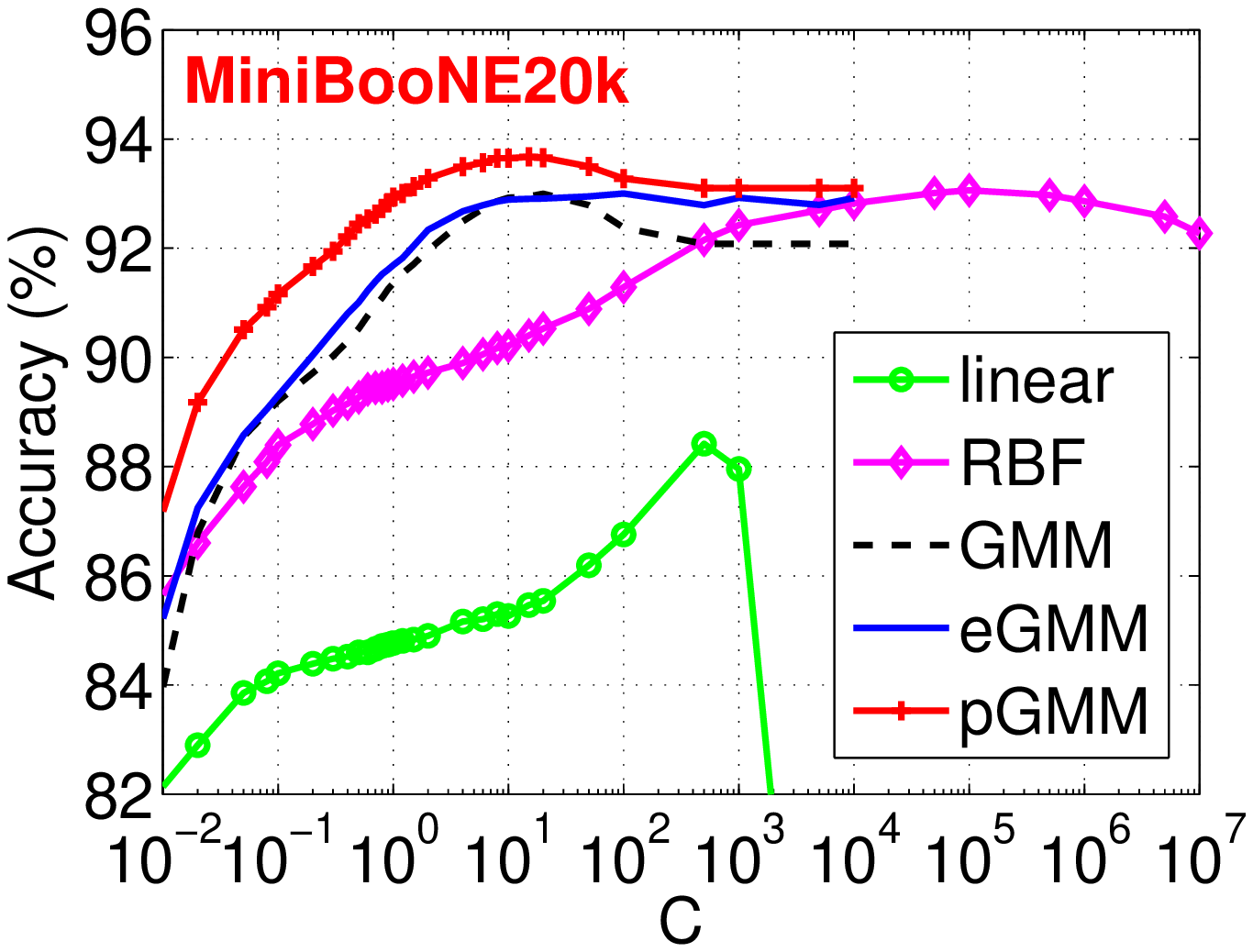}\hspace{-0.14in}
\includegraphics[width=2.3in]{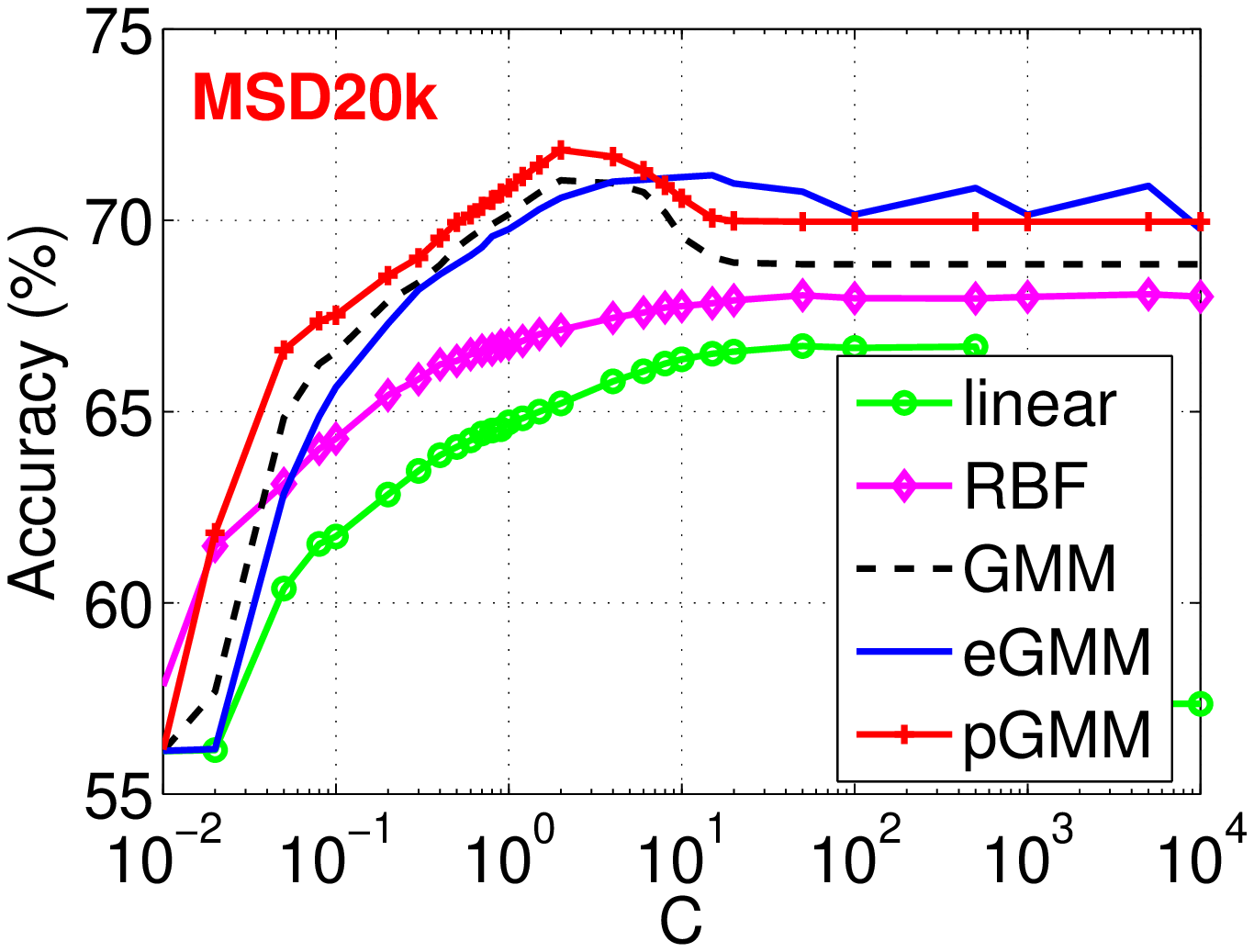}
\includegraphics[width=2.3in]{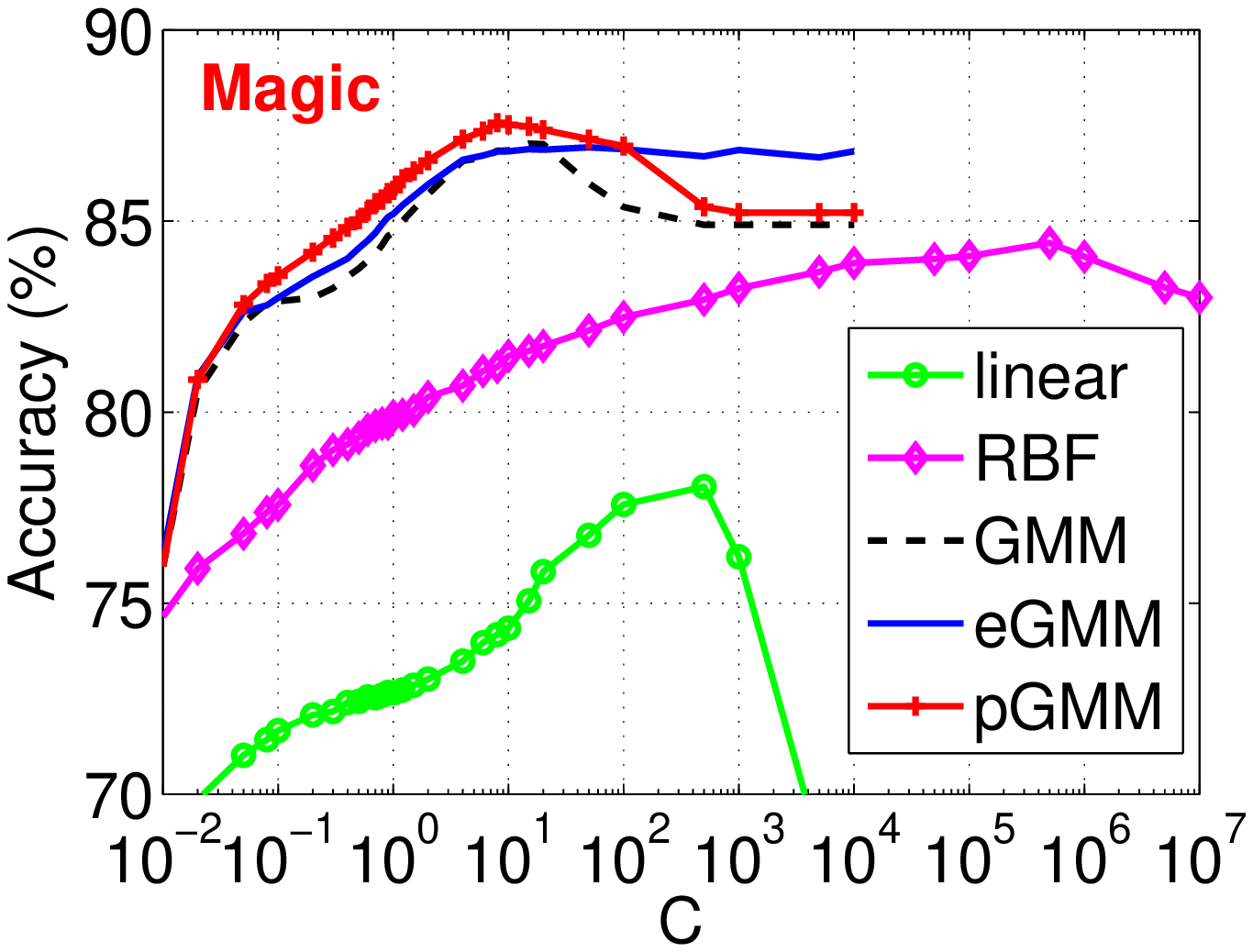}
}

\mbox{

\includegraphics[width=2.3in]{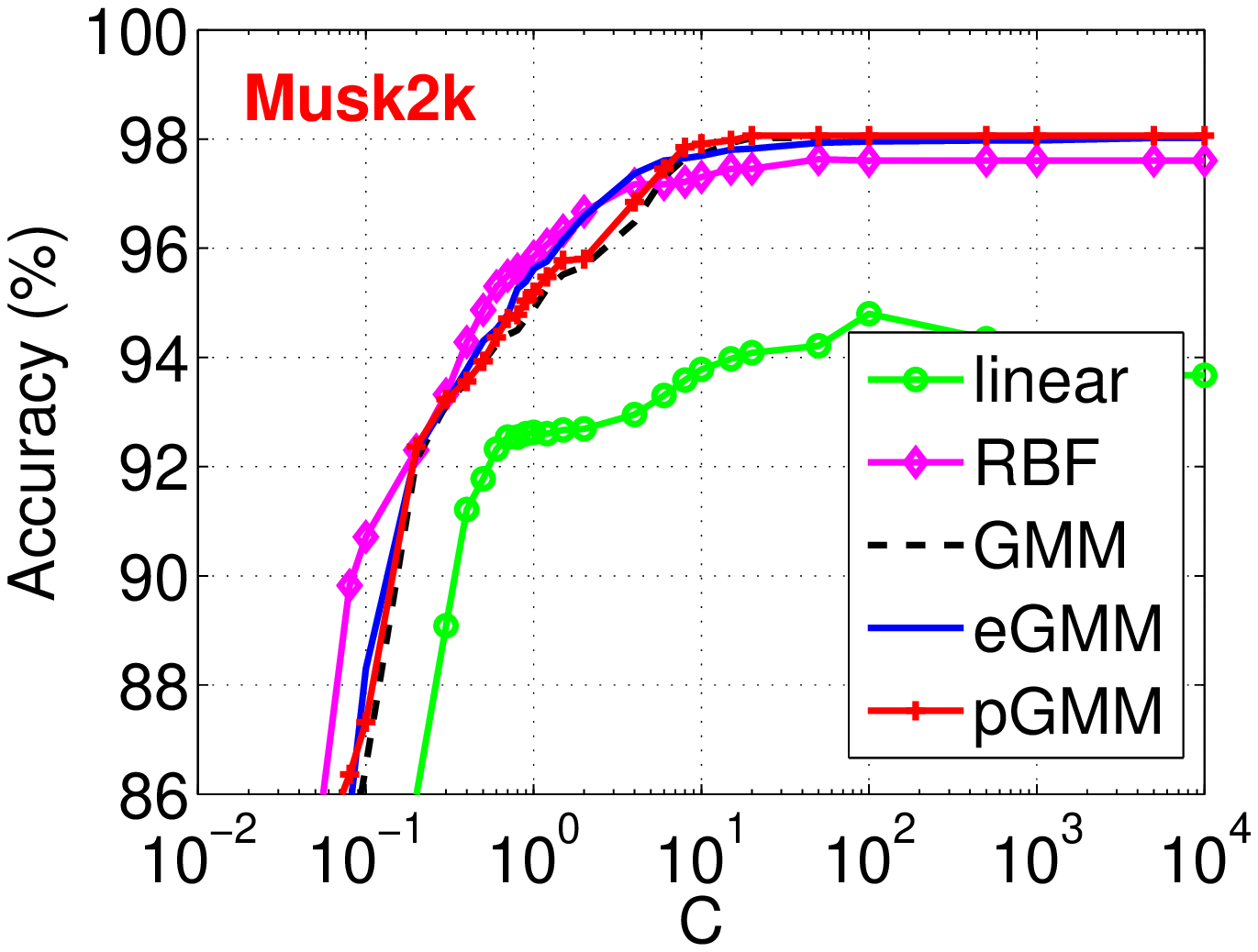}\hspace{-0.14in}
\includegraphics[width=2.3in]{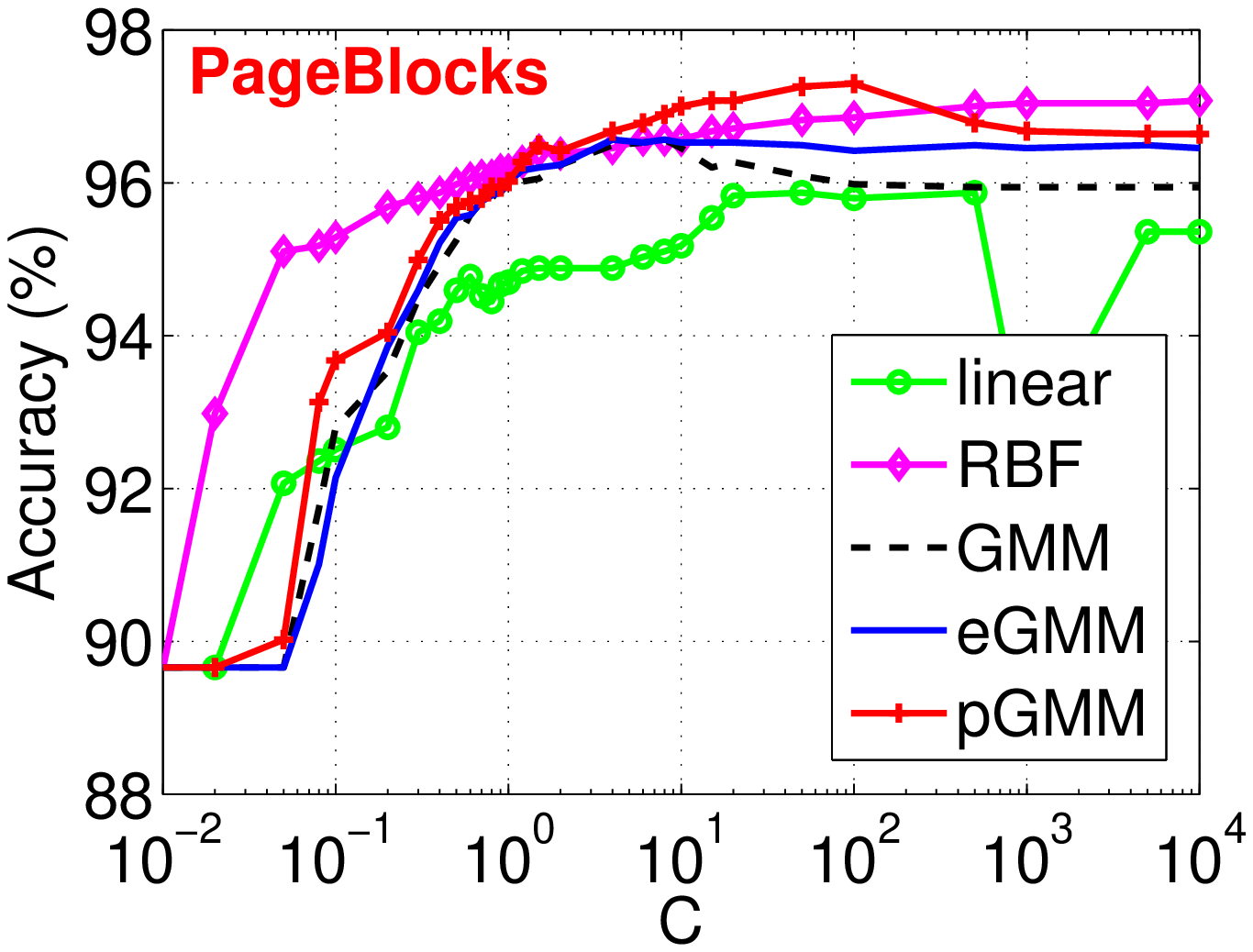}\hspace{-0.14in}
\includegraphics[width=2.3in]{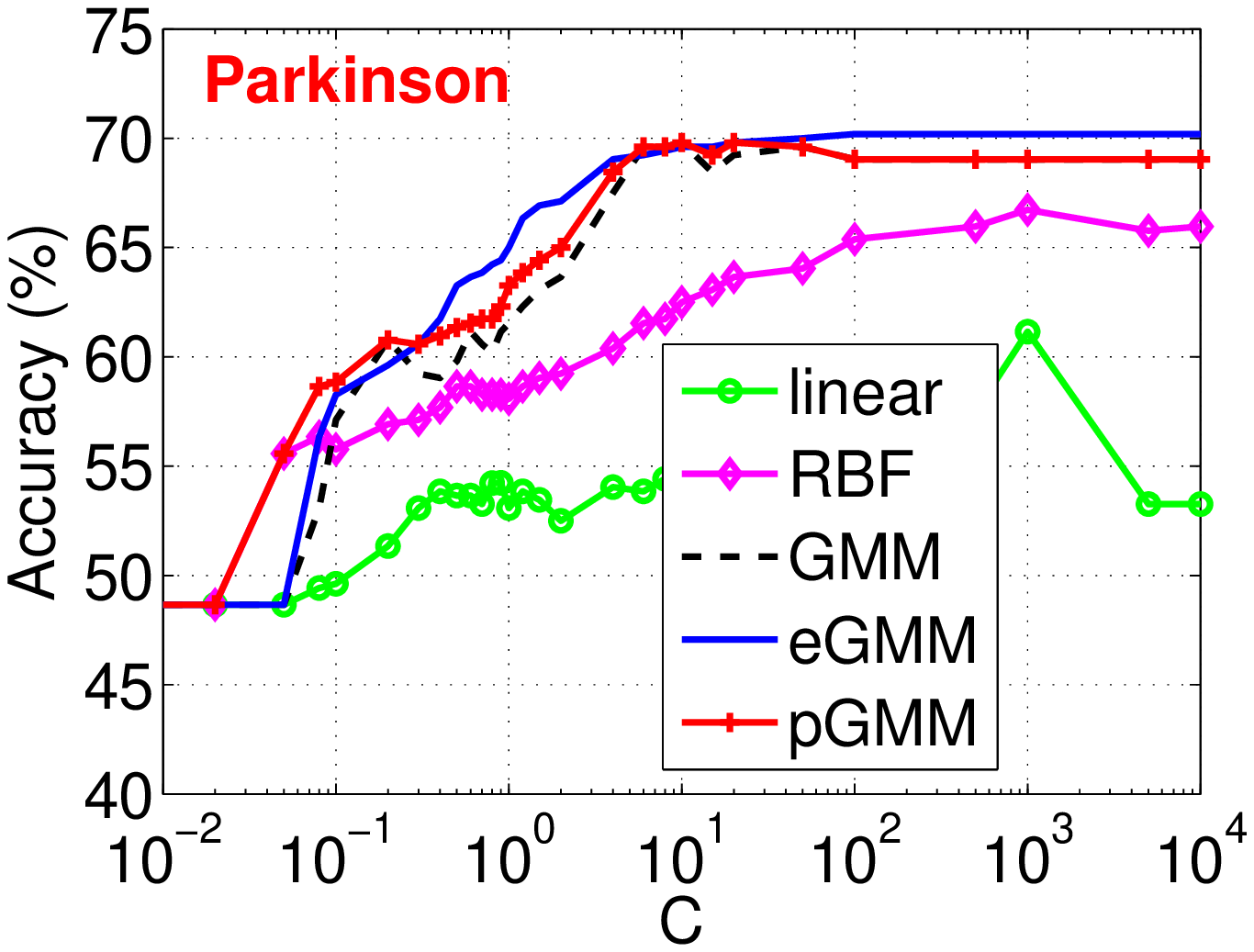}
}

\end{center}
\vspace{-0.3in}
\caption{Test classification accuracies of various kernels using LIBSVM pre-computed kernel functionality. The results are presented  with respect to $C$, which is the $l_2$-regularized kernel SVM parameter. For RBF, eGMM, and pGMM, at each $C$, we report the best test accuracies from a wide range of kernel parameter ($\gamma$) values.}\label{fig_SVM1}
\end{figure}

\begin{figure}[h!]
\begin{center}

\mbox{
\includegraphics[width=2.3in]{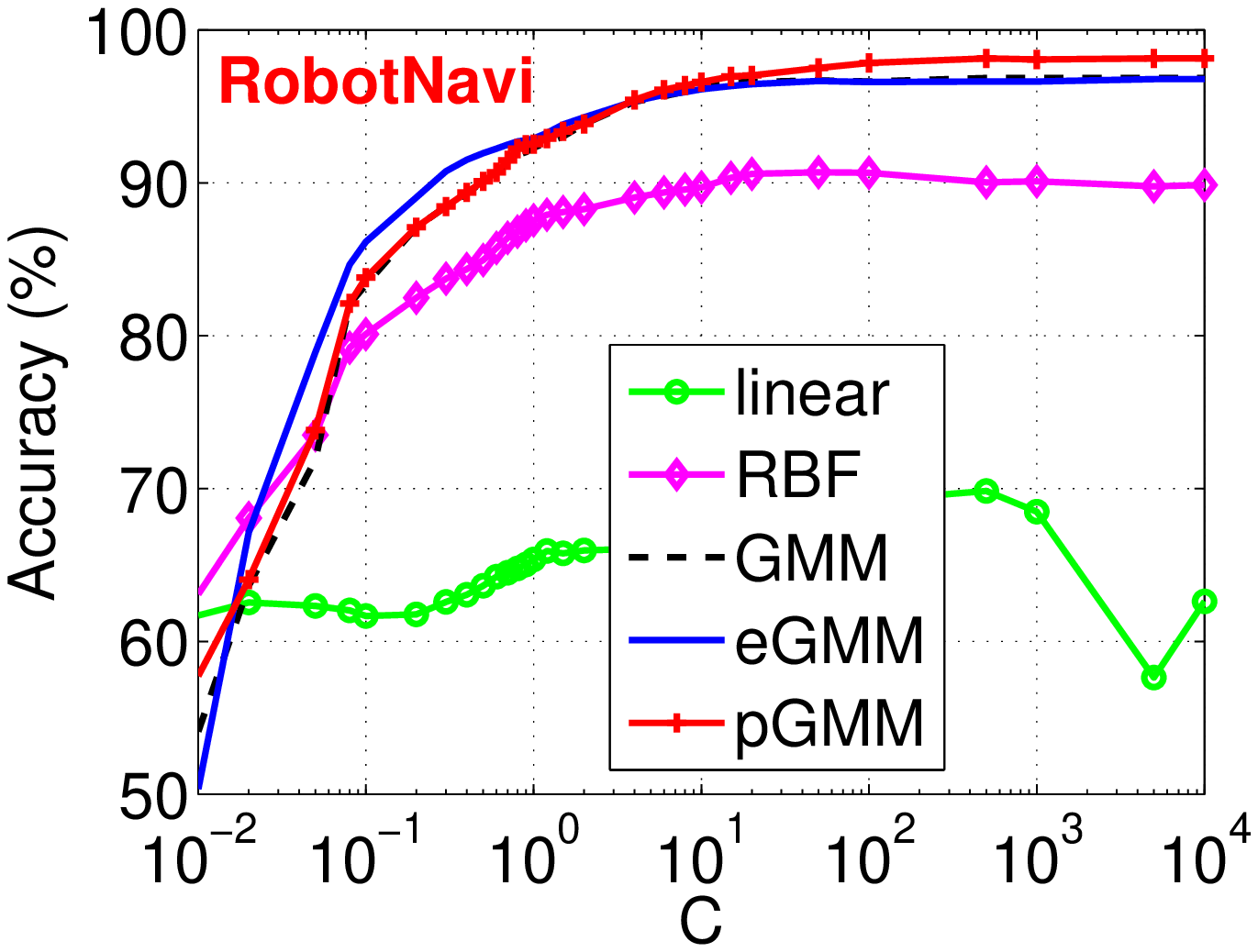}\hspace{-0.14in}
\includegraphics[width=2.3in]{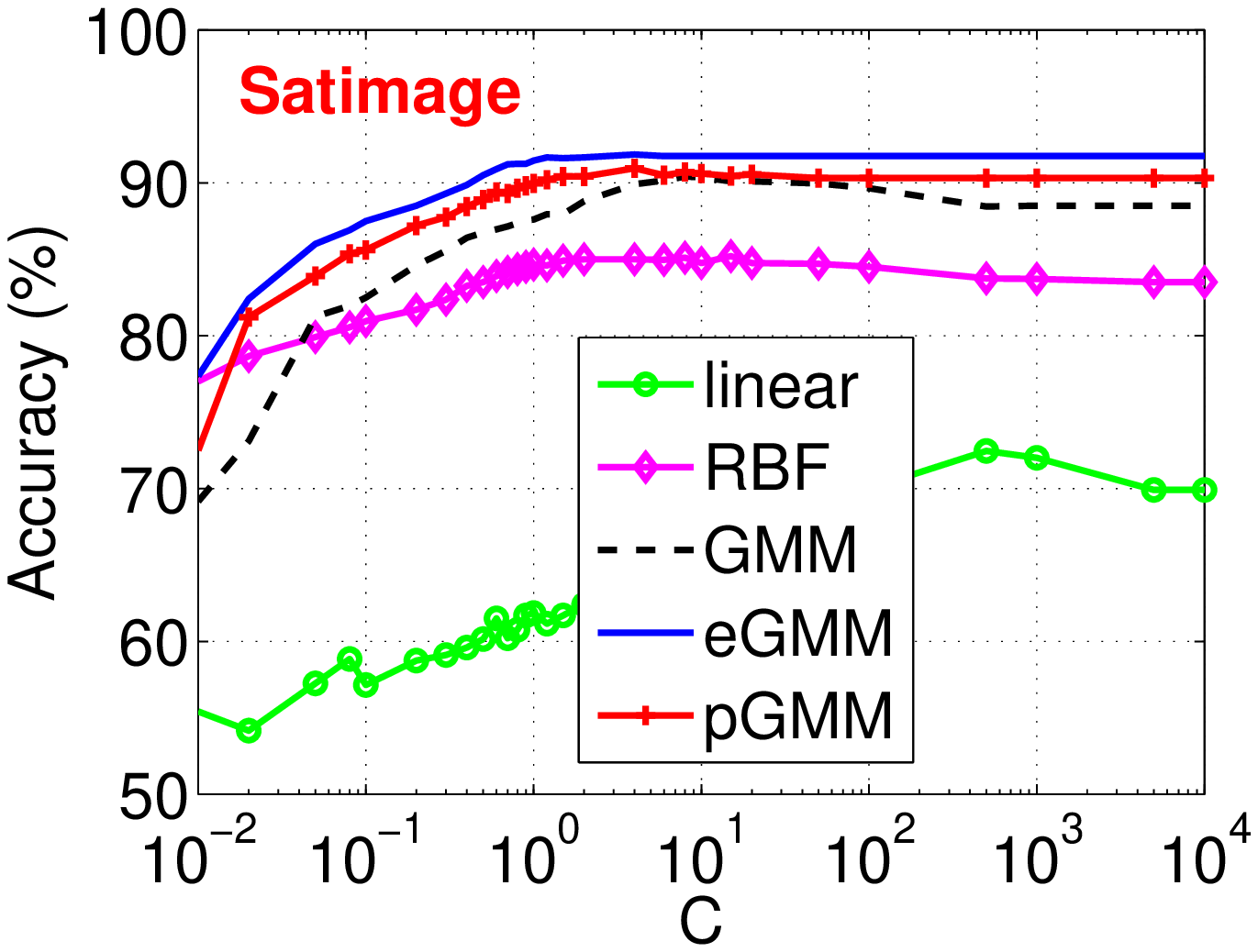}\hspace{-0.14in}
\includegraphics[width=2.3in]{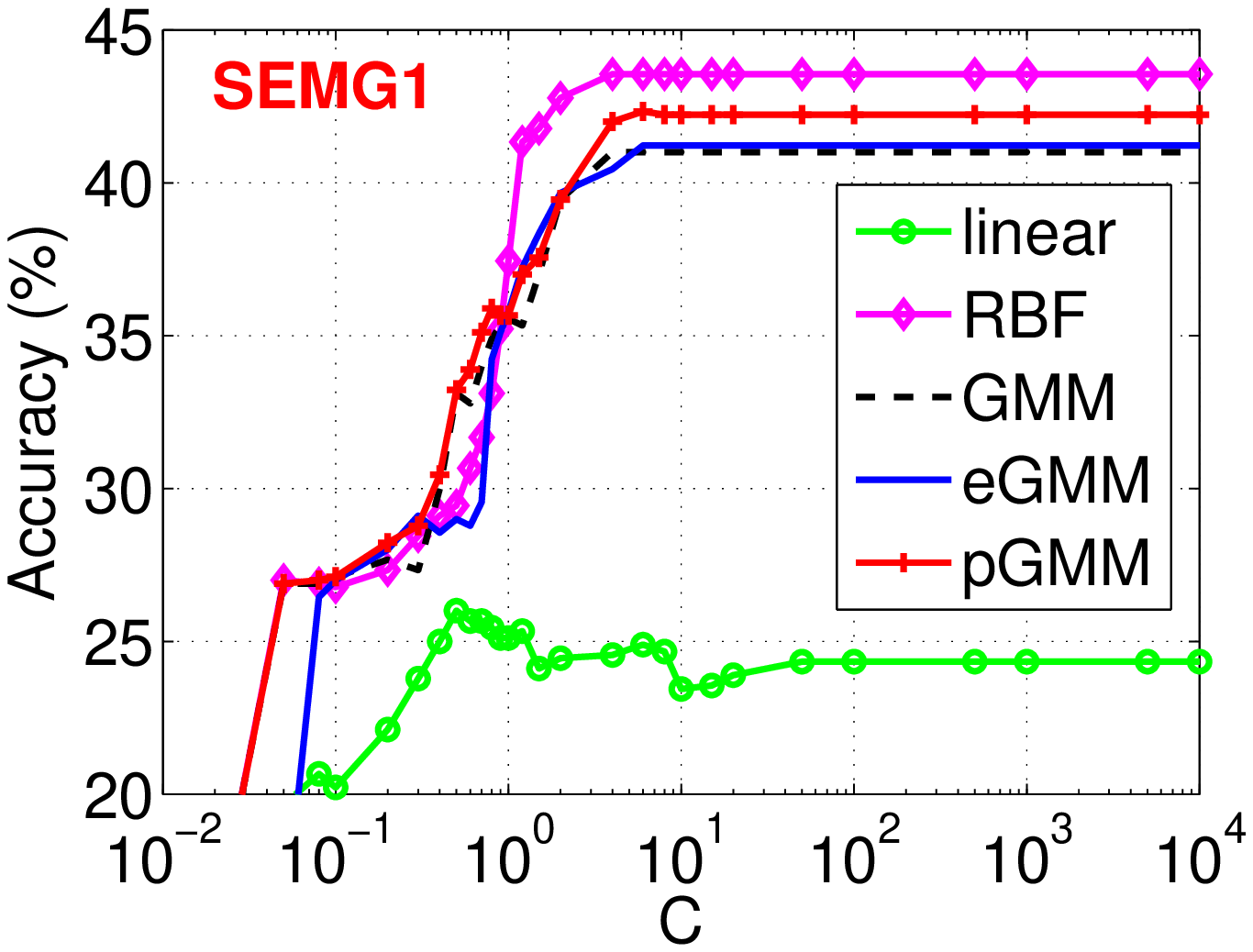}
}

\mbox{
\includegraphics[width=2.3in]{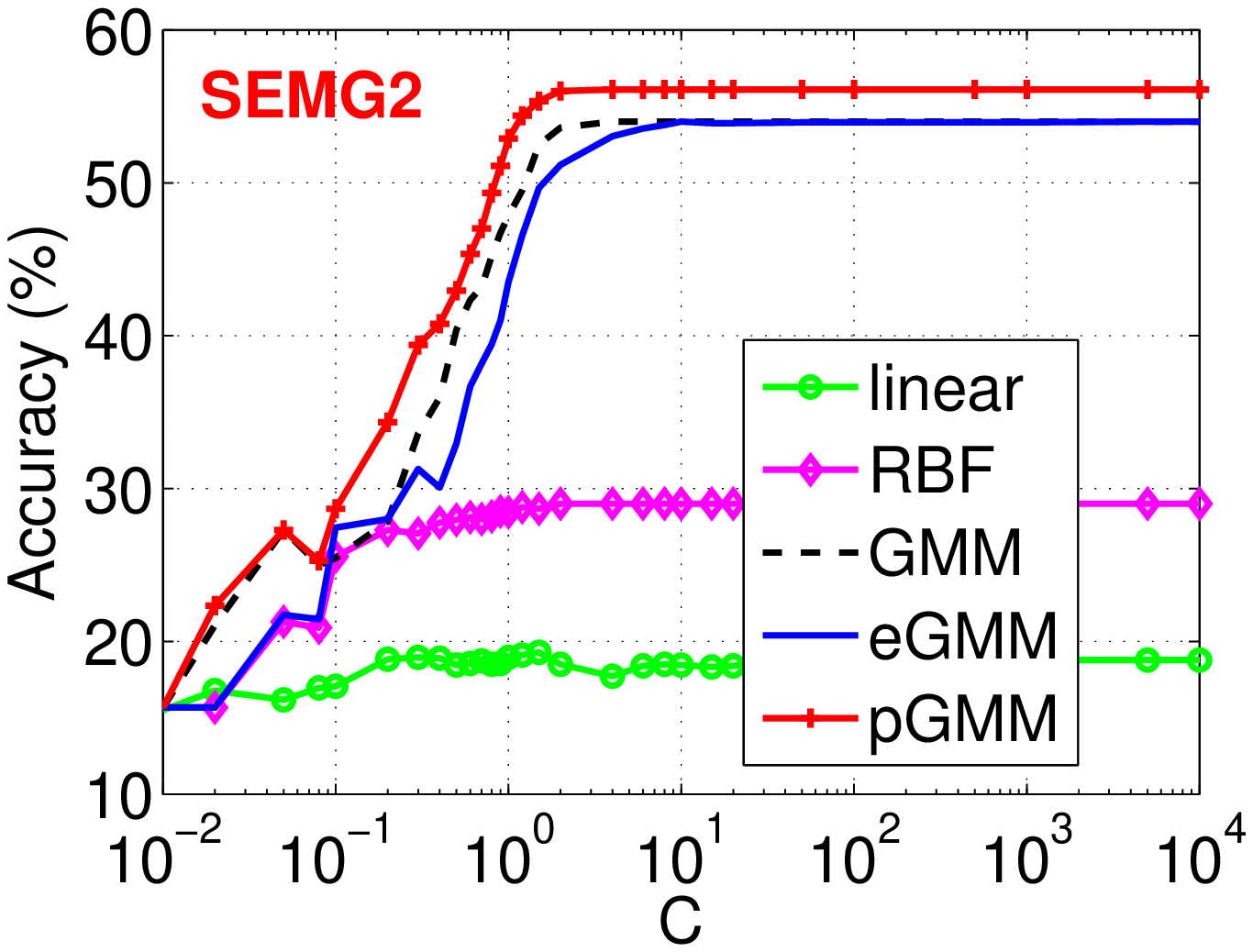}\hspace{-0.14in}
\includegraphics[width=2.3in]{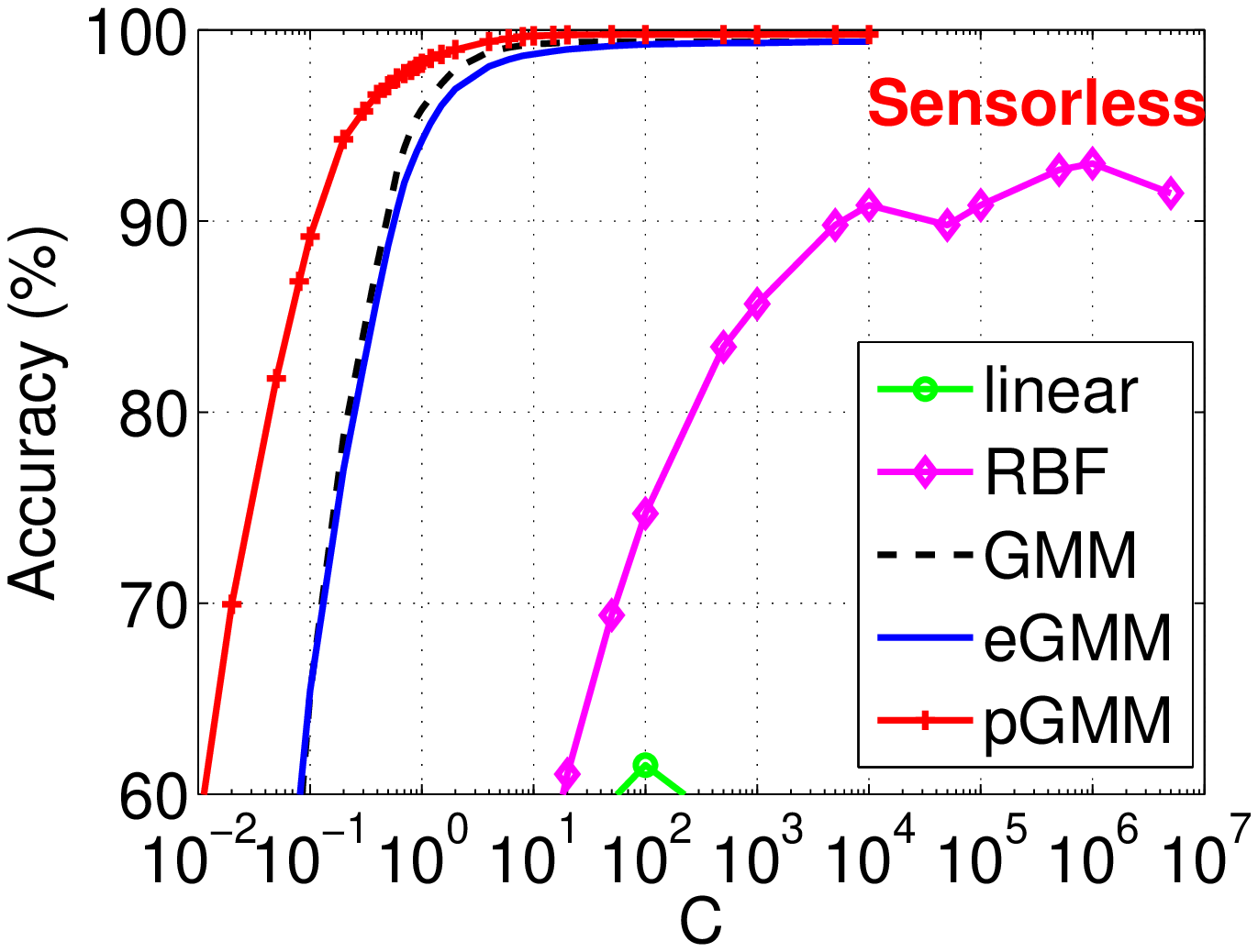}\hspace{-0.14in}
\includegraphics[width=2.3in]{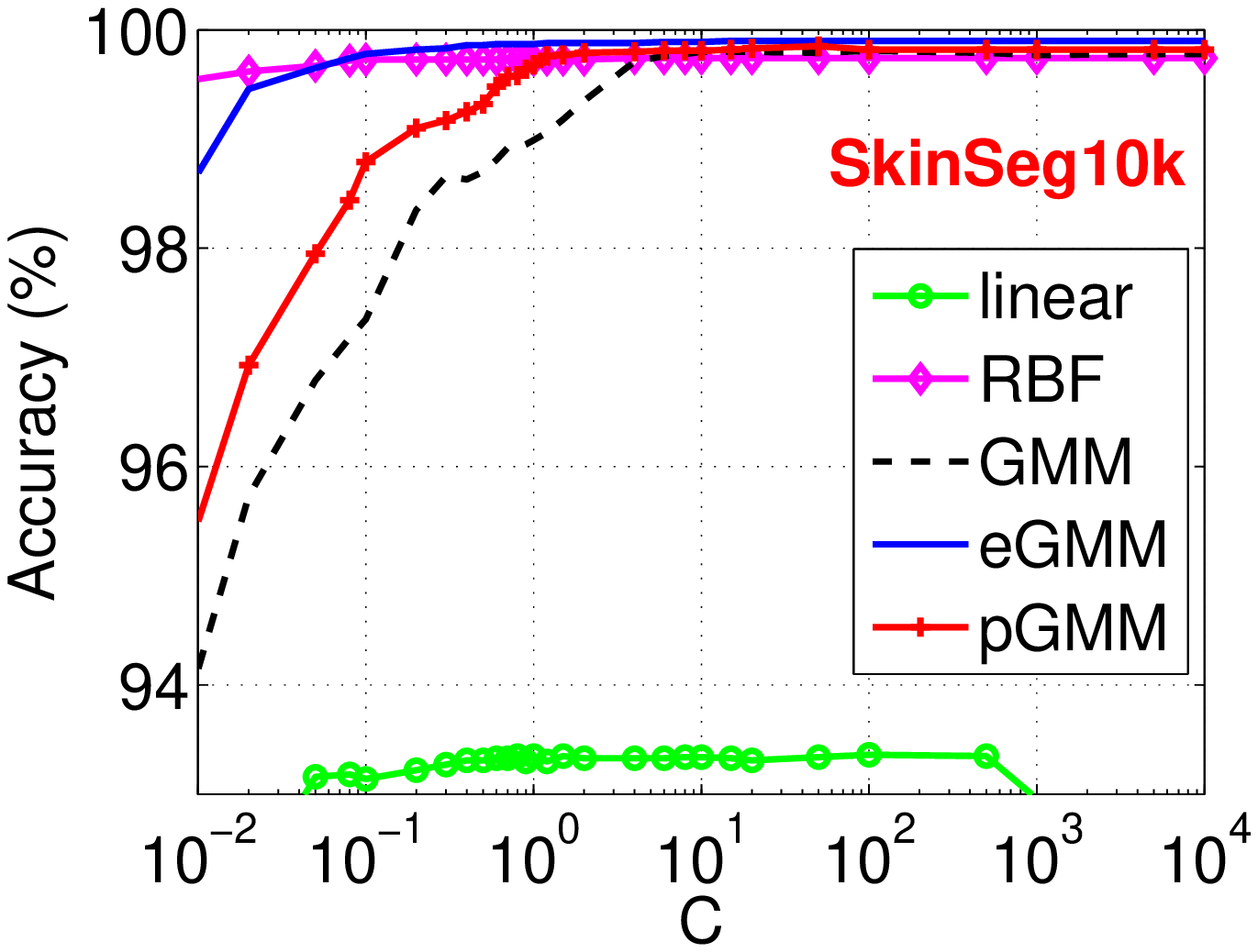}
}

\mbox{
\includegraphics[width=2.3in]{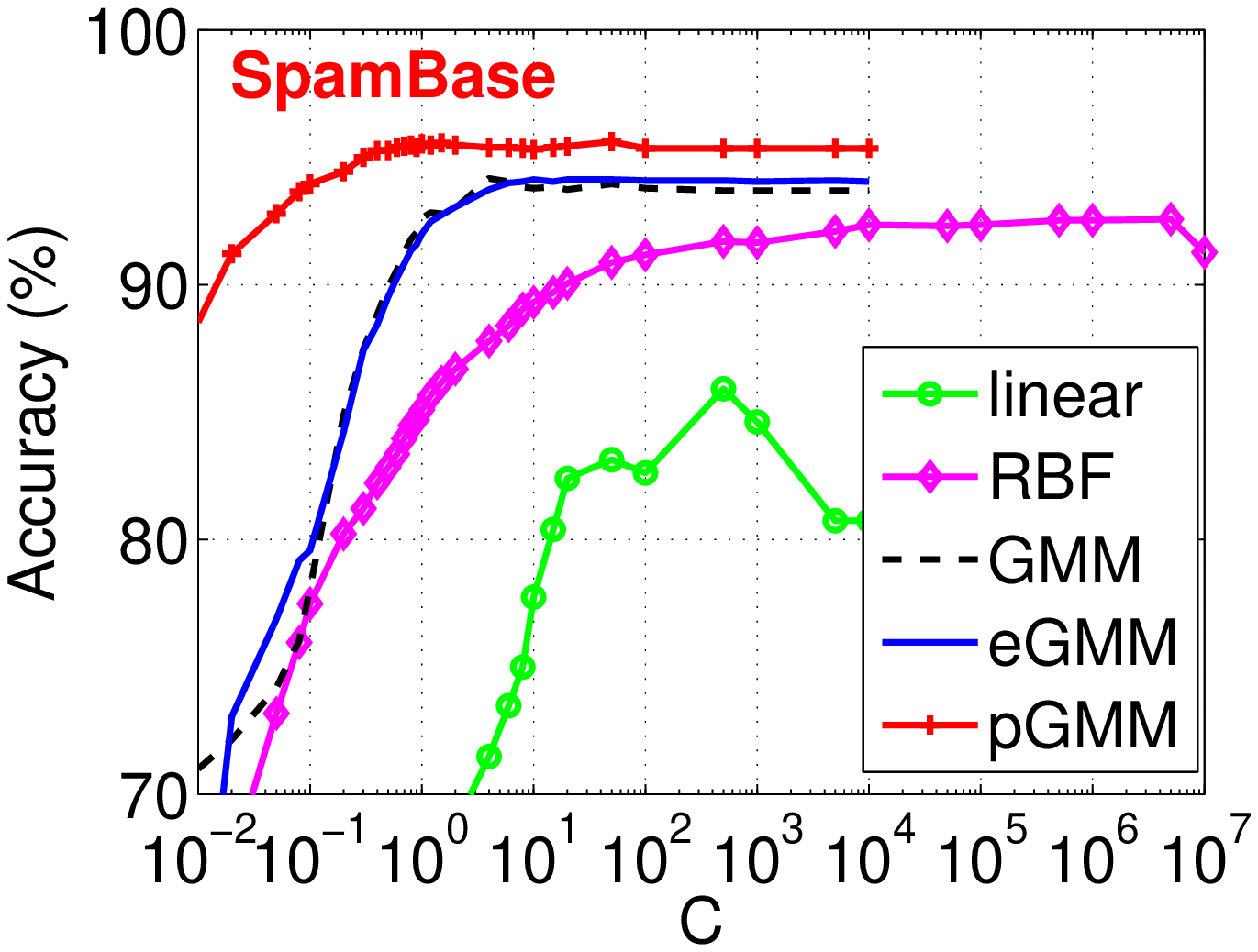}
\includegraphics[width=2.3in]{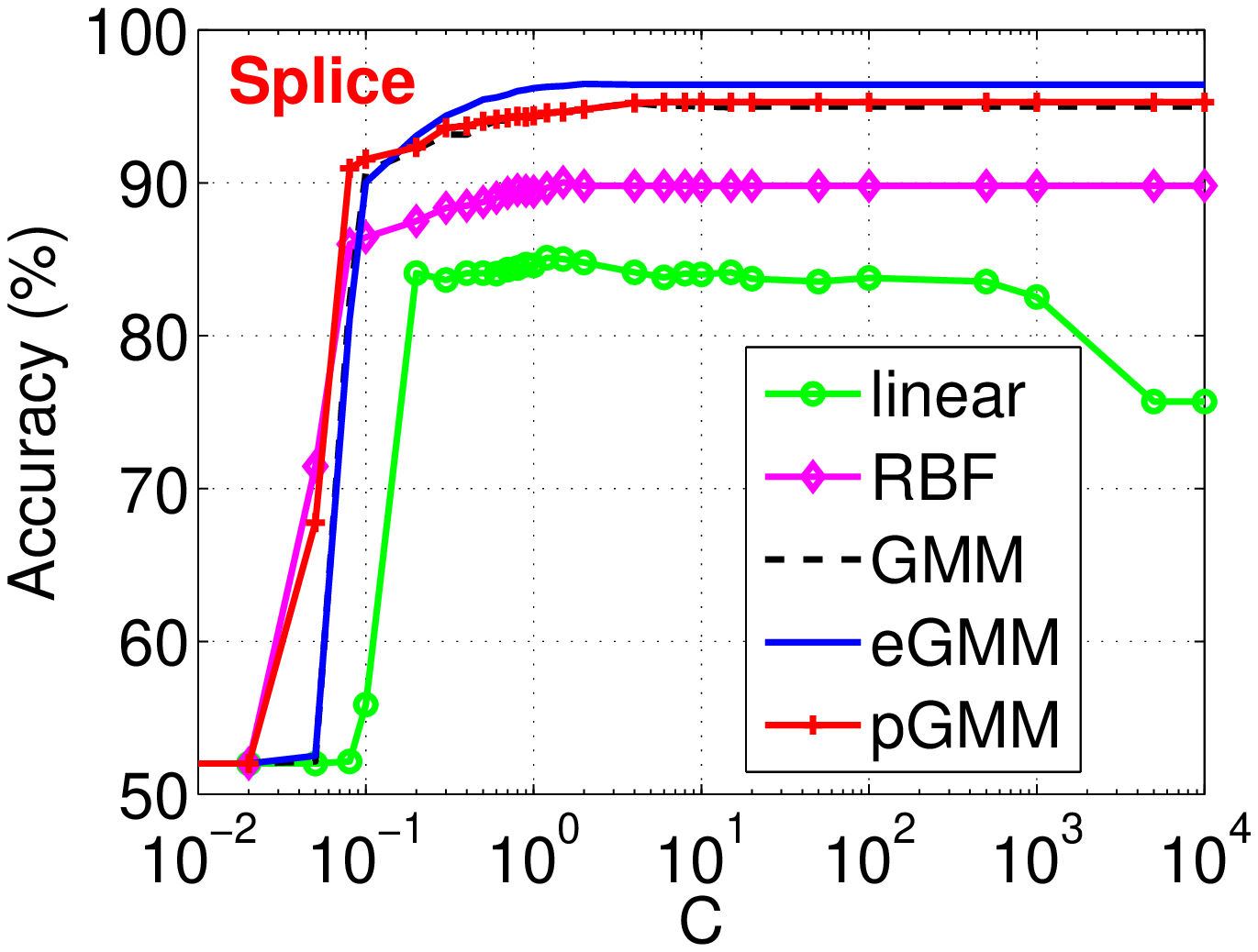}\hspace{-0.14in}
\includegraphics[width=2.3in]{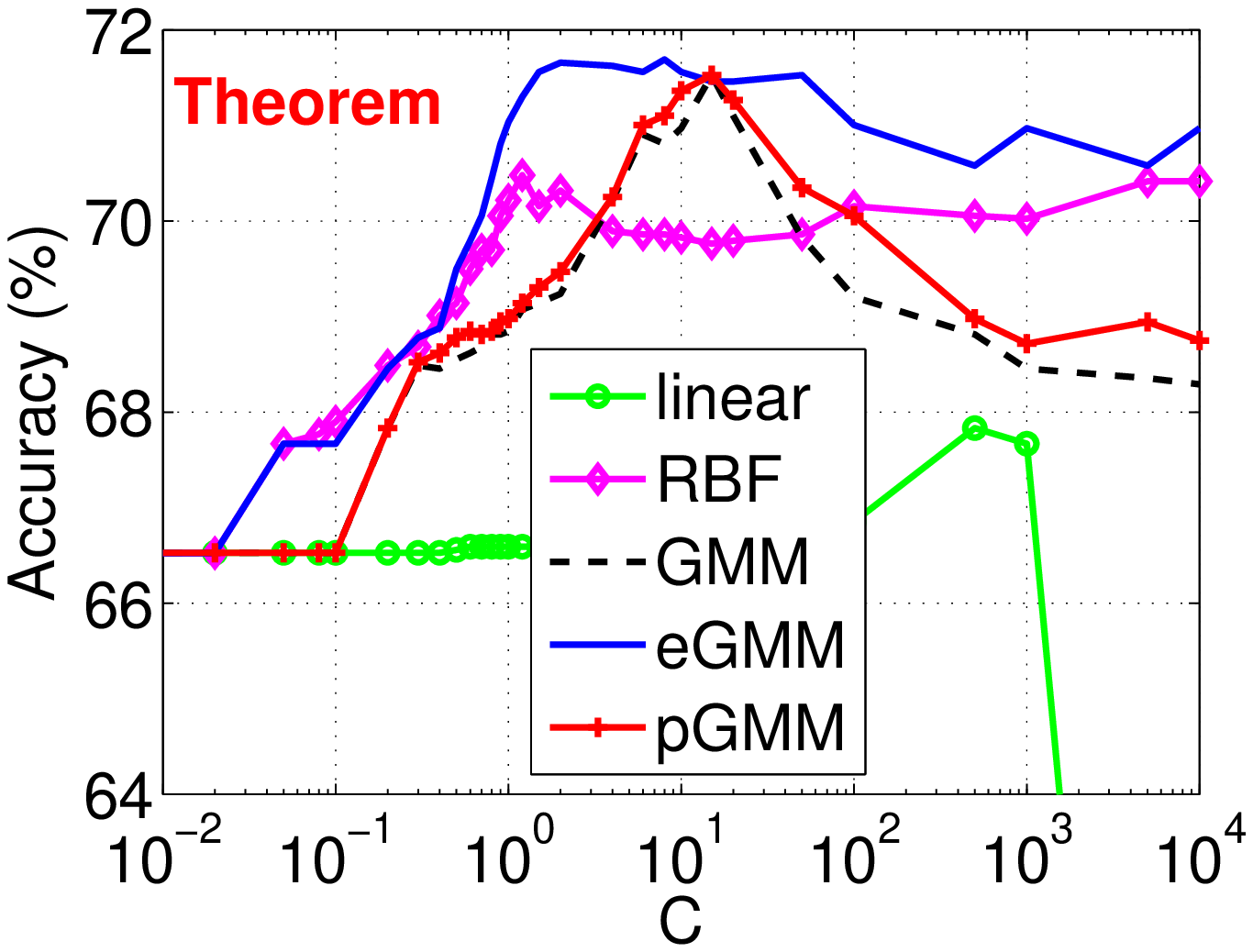}
}

\mbox{
\includegraphics[width=2.3in]{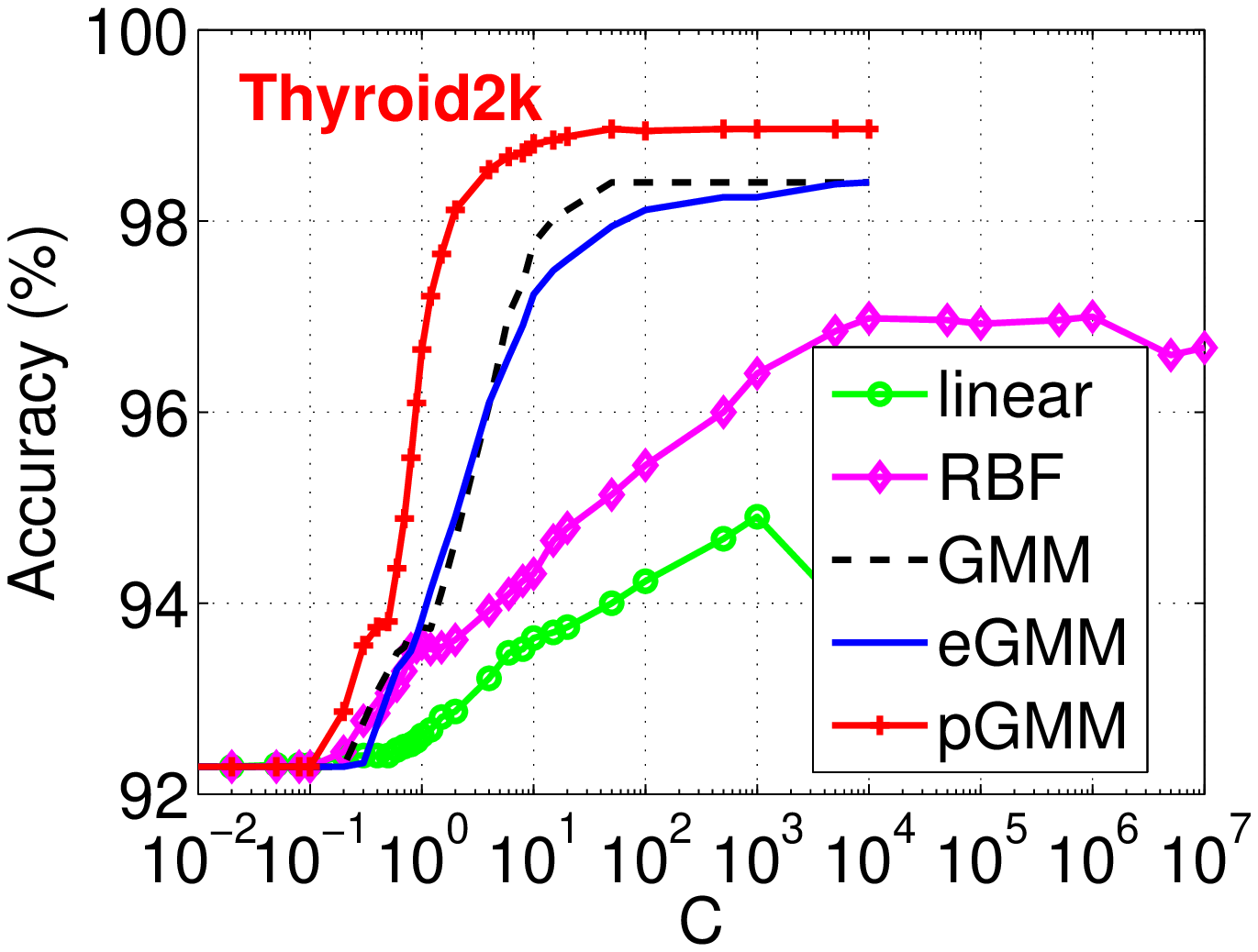}\hspace{-0.14in}
\includegraphics[width=2.3in]{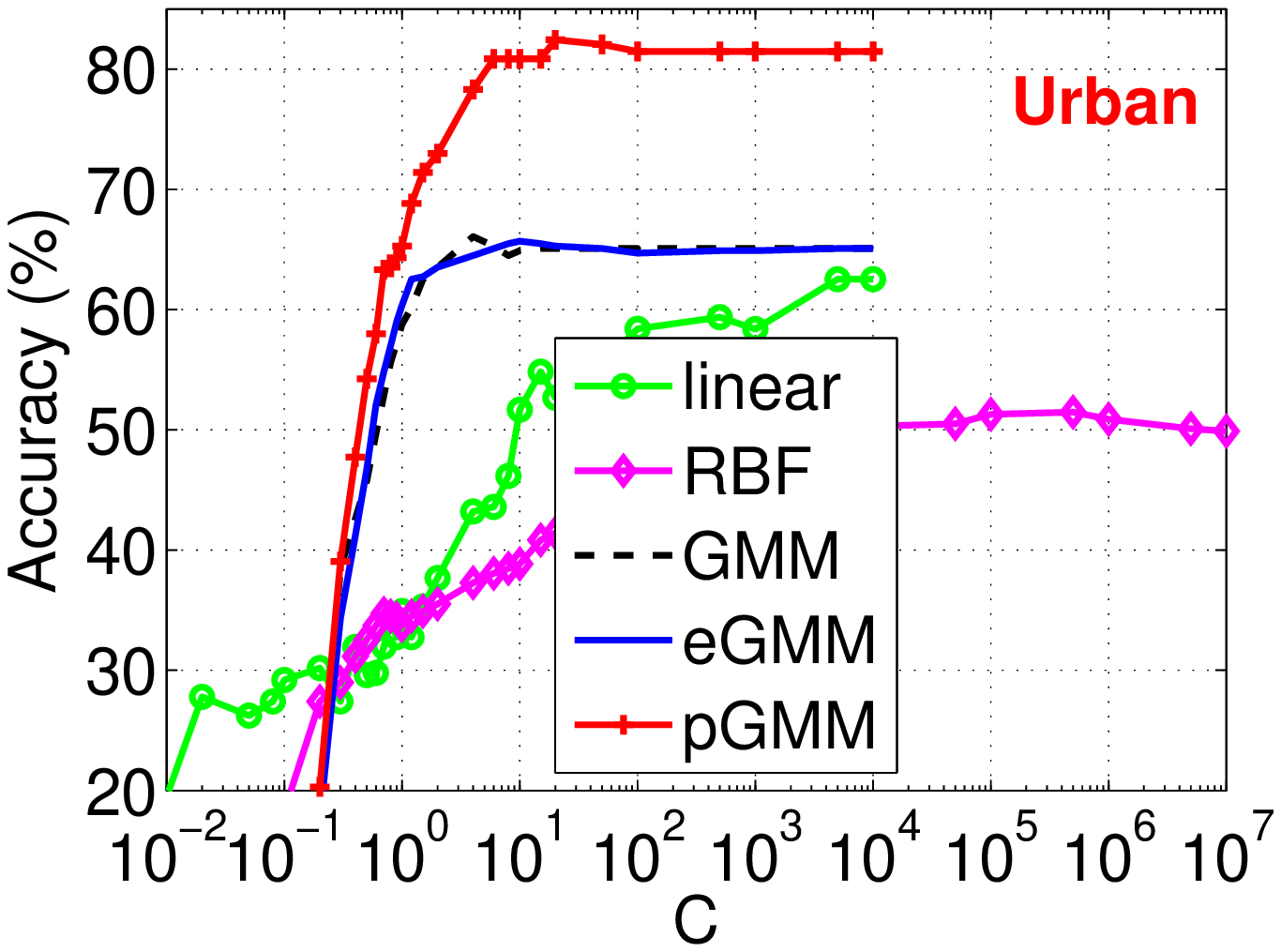}\hspace{-0.14in}
\includegraphics[width=2.3in]{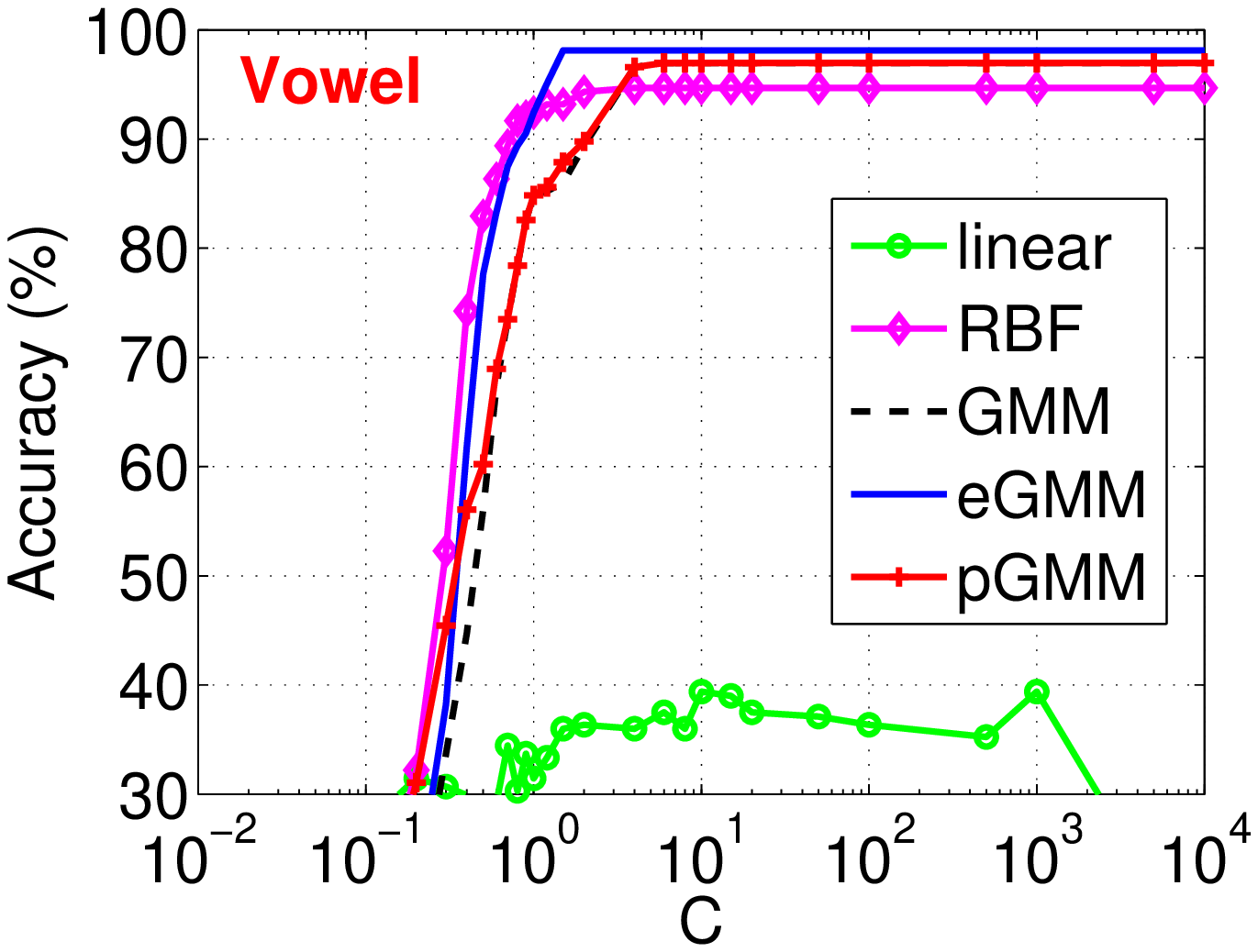}
}

\mbox{
\includegraphics[width=2.3in]{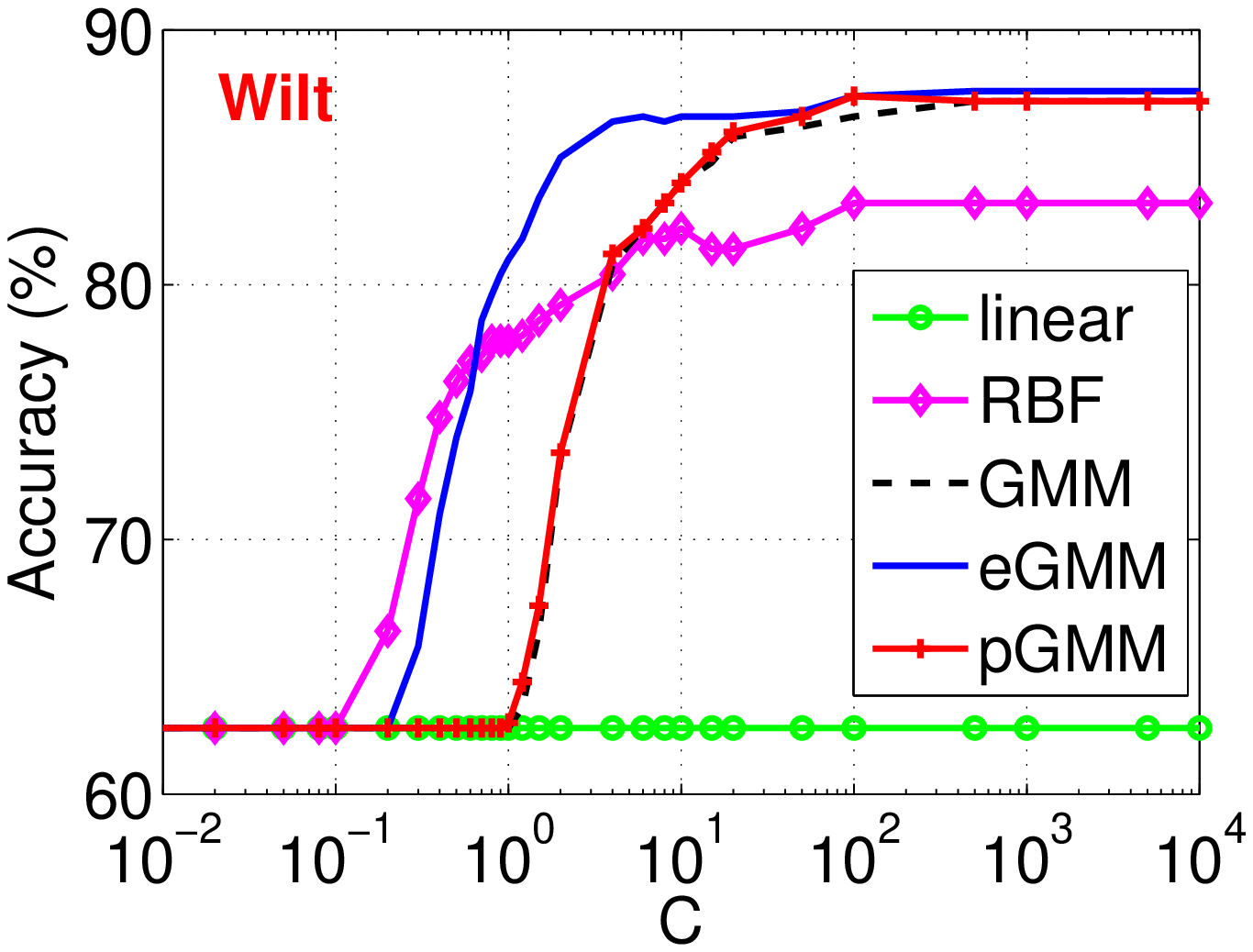}\hspace{-0.14in}
\includegraphics[width=2.3in]{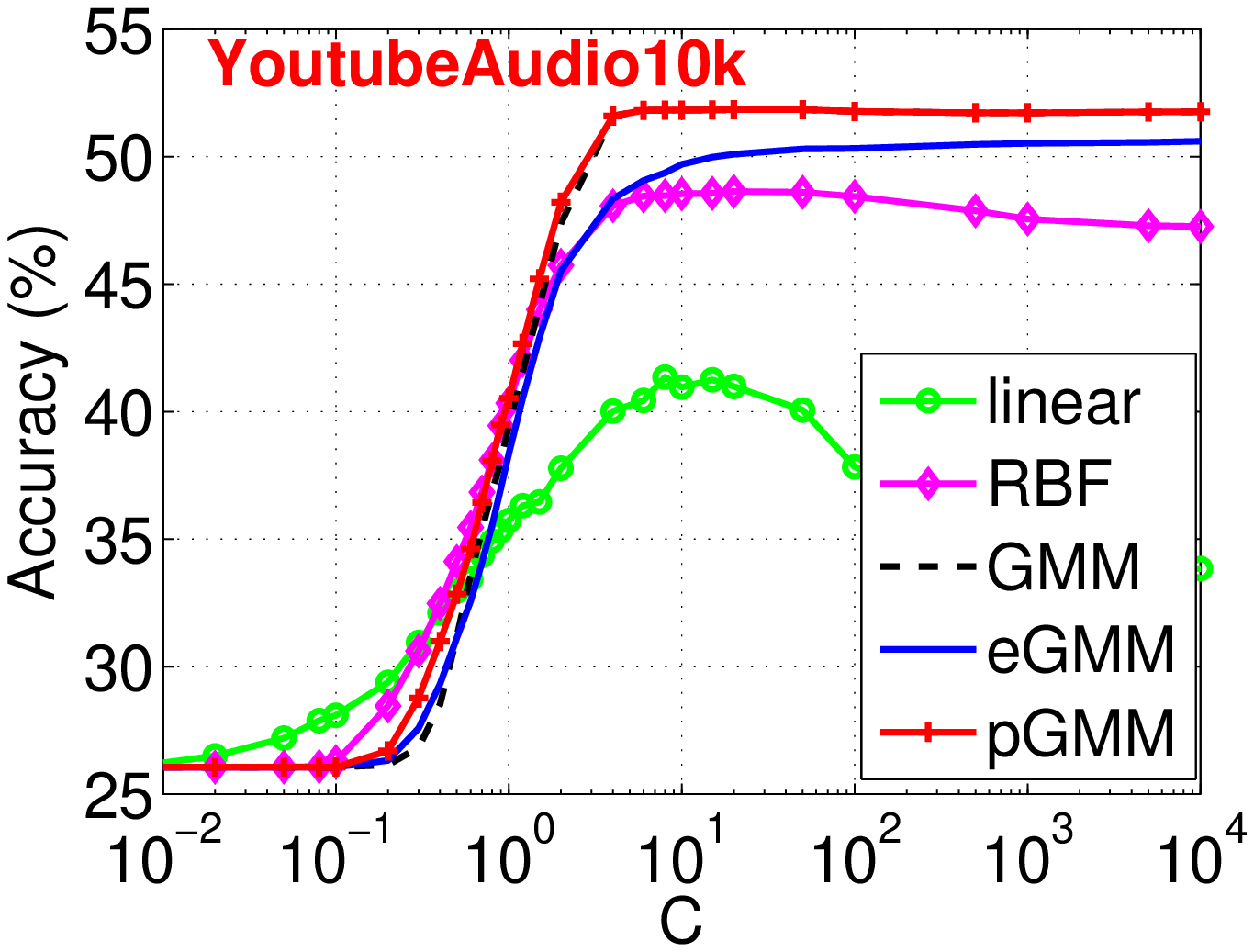}\hspace{-0.14in}
\includegraphics[width=2.3in]{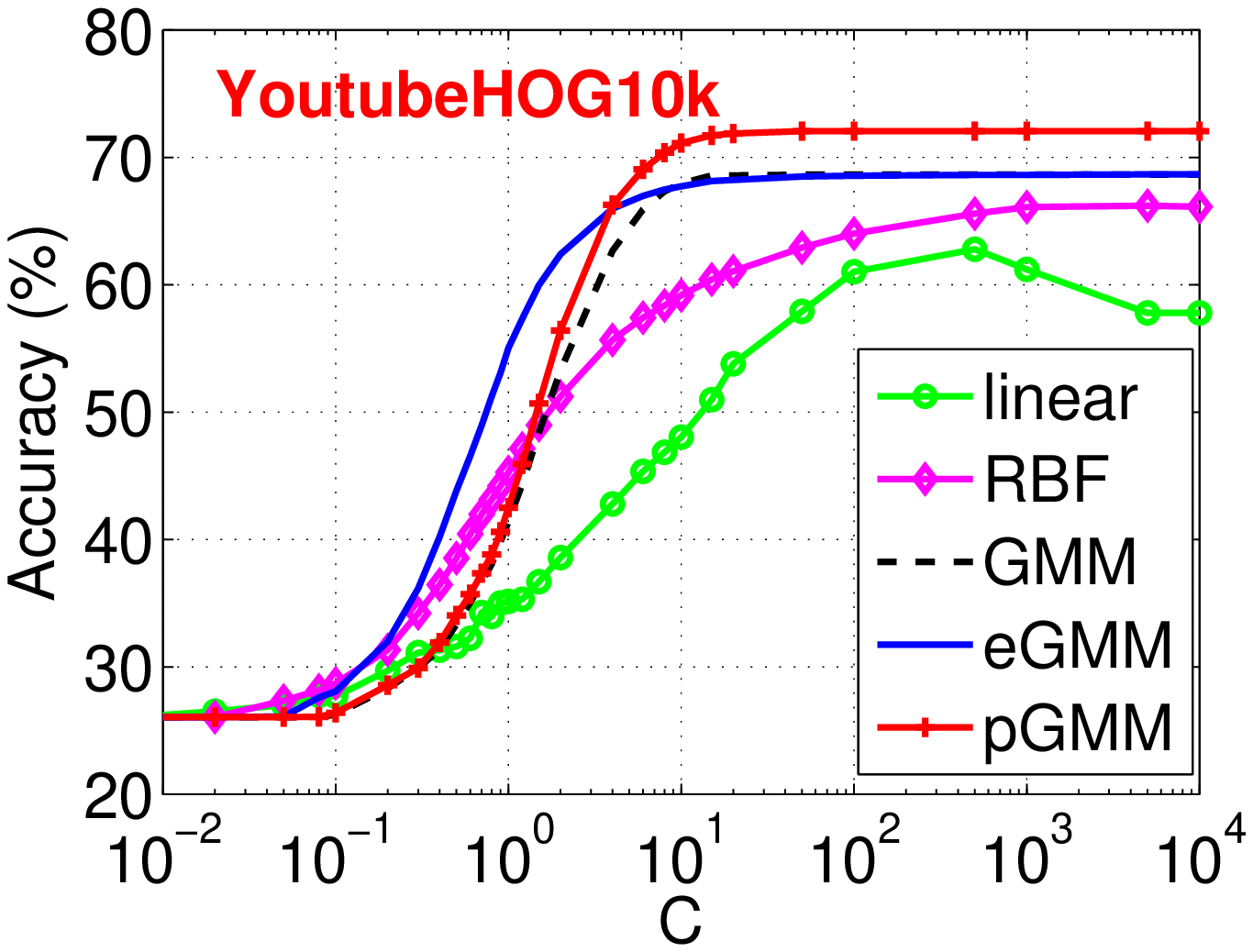}
}

\end{center}
\vspace{-0.3in}
\caption{Test classification accuracies of various kernels using LIBSVM pre-computed kernel functionality. The results are presented  with respect to $C$, which is the $l_2$-regularized kernel SVM parameter. For RBF, eGMM, and pGMM, at each $C$, we report the best test accuracies from a wide range of kernel parameter ($\gamma$) values.}\label{fig_SVM2}
\end{figure}

\begin{figure}[h!]
\begin{center}

\mbox{
\includegraphics[width=2.3in]{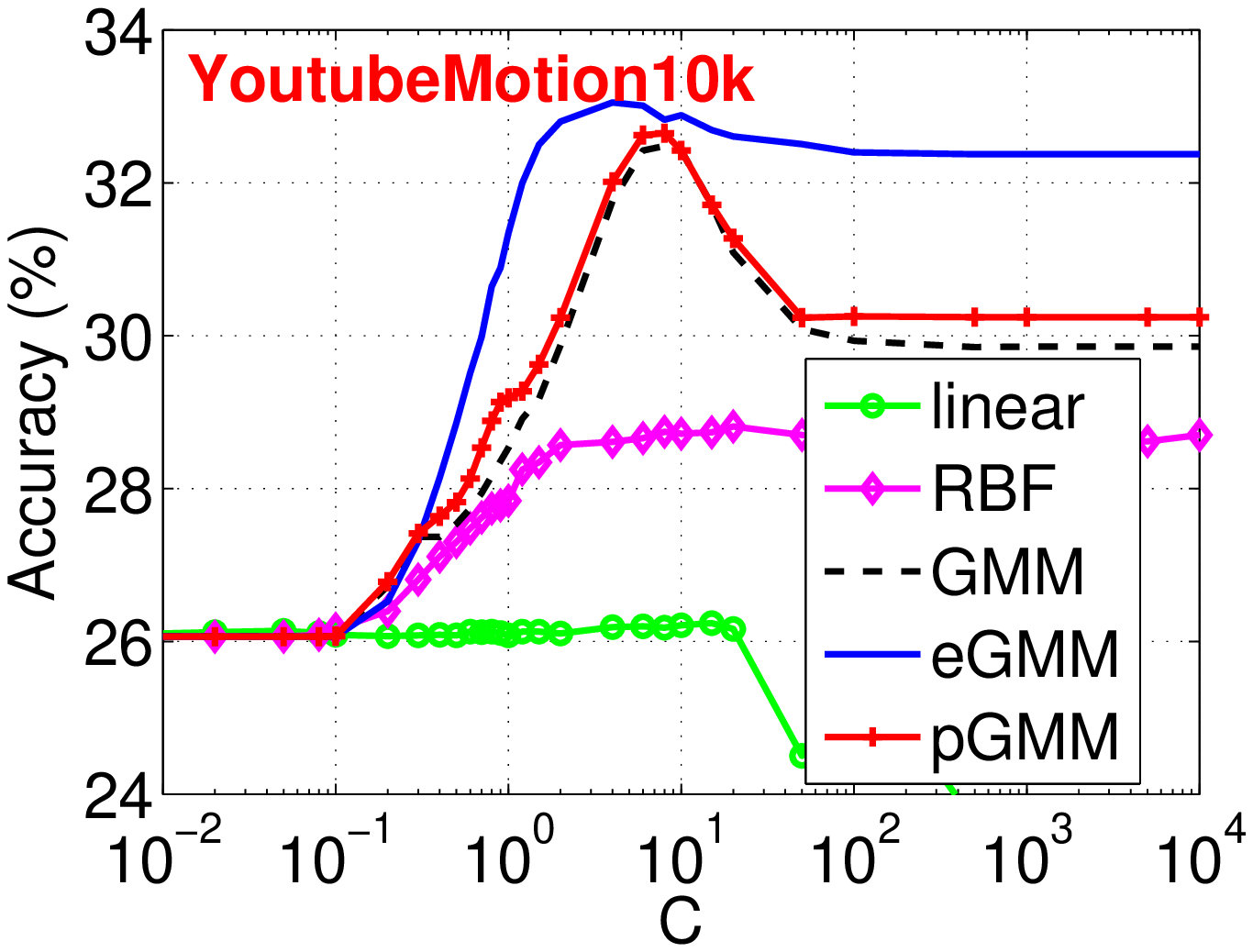}\hspace{-0.14in}
\includegraphics[width=2.3in]{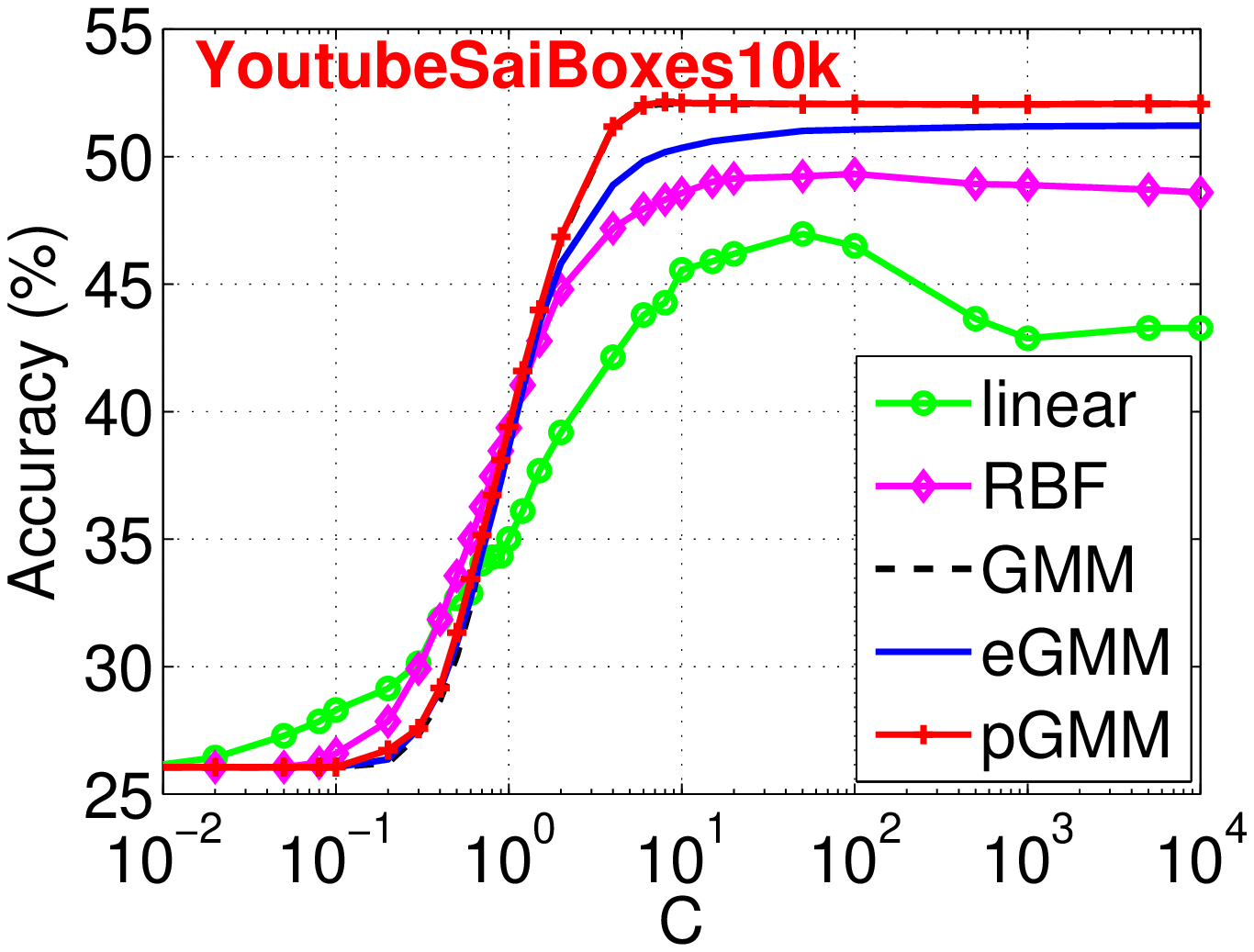}\hspace{-0.14in}
\includegraphics[width=2.3in]{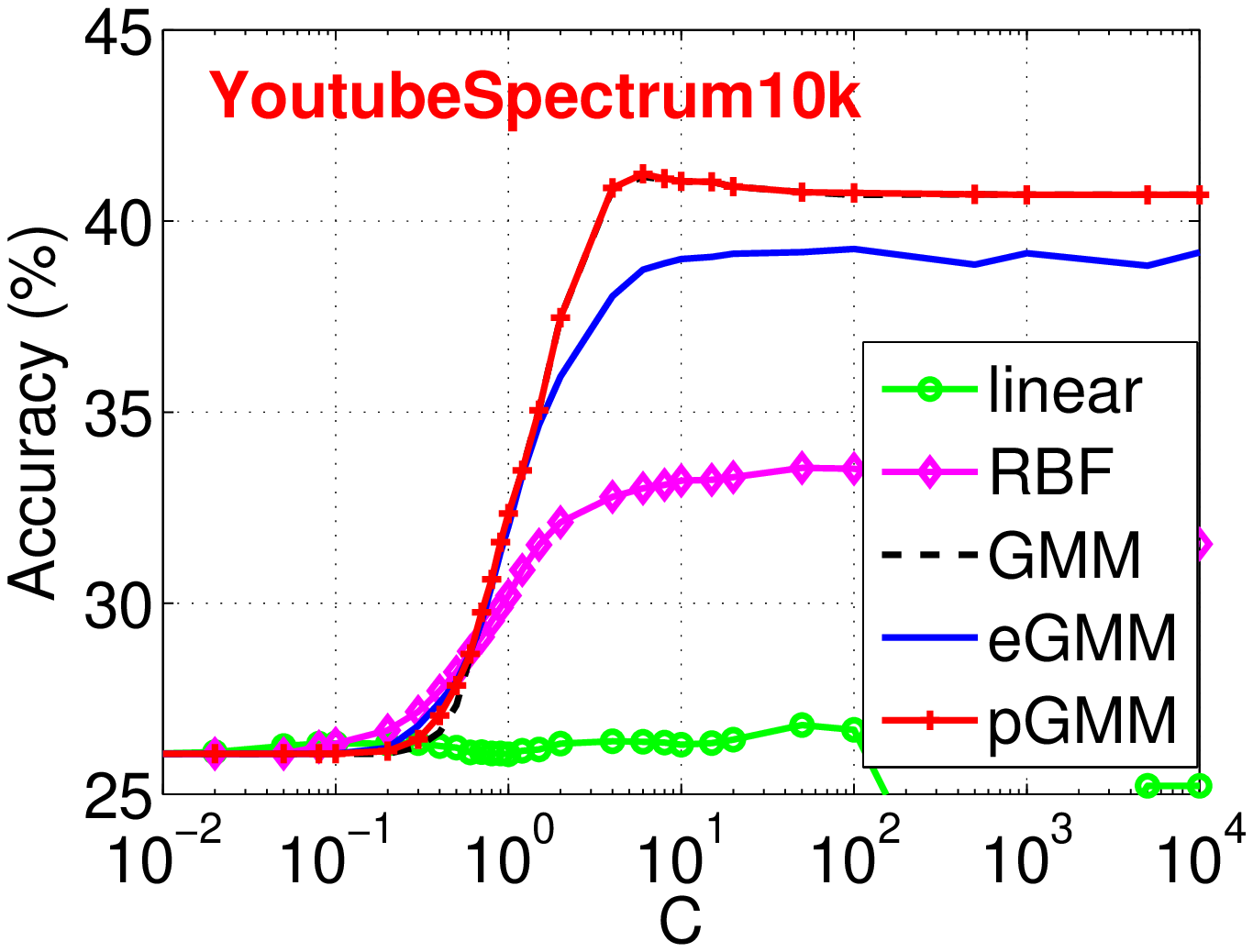}
}

\mbox{
\includegraphics[width=2.3in]{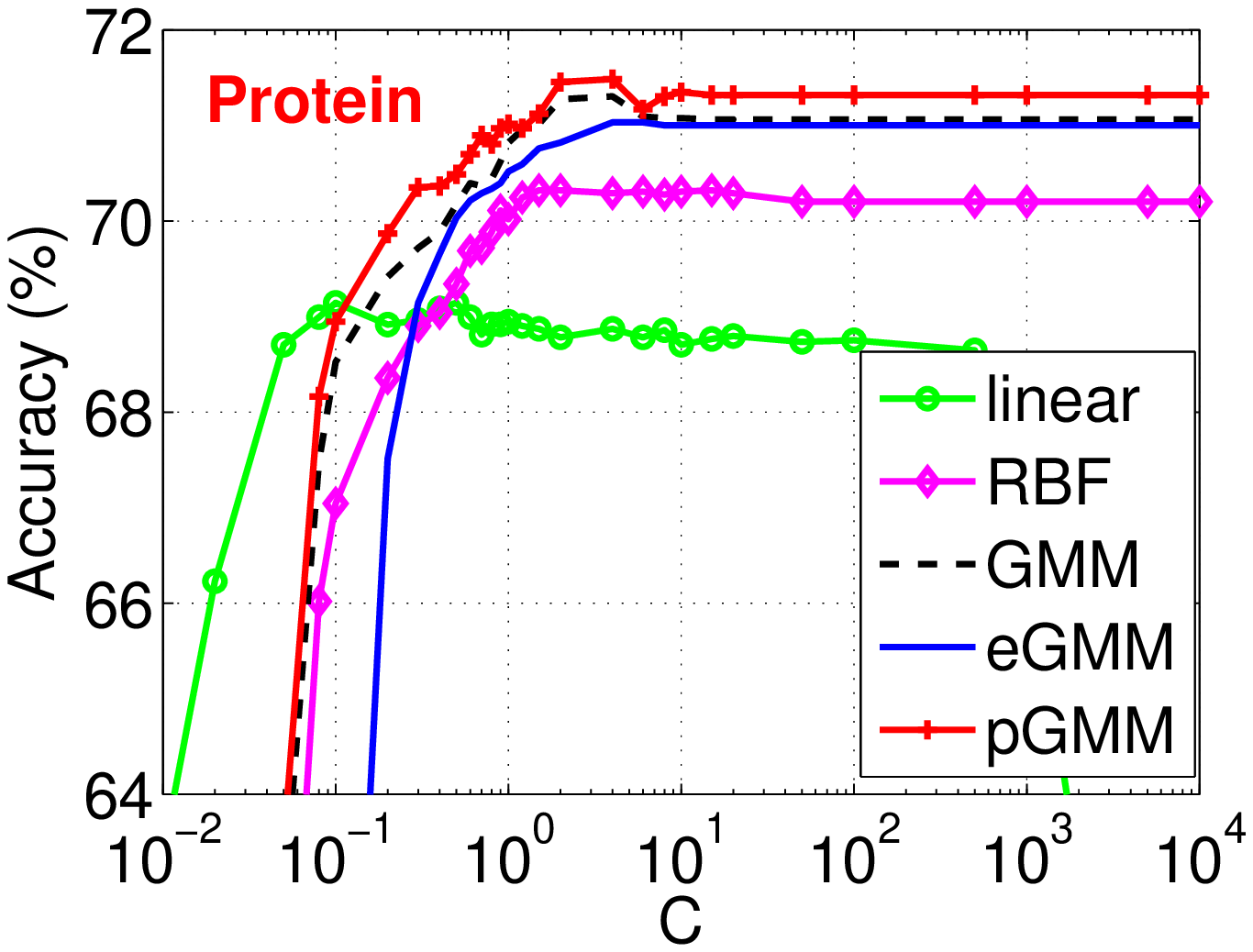}\hspace{-0.14in}
\includegraphics[width=2.3in]{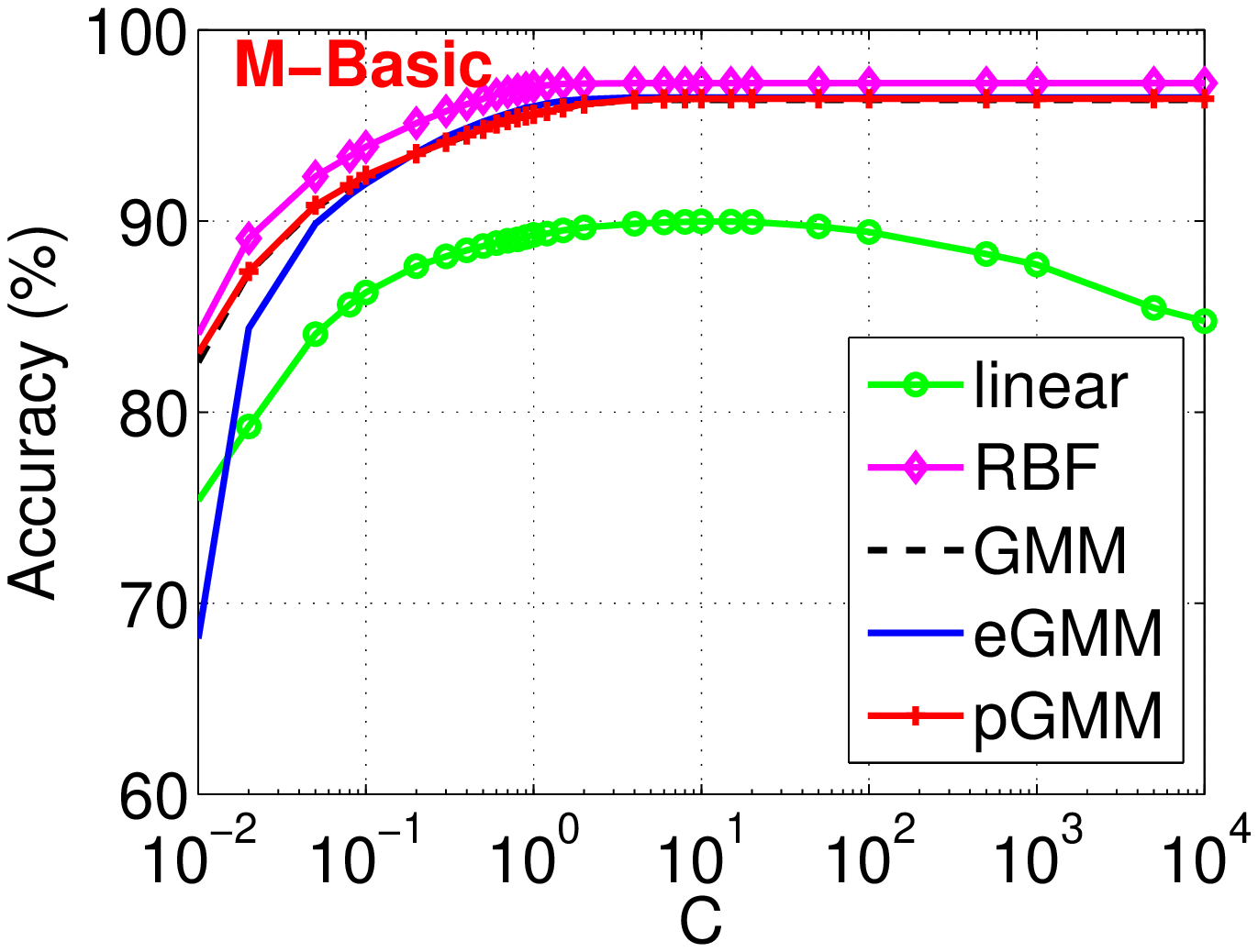}\hspace{-0.14in}
\includegraphics[width=2.3in]{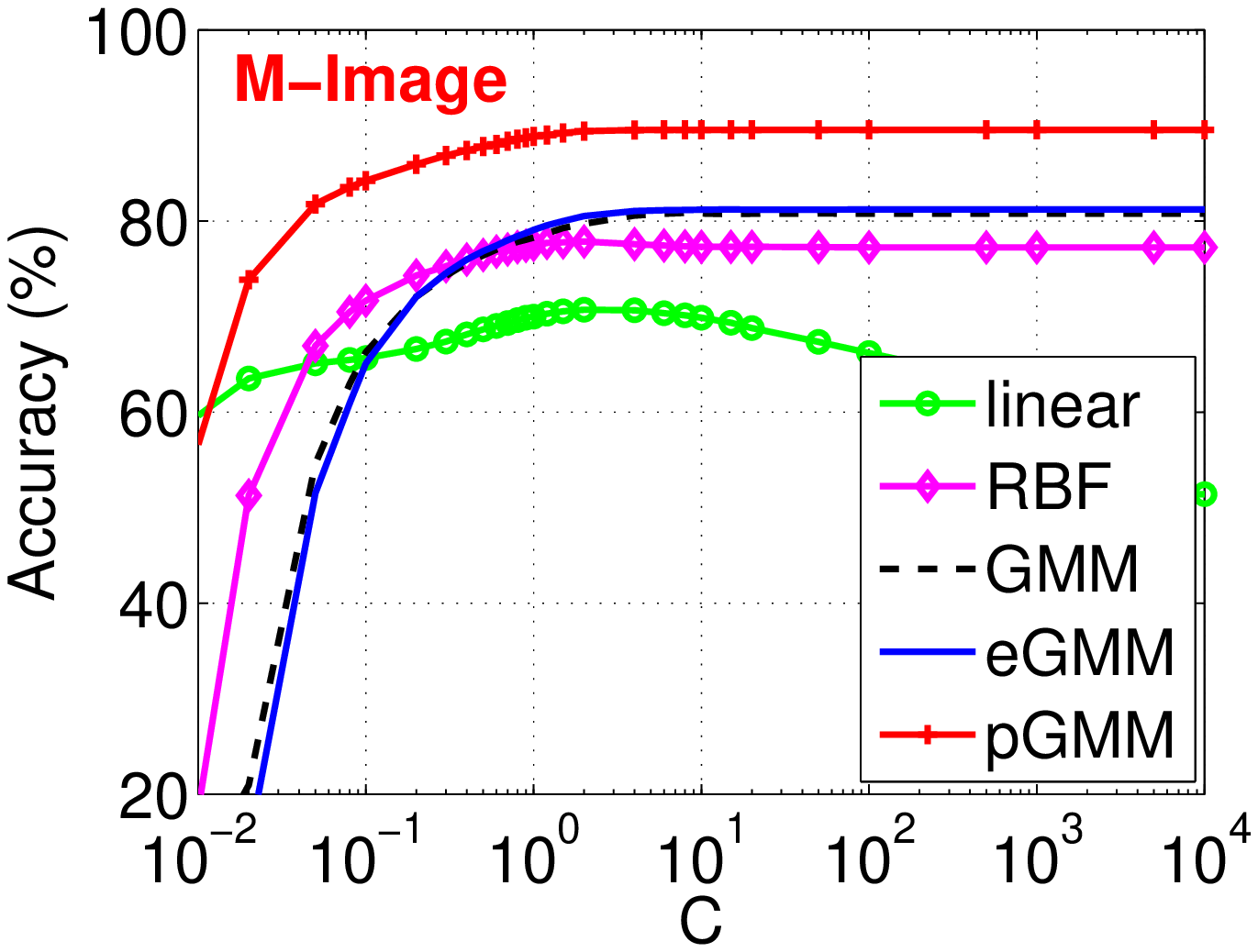}
}

\mbox{
\includegraphics[width=2.3in]{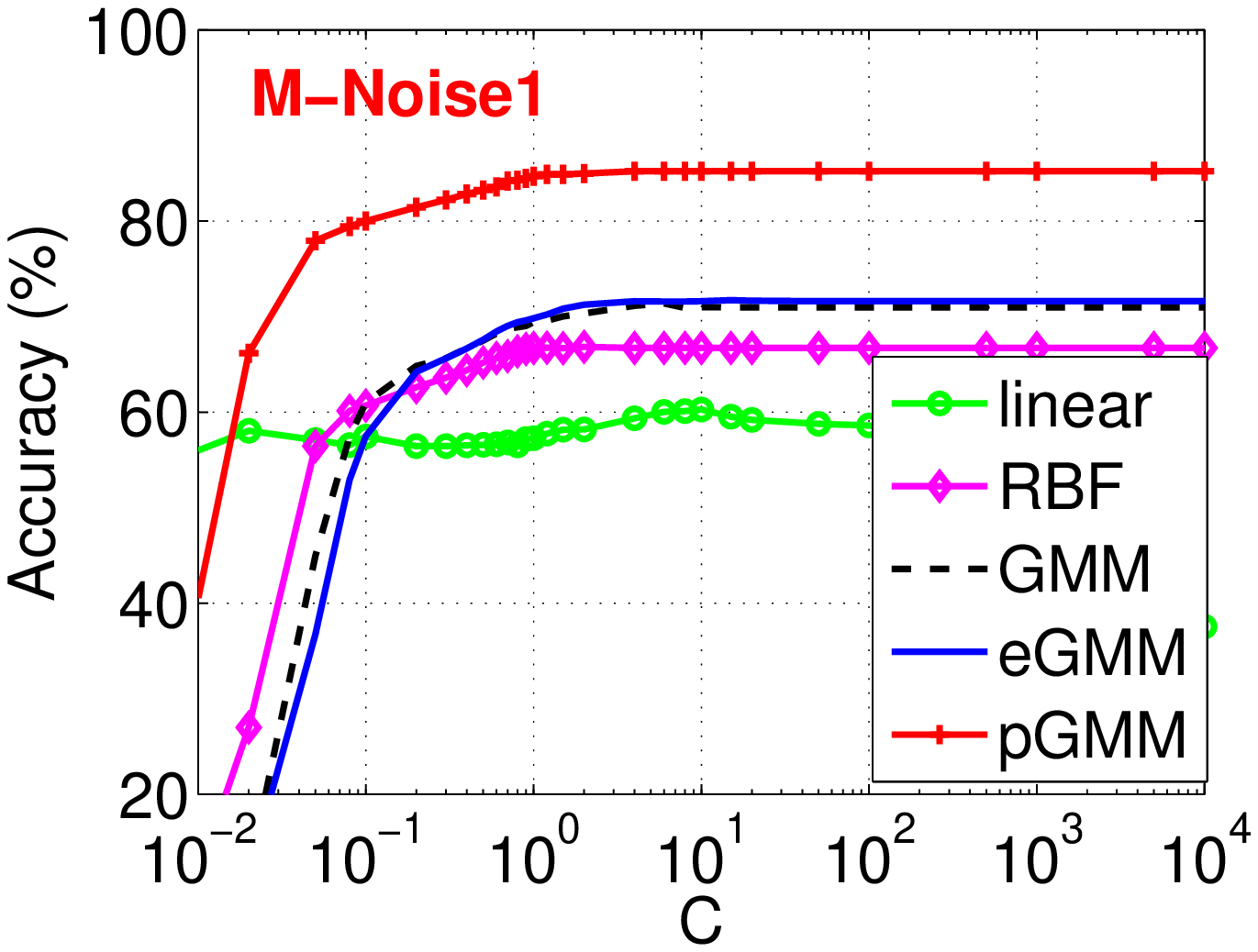}\hspace{-0.14in}
\includegraphics[width=2.3in]{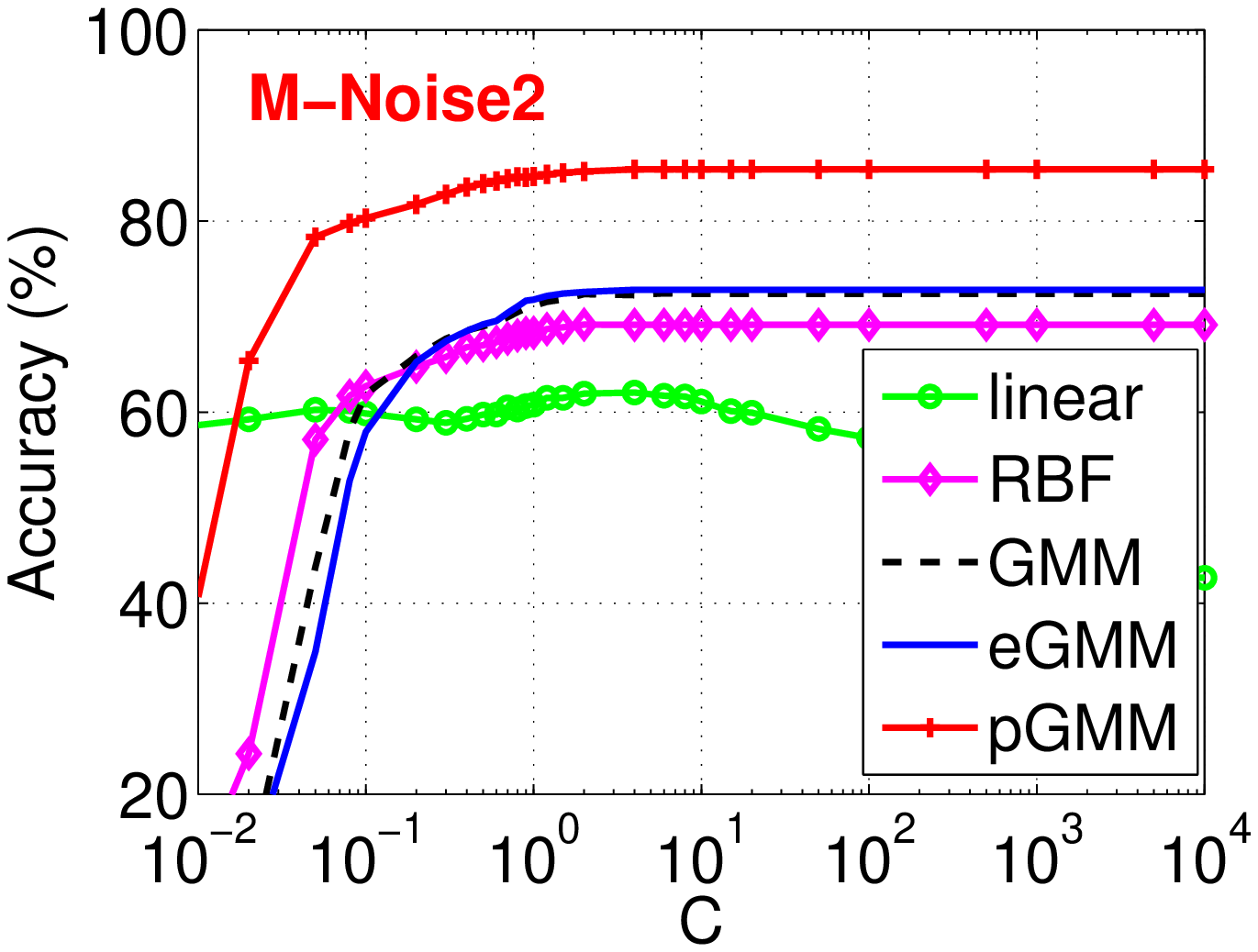}\hspace{-0.14in}
\includegraphics[width=2.3in]{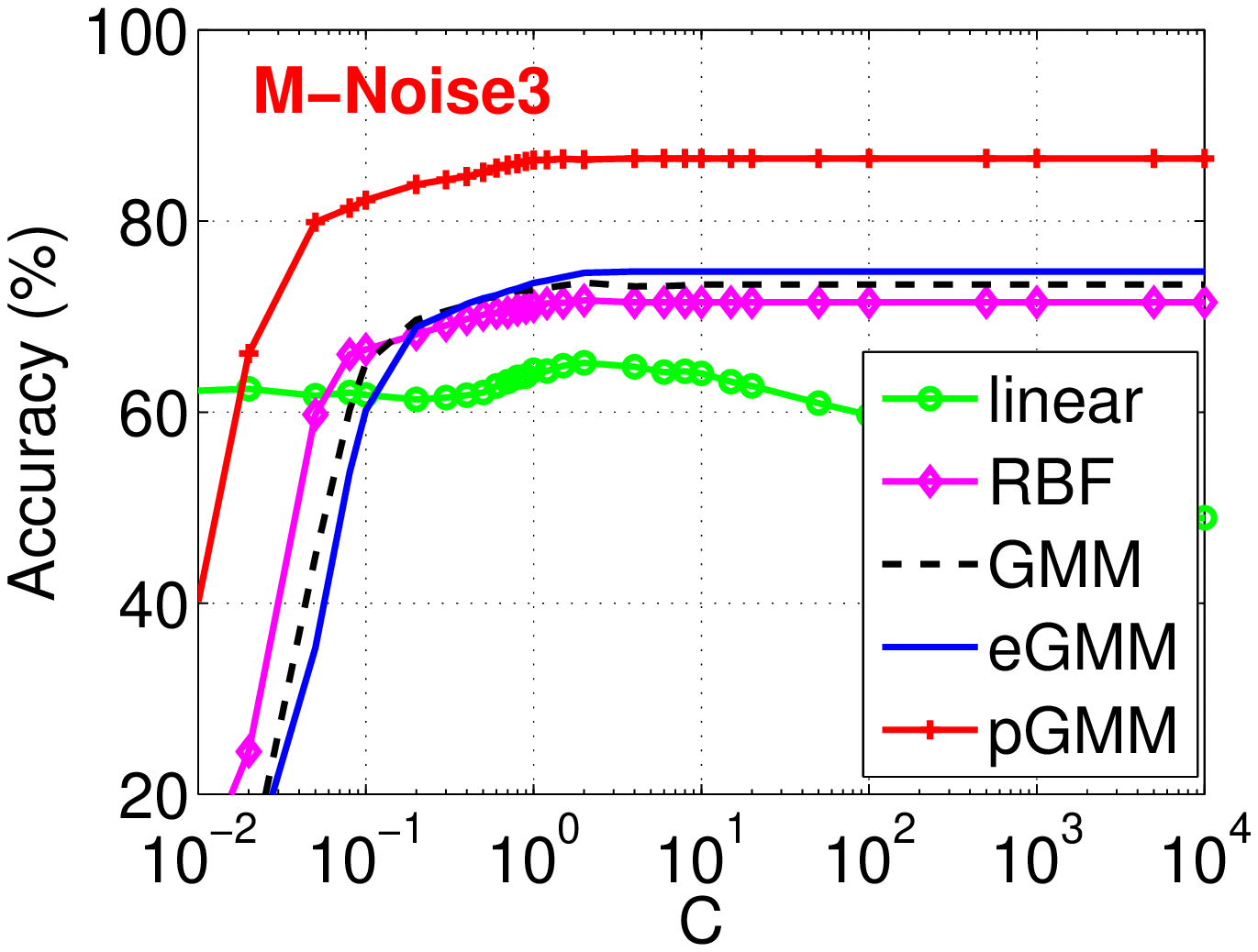}
}

\mbox{
\includegraphics[width=2.3in]{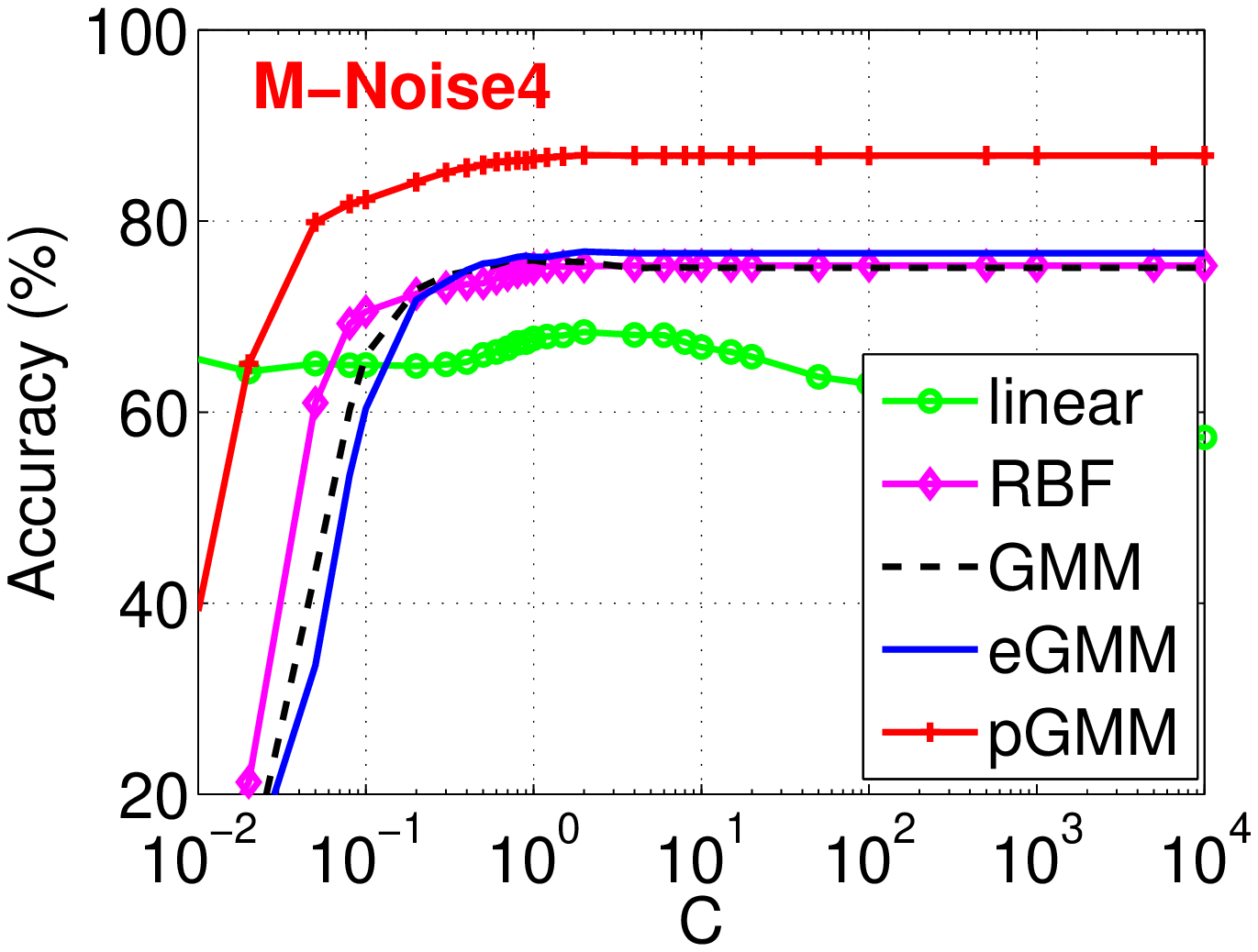}\hspace{-0.14in}
\includegraphics[width=2.3in]{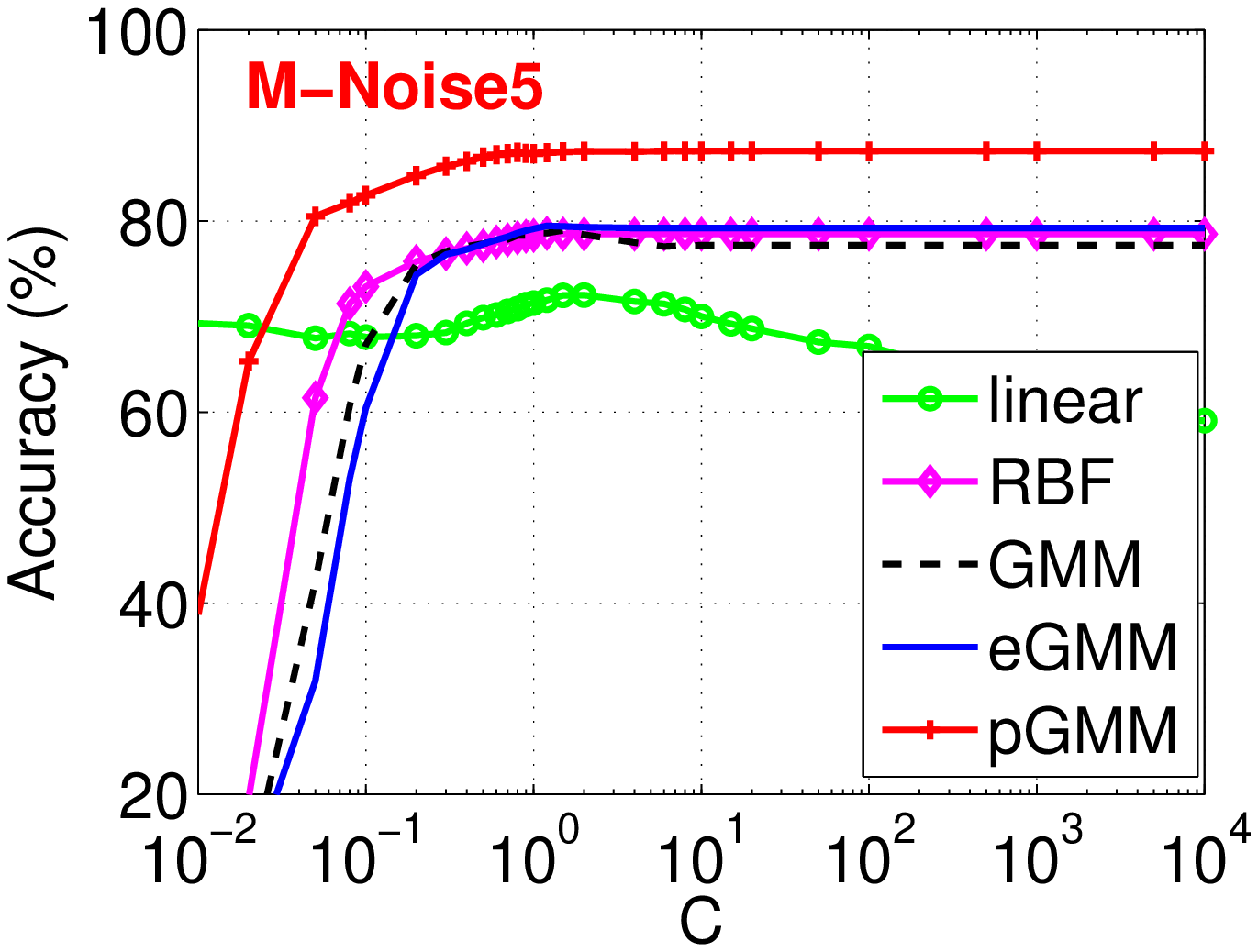}\hspace{-0.14in}
\includegraphics[width=2.3in]{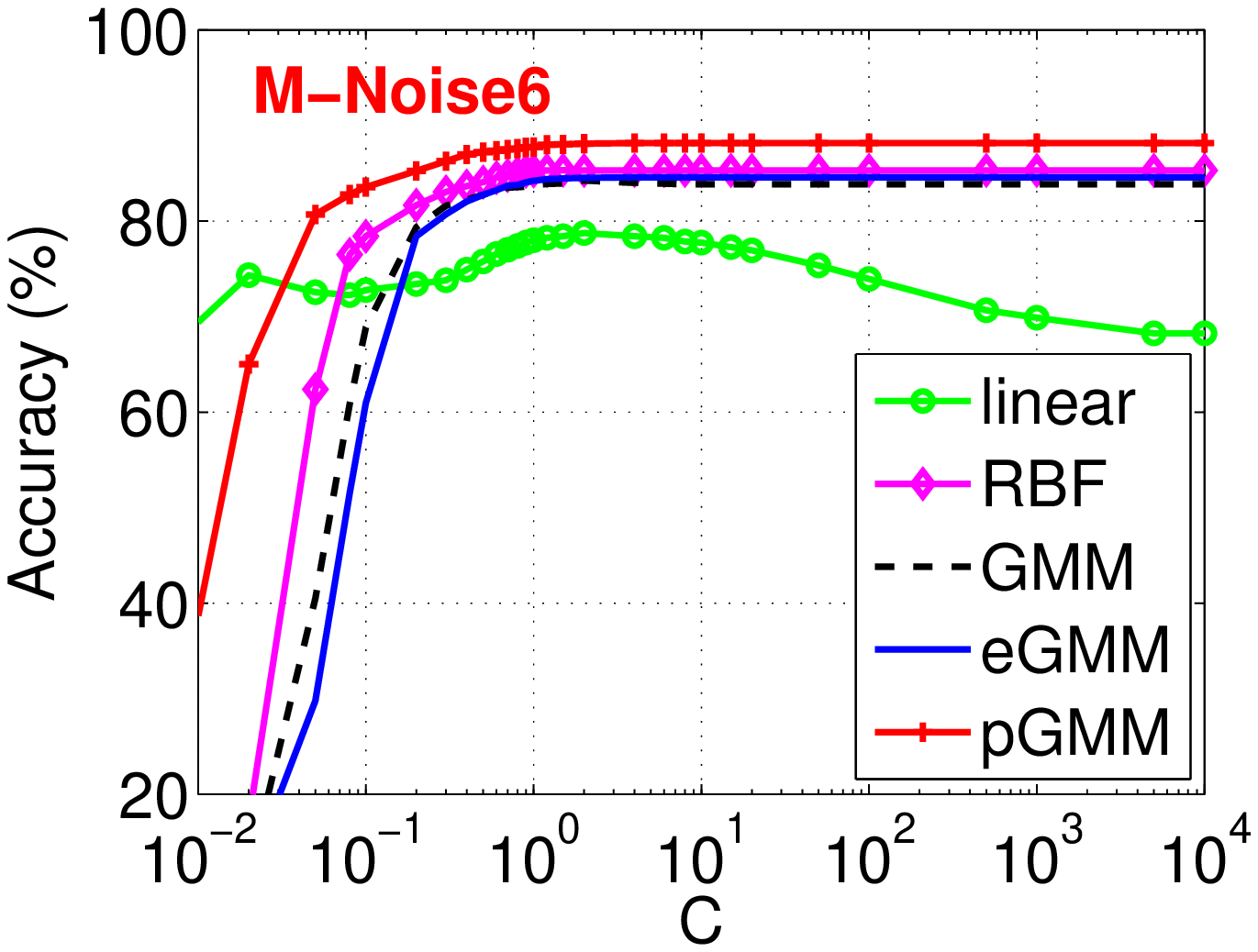}
}

\mbox{
\includegraphics[width=2.3in]{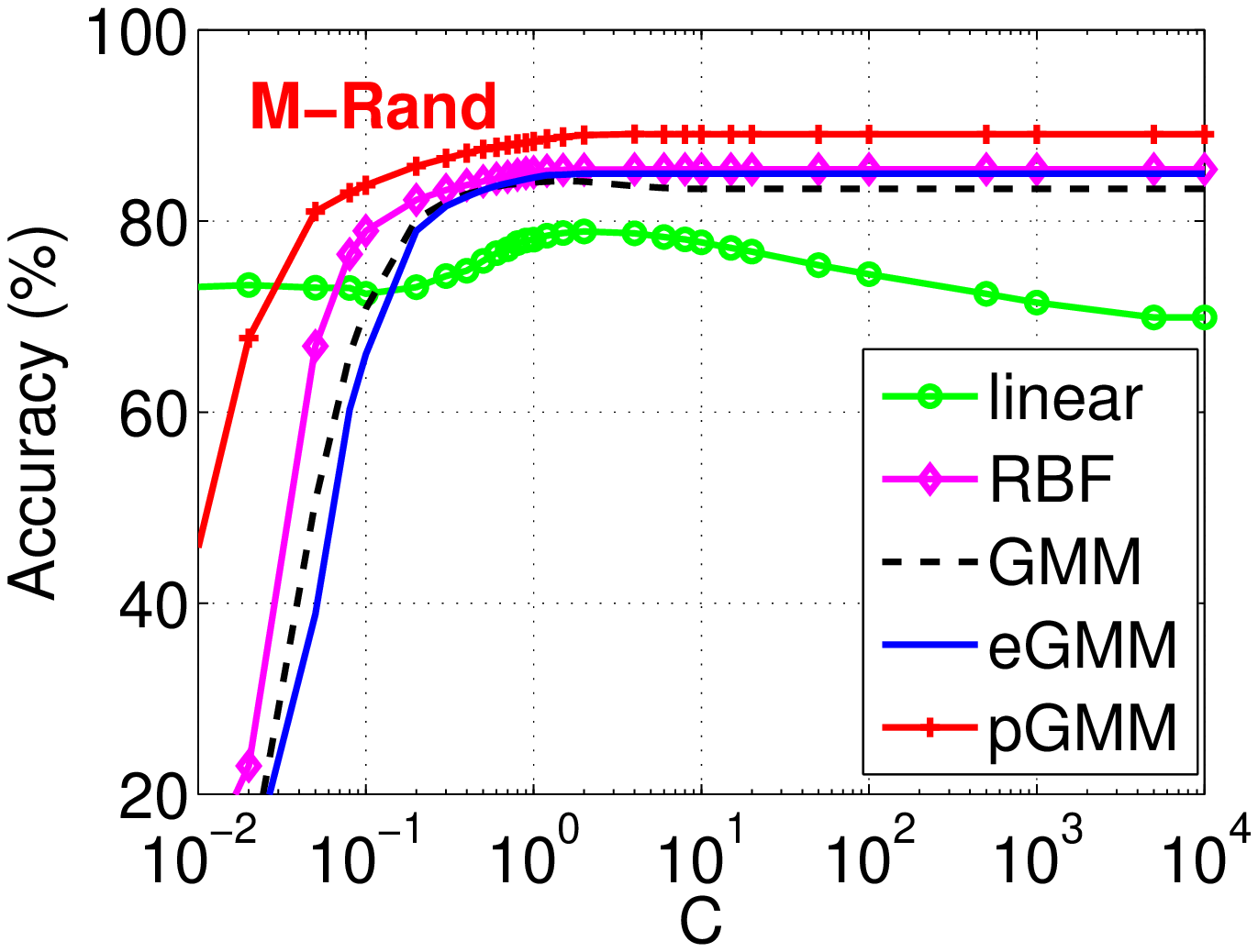}\hspace{-0.14in}
\includegraphics[width=2.3in]{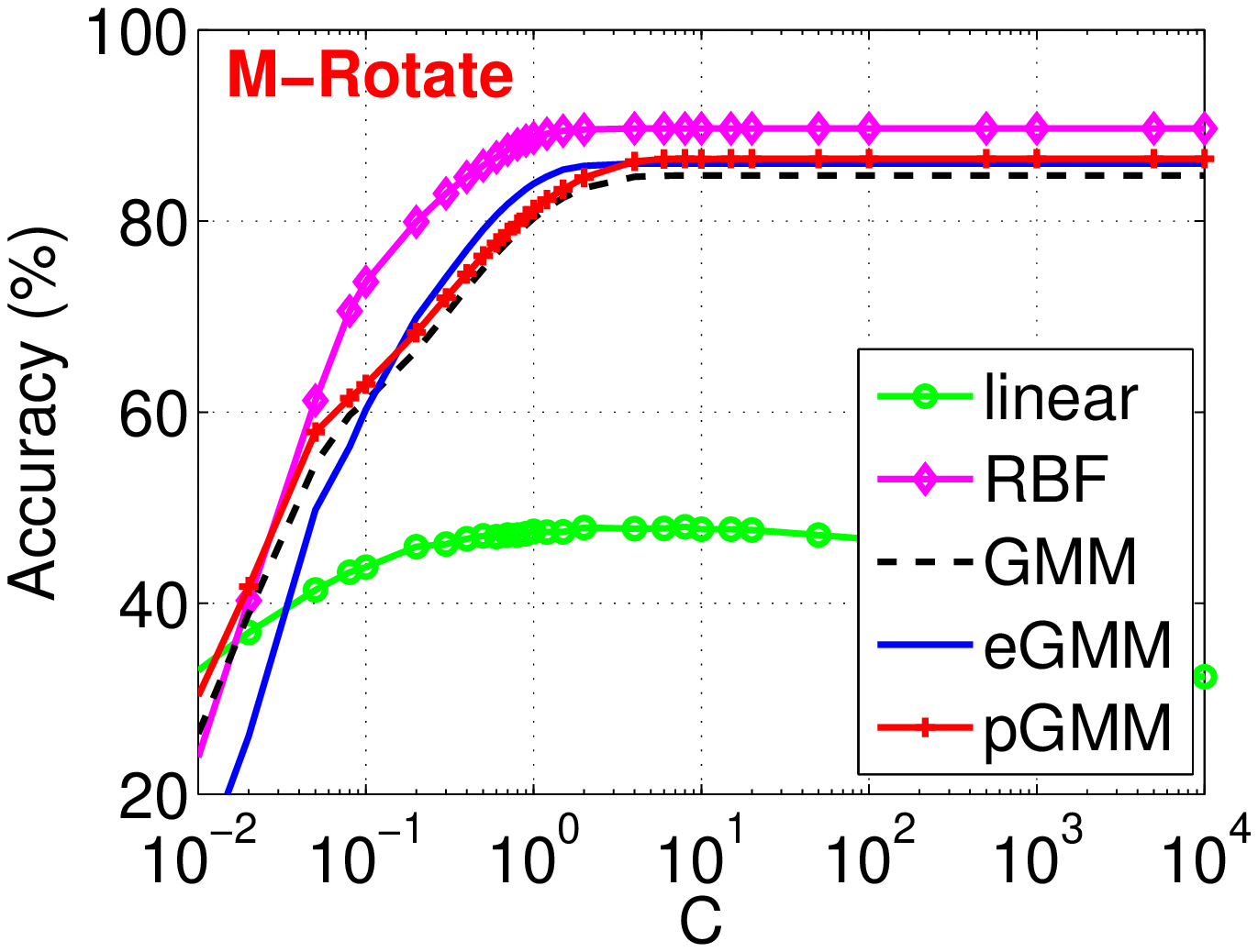}\hspace{-0.14in}
\includegraphics[width=2.3in]{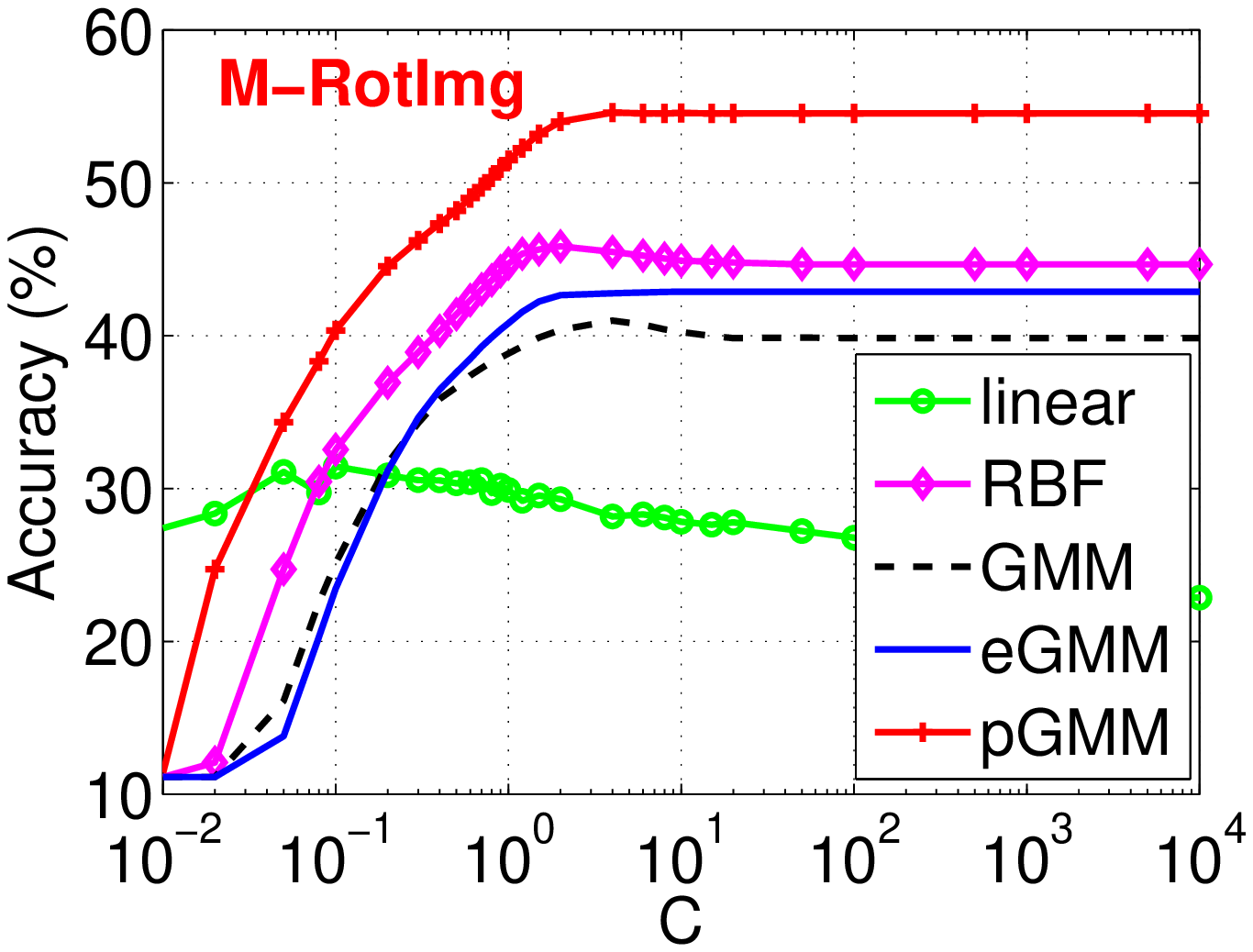}
}

\end{center}
\vspace{-0.3in}
\caption{Test classification accuracies of various kernels using LIBSVM pre-computed kernel functionality. The results are presented  with respect to $C$, which is the $l_2$-regularized kernel SVM parameter. For RBF, eGMM, and pGMM, at each $C$, we report the best test accuracies from a wide range of kernel parameter ($\gamma$) values.}\label{fig_SVM3}
\end{figure}

In all the experiments, we adopt the $l_2$-regularization (with a regularization parameter $C$) and report the test classification accuracies at the best $C$ values in Table~\ref{tab_data} and Table~\ref{tab_data2}. More detailed results for a wide range of $C$ values are reported in Figures~\ref{fig_SVM1}, \ref{fig_SVM2}, and \ref{fig_SVM3}. To ensure repeatability, we use the LIBSVM pre-computed kernel functionality, at the significant cost of disk space. For the RBF kernel, we follow~\cite{Report:Li_GMM16}, by exhaustively experimenting with 58 different values of $\gamma\in\{$0.001, 0.01, 0.1:0.1:2, 2.5, 3:1:20 25:5:50, 60:10:100, 120, 150, 200, 300, 500, 1000$\}$. Basically, Table~\ref{tab_data} and Table~\ref{tab_data2} report the best RBF results among all $\gamma$ and $C$ values in our experiments. Here, 3:1:20 is the matlab notation, meaning that the iterations stat at 3 and terminate at 20, at a space of 1. \\

 For the eGMM kernel, we experiment with the same set of (58) $\gamma$ values as for the RBF kernel. For the pGMM kernel, however, because we have to materialize (store) a kernel matrix for each $\gamma$, disk space becomes a  serious concern. Therefore, for the pGMM kernel, we only search in the range of $\gamma\in\{0.05, 0.1, 0.15, 0.2, 0.25, 0.3, 0.4 0.5, 0.6, 0.75, 1, 1.25, 1.5,  2, 5, 10, 15, 20, 25, 30:10:100\}$. In other words, the performance of the pGMM kernel (and the epGMM kernel) would be further improved if we expand the range of search or granularity of spacing.  \\

The classification results in Table~\ref{tab_data} and~\ref{tab_data2} and Figures~\ref{fig_SVM1}, ~\ref{fig_SVM2} and~\ref{fig_SVM3} confirm that the eGMM and pGMM kernels typically improve the original GMM kernel. On a good fraction of datasets, the improvements can be very significant. In fact, Section~\ref{sec_epGMM} will show that using the epGMM kernel can bring in further improvements. Nevertheless, the RBF kernel still exhibits the best performance on a very small number of datasets. This is great because it means there is still room for improvement in future study.

\begin{table}[h!]
\caption{\textbf{Public (UCI) classification datasets  and $l_2$-regularized kernel SVM results}. We report the test classification accuracies for the linear kernel, the best-tuned RBF kernel, the original (tuning-free) GMM kernel, the best-tuned eGMM kernel, and the best-tuned pGMM kernel, at their individually-best SVM regularization  $C$ values.
}
\begin{center}{\small
{\begin{tabular}{l r r r c c c c c}
\hline \hline
Dataset     &\# train  &\# test  &\# dim &linear  &RBF  &GMM &eGMM &pGMM  \\
\hline
Car &864 &864 &6 &71.53   &94.91  &98.96   &99.31    &{\bf99.54} \\
Covertype25k &25000 &25000 &54 &62.64 &{82.66} &82.65  &{\bf88.32}   &83.25 \\
CTG &1063 &1063 &35 &60.59   &89.75  &88.81   &88.81  &{\bf100.00}\\
DailySports &4560 &4560 &5625  & 77.70 &97.61 &\textbf{99.61} &\textbf{99.61}  &\textbf{99.61} \\
DailySports2k&2000&7120&5625& 72.16  &93.71   &98.99  &99.00  &{\bf99.07}  \\
Dexter &300 &300 &19999 &92.67   &93.00  &94.00   &94.00  &{\bf94.67} \\
Gesture &4937 &4936 &32 & 37.22   &61.06   &{65.50}  &{\bf66.67}   &66.33\\
ImageSeg &210 &2100 &19 &83.81   &91.38   &{95.05}   &95.38  & {\bf95.57}\\
Isolet2k &2000 &5797 &617 & 93.95   &{\bf95.55} &95.53 &{\bf95.55}&95.53\\
MHealth20k&20000&20000&23&72.62   &82.65   &{85.28} &85.33   &{\bf86.69}  \\
MiniBooNE20k&20000&20000&50&88.42   &93.06   &{93.00}  &93.01  &{\bf93.72} \\
MSD20k &20000 &20000 &90 &66.72 &68.07 &{71.05}  &71.18   &{\bf71.84}\\
Magic &9150 &9150 &10 &78.04   &84.43  &{87.02} &86.93   &{\bf87.57}  \\
Musk &3299 &3299 &166  & 95.09 &\textbf{99.33} &99.24 &99.24 &99.24 \\
Musk2k&2000&4598&166&94.80   &97.63   &{98.02}   &{98.02}   &{\bf98.06} \\
PageBlocks &2737  &2726 &10 &95.87   &{97.08}   &96.56   &{96.56} &{\bf97.33} \\
Parkinson &520&520&26&61.15   &66.73   &{69.81}   &{\bf70.19} &69.81\\
PAMAP101 &20000 &20000 &51 &76.86   &96.68  &{98.91}  &98.91 &{\bf99.00} \\
PAMAP102 &20000 &20000 &51 &81.22   &95.67  &{\bf98.78}  &98.77 &{\bf98.78}\\
PAMAP103 &20000 &20000 &51 & 85.54  &97.89  &{99.69}  &{\bf99.70} &99.69\\
PAMAP104 &20000 &20000 &51 &84.03  &97.32   &{99.30}  &{\bf99.31} &99.30\\
PAMAP105 &20000 &20000 &51 &79.43  &97.34   &{99.22} &{\bf99.24} &99.22 \\
RobotNavi &2728 &2728 &24 &69.83   &90.69   &{96.85}   &96.77  &{\bf98.20} \\
Satimage &4435 &2000 &36 &72.45   &85.20   &{90.40}   &{\bf91.85}   &90.95\\
SEMG1 &900 &900 &3000  &26.00   &\textbf{43.56}   &41.00 &41.22 &42.89 \\
SEMG2 &1800 &1800 &2500  &19.28  &29.00    &{54.00} &54.00 &{\bf56.11} \\
Sensorless &29255 &29254 &48 &61.53 &93.01  &{99.39} &99.38 &{\bf99.76}  \\
Shuttle500 &500 &14500 &9 &91.81   &99.52  &{99.65}   &99.65  &{\bf99.66} \\
SkinSeg10k&10000&10000&3& 93.36   &99.74 &{99.81} &{\bf99.90}   &99.85   \\
SpamBase &2301&2300&57&   85.91&   92.57 & {94.17}   &94.13 &{\bf95.78} \\
Splice &1000&2175&60&85.10   &90.02   &{95.22}   &{\bf96.46}    &95.26 \\
Theorem &3059&3059&51&   67.83   &70.48   &{71.53}   &{\bf71.69}   &71.53 \\
Thyroid &3772 &3428 &21 &95.48   &97.67   &{98.31}   &98.34   &{\bf99.10}\\
Thyroid2k &2000&5200&21&   94.90   &97.00  &{98.40}   &98.40   &{\bf98.96} \\
Urban &168&507&147&   62.52   &51.48  &{66.08}   & 65.68   &{\bf83.04} \\
Vertebral &155&155&6&   80.65   &83.23 &{89.04}   &{\bf89.68}   &89.04 \\
Vowel &264 &264 &10 &39.39 &94.70  &{96.97}  &{\bf98.11}   &96.97 \\
Wholesale &220&220&6&   89.55   &90.91   &{93.18}   &93.18   &{\bf93.64} \\
Wilt &4339&500&5&62.60   &83.20   &{87.20}  &{\bf87.60} & 87.40 \\
YoutubeAudio10k&10000&11930&2000&   41.35   &48.63   &50.59 &50.60  &{\bf51.84}   \\ 
YoutubeHOG10k&10000&11930&647&   62.77   &66.20   &{68.63}  &68.65  &{\bf72.06} \\
YoutubeMotion10k&10000&11930&64& 26.24  &28.81   &{31.95} &{\bf33.05} &32.65 \\
YoutubeSaiBoxes10k&10000&11930&7168&46.97   &49.31   &51.28  &51.22   &{\bf52.15} \\
YoutubeSpectrum10k&10000&11930&1024&26.81   &33.54   &{39.23}  &39.27   &{\bf41.23} \\
\hline\hline
\end{tabular}}
}
\end{center}\label{tab_data}

\end{table}

\begin{table}[h!]
\caption{Datasets in group 1 are from the LIBSVM website. Datasets in group 2 were used by~\cite{Proc:Larochelle_ICML07,Proc:ABC_UAI10} for testing deep learning algorithms and tree methods.
}
\begin{center}{\small
{\begin{tabular}{c l r r r c c c c c}
\hline \hline
Group &Dataset     &\# train  &\# test  &\# dim &linear  &RBF   &GMM  &eGMM &pGMM \\
\hline
&Letter&15000 &5000 &16 &61.66 &{97.44}  &97.26 &{\bf97.68 }  &97.32 \\
1&Protein&17766    & 6621      & 357  &69.14   &70.32    &{70.64} &71.03   &{\bf71.48} \\
&SensIT20k&20000 &19705&100& 80.42   &83.15  &{84.57}  &84.69   &{\bf84.90} \\
&Webspam20k&20000 &60000 &254&93.00  &{97.99} &97.88 &{\bf98.21}   &97.93 \\\hline
&M-Basic   &12000 &50000 &784 & 89.98   &{\bf97.21}   &96.34  &96.47  & 96.40  \\
&M-Image &12000 &50000 &784 & 70.71 &77.84  &{80.85} &81.20 &\textbf{89.53} \\
&M-Noise1 &10000 &4000 &784 &60.28   &66.83    &{71.38} &71.70   &\textbf{85.20} \\
&M-Noise2 &10000 &4000 &784 & 62.05  & 69.15   &{72.43} &72.80   &\textbf{85.40} \\
&M-Noise3 &10000 &4000 &784 &65.15   &71.68   &{73.55} &74.70   &\textbf{86.55} \\
2&M-Noise4 &10000 &4000 &784 & 68.38  &75.33   &{76.05} &76.80  &\textbf{86.88}\\
&M-Noise5 &10000 &4000 &784 &72.25   &78.70  &{79.03} &79.48   &\textbf{87.33}\\
&M-Noise6 &10000 &4000 &784 &78.73   &{85.33} &84.23  &84.58  &\textbf{88.15}\\
&M-Rand &12000 &50000 &784 & 78.90   &{85.39}   &84.22 &84.95   &\textbf{89.09}\\
&M-Rotate &12000 &50000   &784 &47.99  &\textbf{89.68}  & 84.76 &86.02 &{86.52}\\
&M-RotImg &12000 &50000 &784 &31.44  &{45.84}   & 40.98 &42.88   &\textbf{54.58} \\
\hline\hline
\end{tabular}}
}
\end{center}\label{tab_data2}
\end{table}

\newpage\clearpage

\newpage\clearpage

\section{The epGMM Kernel, Comparisons with Deep Nets and Trees}\label{sec_epGMM}

Given two data vectors $u$ and $v$, the epGMM kernel is defined as
\begin{align}\notag
epGMM(u,v;\gamma_1,\gamma_2) =  e^{-\gamma_2\left(1-\frac{\sum_{i=1}^{2D}\left(\min\{\tilde{u}_i,\tilde{v}_i\}\right)^{\gamma_1}}{\sum_{i=1}^{2D}\left( \max\{\tilde{u}_i,\tilde{v}_i\}\right)^{\gamma_1}}\right)}
\end{align}
after applying the transformation in (\ref{eqn_transform}) to $u$ and $v$. When $\gamma_1=1$, this becomes the eGMM kernel. \\

In our experiments with the pGMM kernel, we searched for the best $\gamma$ (i.e., the $\gamma_1$ here) parameter in the range of $\gamma\in\{0.05, 0.1, 0.15, 0.2, 0.25, 0.3, 0.4 0.5, 0.6, 0.75, 1, 1.25, 1.5,  2, 5, 10, 15, 20, 25, 30:10:100\}$. Note that since we have to store a kernel matrix at each $\gamma$, the experiments are costly. For testing the epGMM kernel, we  re-use the
those pre-computed kernels and experiment with the epGMM kernel using the same $\gamma$ values (which is the $\gamma_2$ here) as for the RBF and eGMM kernels.\\

The experimental results are reported in Table~\ref{tab_epGMM} (the last column).  We can see that the epGMM kernel indeed improves over the eGMM and pGMM kernels, as one would have expected. The improvements can be quite noticeable on those datasets.

\begin{table}[h!]
\caption{We add the results (test classification accuracies) for the epGMM kernel as the last column. We mainly focus on the datasets in group 1, for the purpose of comparing with deep nets and trees.}
\begin{center}{\small
{\begin{tabular}{c l r r r c c c c c c}
\hline \hline
Group &Dataset     &\# train  &\# test  &\# dim &linear  &RBF  &GMM &eGMM &pGMM&epGMM\\
\hline
&M-Basic   &12000 &50000 &784 & 89.98   &{\bf97.21}   &96.34  &96.47  & 96.40  &{96.71}  \\
&M-Image &12000 &50000 &784 & 70.71 &77.84  &{80.85} &81.20 &89.53  &\textbf{89.96}\\
&M-Noise1 &10000 &4000 &784 &60.28   &66.83    &{71.38} &71.70   &85.20 &\textbf{85.58}  \\
&M-Noise2 &10000 &4000 &784 & 62.05  & 69.15   &{72.43} &72.80   &85.40  &\textbf{86.05} \\
&M-Noise3 &10000 &4000 &784 &65.15   &71.68   &{73.55} &74.70   &86.55 &\textbf{87.10} \\
1&M-Noise4 &10000 &4000 &784 & 68.38  &75.33   &{76.05} &76.80  &86.88 &\textbf{87.43} \\
&M-Noise5 &10000 &4000 &784 &72.25   &78.70  &{79.03} &79.48   &87.33  &\textbf{88.30} \\
&M-Noise6 &10000 &4000 &784 &78.73   &{85.33} &84.23  &84.58  &88.15 &\textbf{88.85} \\
&M-Rand &12000 &50000 &784 & 78.90   &{85.39}   &84.22 &84.95   &89.09  &\textbf{89.43} \\
&M-Rotate &12000 &50000   &784 &47.99  &\textbf{89.68}  & 84.76 &86.02 &86.56  &{88.36} \\
&M-RotImg &12000 &50000 &784 &31.44  &{45.84}   & 40.98 &42.88   &54.58  &\textbf{55.73}\\\hline
&Protein&17766    & 6621      & 357  &69.14   &70.32    &{70.64} &71.03   &{71.48} &\textbf{71.97}\\
&Webspam20k&20000 &60000 &254&93.00  &{97.99} &97.88 &{98.21}   &97.93 &\textbf{98.49} \\
2&Covertype25k &25000 &25000 &54 &62.64 &{82.66} &82.65  &{88.32}   &83.14 &\textbf{88.77} \\
&Gesture &4937 &4936 &32 & 37.22   &61.06   &{65.50}  &{66.67}   &66.33 & \textbf{68.09}\\
&YoutubeMotion10k&10000&11930&64& 26.24  &28.81   &{31.95} &{33.05} &32.65  &\textbf{34.79}\\
\hline\hline
\end{tabular}}
}
\end{center}\label{tab_epGMM}

\end{table}

\begin{figure}[h!]
\begin{center}
\includegraphics[width=3in]{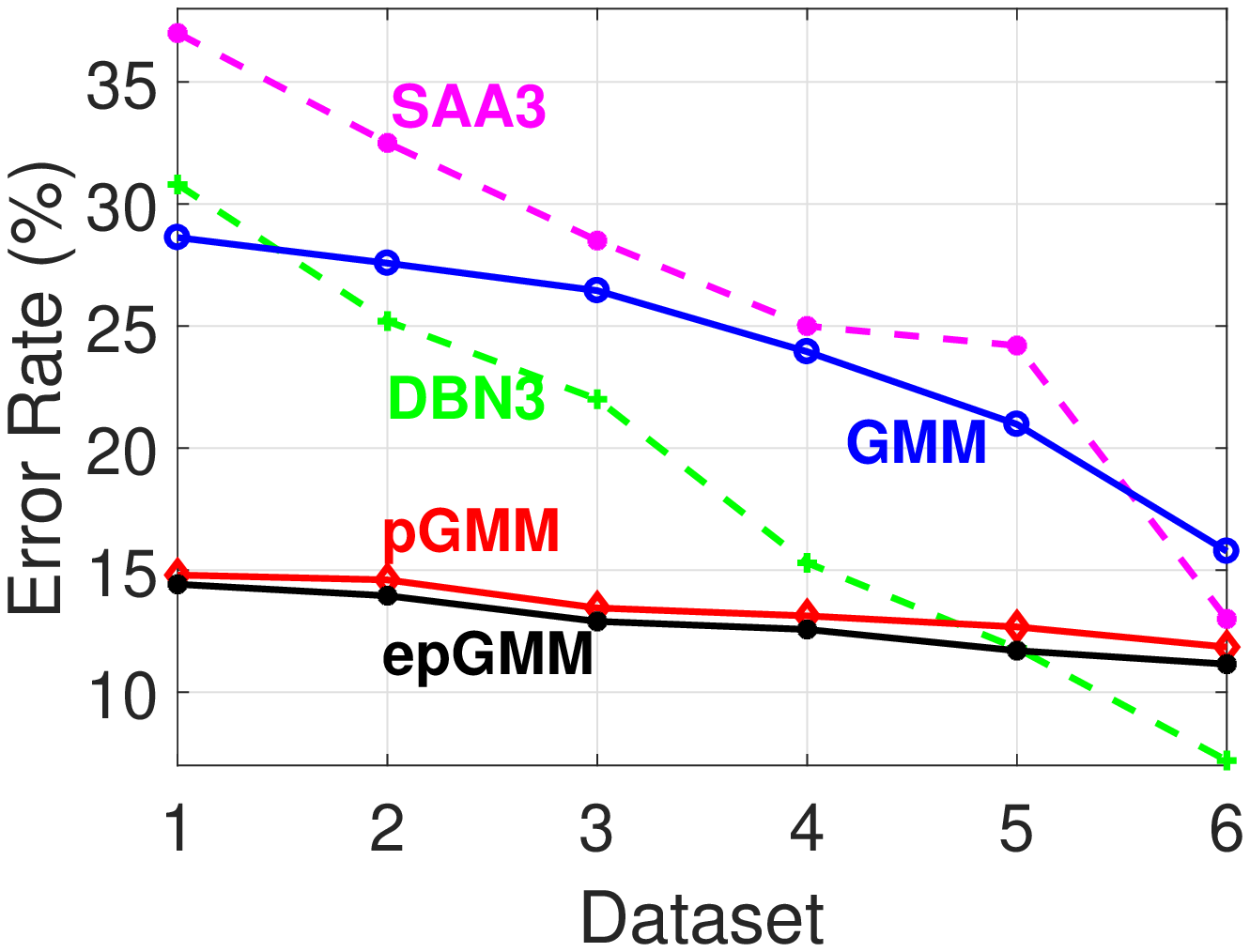}
\includegraphics[width=3in]{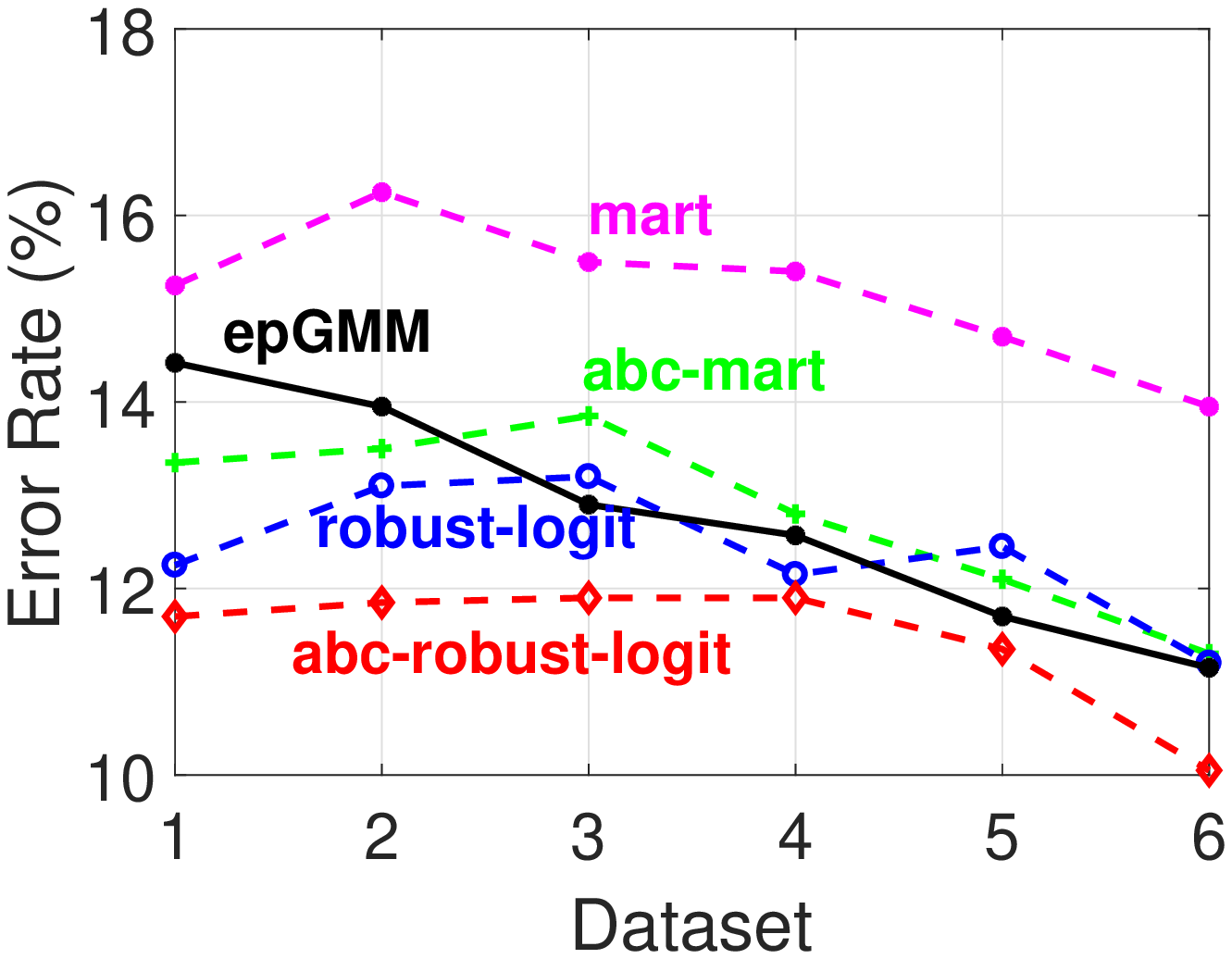}

\end{center}
\vspace{-0.3in}
\caption{Classification test error rates on M-Noise1, M-Noise2, ..., M-Noise6 datasets. The left panel compares GMM, pGMM, and epGMM with two deep learning algorithms as reported in~\cite{Proc:Larochelle_ICML07}. The right panel compares epGMM with four boosted tree methods as reported in~\cite{Proc:ABC_UAI10}.}\label{fig_Noise6}
\end{figure}

The 11 datasets in Group 1 of Table~\ref{tab_epGMM} were already used for testing deep learning algorithms and tree methods~\cite{Proc:Larochelle_ICML07,Proc:ABC_UAI10}. It is perhaps surprising that the performance of the pGMM kernel (and the epGMM kernel) can be largely comparable to deep nets and boosted trees, as shown in Figure~\ref{fig_Noise6} and Table~\ref{tab_deep}. These results are exciting, because, that this point, we merely use kernel SVM with single kernels. It is  reasonable to expect that additional improvements  might be achieved in future studies.

\newpage\clearpage

In their  studies, \cite{Article:Li_ABC_arXiv08,Proc:ABC_ICML09,Proc:ABC_UAI10} developed tree methods including ``abc-mart'', ``robust logitboost'', and ``abc-robust-logitboost'' and demonstrated their excellent performance on those 11 datasets (and other datasets), by establishing the second-order tree-split formula and new derivatives for multi-class logistic loss function. They always used a special histogram-based implementation named ``adaptive binning'', and the ``best-first'' strategy for determining the region for the next split (thus, the trees were not balanced as they did not directly control the levels of depth.).\\

Figure~\ref{fig_Noise6} reports the test classification error rates (lower is better) for six datasets: M-Noise1, M-Noise2, ..., M-Noise6. In the left panel, we plot the results of the GMM kernel, the eGMM kernel, and the epGMM kernel, together with the results of two deep learning algorithms as reported in~\cite{Proc:Larochelle_ICML07}. We can see that for most of those six datasets, the pGMM kernel and the epGMM kernel achieve the best accuracy. In the right panel of Figure~\ref{fig_Noise6}, we compare epGMM with four boosted tree methods: mart, abc-mart, robust logitboost, and abc-robust-logitboost. \\

The ``mart'' tree algorithm~\cite{Article:Friedman_AS01} has been  popular in industry practice, especially in search. At each boosting step, it uses the first derivative of the logistic loss function as the residual response to fit regression trees, to achieve excellent robustness and fairly good accuracy. The earlier work on ``logitboost''~\cite{Article:FHT_AS00} were believed to exhibit numerical issues (which in part motivated the development of mart). It turns out that the numerical issue does not actually exist after~\cite{Proc:ABC_UAI10} derived the tree-split formula using both the first and second order derivatives of the logistic loss function. ~\cite{Proc:ABC_UAI10} showed the ``robust logitboost'' in general improves ``mart'', as can be seen from Figure~\ref{fig_Noise6} (right panel).

\cite{Article:Li_ABC_arXiv08,Proc:ABC_ICML09,Proc:ABC_UAI10} made an interesting (and perhaps brave) observation that the derivatives (as in text books) of the classical logistic loss function can be written in a different form for the multi-class case, by enforcing the ``sum-to-zero'' constraints.  At each boosting step, they identify a ``base class'' either by the ``worst-class'' criterion~\cite{Article:Li_ABC_arXiv08} or the exhaustive search method as reported in~\cite{Proc:ABC_ICML09,Proc:ABC_UAI10}. This ``adaptive base class (abc)'' strategy can be combined with either mart or robust logitboost; hence the names ``abc-mart'' and ``abc-robust-logitboost''. The improvements due to the use of ``abc'' strategy can also be substantial. Again, as mentioned earlier, in all the tree implementations, they~\cite{Article:Li_ABC_arXiv08,Proc:ABC_ICML09,Proc:ABC_UAI10} always used the adaptive-binning strategy for simplifying the implementation and speeding up training. Also, they followed the ``best-first'' criterion whereas  many tree implementations used balanced trees (which may cause ``data-imbalance'' and reduce accuracy).

\begin{table}[h]
\caption{Test error rates of five additional datasets reported in~\cite{Proc:Larochelle_ICML07,Proc:ABC_UAI10}. The results in group 1 are from~\cite{Proc:Larochelle_ICML07}, where they compared kernel SVM, neural nets, and deep learning. The results in group 3 are  from~\cite{Proc:ABC_UAI10}, which compared four boosted tree methods with deep nets.}
\begin{center}{
\begin{tabular}{c l c c c c c c}
\hline \hline
Group &Method &M-Basic & M-Rotate &M-Image &M-Rand & M-RotImg\\\hline
&SVM-RBF &{\bf3.05}\% &11.11\% &22.61\% &14.58\% &55.18\%\\
&SVM-POLY &3.69\% &15.42\% &24.01\% &16.62\% &56.41\%\\
1&NNET &4.69\% &18.11\% &27.41\% &20.04\% &62.16\%\\
&DBN-3 &3.11\% &{\bf10.30\%} &16.31\% &{\bf6.73\%} &47.39\%\\
&SAA-3 &3.46\% &{\bf10.30\%} &23.00\% &11.28\% &51.93\%\\
&DBN-1 &3.94\% &14.69\% &16.15\% &9.80\% &52.21\%\\\hline
&Linear &10.02\% &52.01\% &29.29\% &21.10\% &68.56\% \\
&RBF &2.79\% &{\bf10.30}\% &22.16\% &14.61\% &54.16\%\\
2&GMM &3.80\% &15.24\% &19.15\% &15.78\% &59.02\%\\
&eGMM &3.53\%&13.98\% &18.80\%&15.05\% &57.12\%\\
&pGMM &3.63\%&13.44\% &10.47\%&10.91\% &45.42\%\\
&epGMM &3.29\%&11.81\%&10.04\%&10.57\%&{\bf44.27}\%\\\hline
&{mart} & 4.12\% &15.35\% &11.64\% & 13.15\% &49.82\%\\
3&{abc-mart} &3.69\% &13.27\% &9.45\% & 10.60\% &46.14\%\\
&{robust logit} &3.45\% &13.63\% & 9.41\% &10.04\%&45.92\%\\
&{abc-robust-logit} &3.20\% &11.92\% & \textbf{8.54\%} &9.45\%&{44.69\%}\\
\hline\hline
\end{tabular}

}
\end{center}
\label{tab_deep}
\end{table}

Table~\ref{tab_deep} reports the test error rates on five other datasets: M-Basic, M-Rotate, M-Image, M-Rand, and M-RotImg. In group 1 (as  reported in~\cite{Proc:Larochelle_ICML07}), the results show that (i) the kernel SVM with RBF kernel outperforms the kernel SVM with polynomial kernel; (ii) deep learning algorithms usually beat kernel SVM and neural nets. Group 2 presents the same results as in Table~\ref{tab_epGMM} (in terms of error rates as opposed to accuracies). We can see that pGMM and epGMM outperform deep learning methods except for M-Rand. In group 3, overall the tree methods especially abc-robust-logitboost  achieve very good accuracies. The results of pGMM and epGMM are largely comparable to the results of tree methods. \\

The training of boosted trees is typically  slow (especially in high-dimensional data) because a large number of trees are usually needed in order to achieve good accuracies. Consequently, the model sizes of tree methods are usually  large. Therefore, it would be exciting to have methods which are  simpler than trees and achieve comparable accuracies.

\newpage\clearpage

\section{Hashing the pGMM Kernel}

It is now well-understood that it is highly beneficial to be able to linearize nonlinear kernels so that learning algorithms can be easily scaled to massive data. The prior work~\cite{Report:Li_GMM16} has already demonstrated the effectiveness of the generalized consistent weighted sampling (GCWS)~\cite{Report:Manasse_CWS10,Proc:Ioffe_ICDM10,Proc:Li_KDD15} for hashing the GMM kernel. In this study, we  modify GCWS for linearizing the pGMM kernel  as summarized in Algorithm~\ref{alg_GCWS}.
\begin{algorithm}{
\textbf{Input:} Data vector $u_i$  ($i=1$ to $D$)

Generate vector $\tilde{u}$ in $2D$-dim by (\ref{eqn_transform}).

\vspace{0.08in}

For $i$ from 1 to $2D$

\hspace{0.1in}$r_i\sim Gamma(2, 1)$, $c_i\sim Gamma(2, 1)$,  $\beta_i\sim Uniform(0, 1)$

\hspace{0.05in} $t_i\leftarrow \lfloor \gamma\frac{\log \tilde{u}_i }{r_i}+\beta_i\rfloor$, $a_i\leftarrow \log(c_i)- r_i(t_i+1-\beta_i)$

End For

\textbf{Output:} $i^* \leftarrow arg\min_i \ a_i$,\hspace{0.3in}  $t^* \leftarrow t_{i^*}$
}\caption{Modified generalized consistent weighted sampling (GCWS) for hashing the pGMM kernel with a tuning parameter $\gamma$.}
\label{alg_GCWS}
\end{algorithm}

\noindent With $k$ samples, we  can  estimate $pGMM(u,v)$ according to the following collision probability:
\begin{align}
&\mathbf{Pr}\left\{i^*_{\tilde{u},j} = i^*_{\tilde{v},j} \ \text{and} \ t^*_{\tilde{u},j} = t^*_{\tilde{v},j}\right\} = pGMM(u,v),
\end{align}
or, for implementation  convenience, the approximate collision probability~\cite{Proc:Li_KDD15}:
\begin{align}\label{eqn_GCWS_Prob}
\mathbf{Pr}\left\{i^*_{\tilde{u},j} =  i^*_{\tilde{v},j}\right\} \approx pGMM({u},{v})
\end{align}

For each  vector $u$, we obtain $k$ random samples $i^*_{\tilde{u},j}$, $j=1$ to $k$. We store only the lowest $b$ bits of $i^*$. We need to view those $k$ integers as locations (of the nonzeros). For example, when $b=2$, we should view $i^*$ as a  binary vector of length $2^b=4$.  We  concatenate all $k$ such vectors into a binary vector of length $2^b\times k$, which contains exactly $k$ 1's.  We then feed the new data vectors to a linear classifier if the task is classification. The storage and computational cost is largely determined by the number of nonzeros in each data vector, i.e., the $k$ in our case. This scheme can of course also be used for  many other tasks including clustering, regression, and near neighbor search.\\

\begin{figure}[h!]
\begin{center}

\mbox{
\includegraphics[width=2.7in]{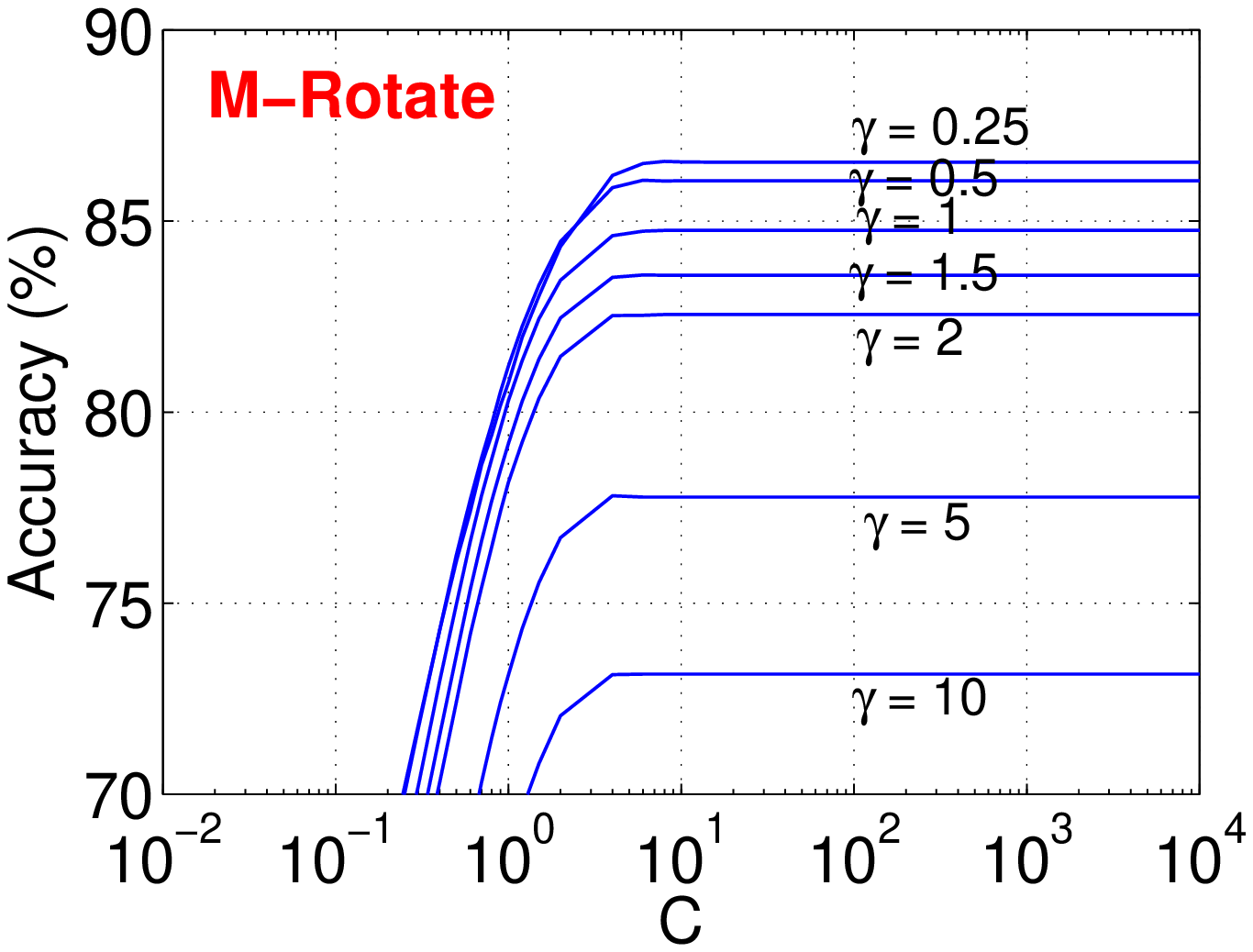}
\includegraphics[width=2.7in]{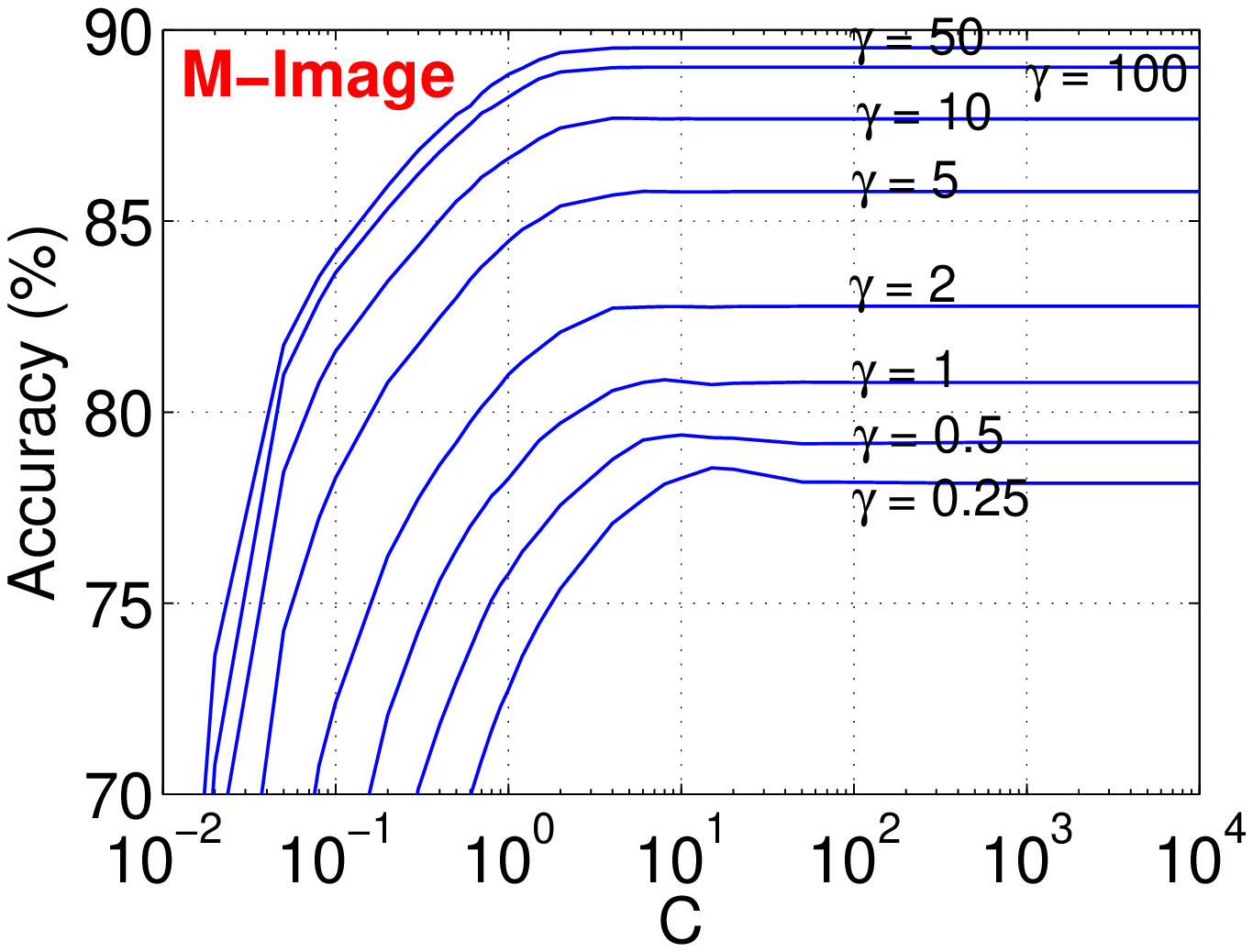}
}

\end{center}
\vspace{-0.3in}
\caption{\textbf{Impact of $\gamma$} on the pGMM kernel SVM classification accuracies for two datasets.}\label{fig_g}-\vspace{-0.3in}
\end{figure}

Note that the performance of pGMM can be heavily impacted by the tuning parameter $\gamma$ in the definition of the pGMM kernel. Figure~\ref{fig_g} presents  examples on  M-Rotate and M-Image.

\newpage\clearpage

\begin{figure}[h!]
\begin{center}

\mbox{
\includegraphics[width=2.7in]{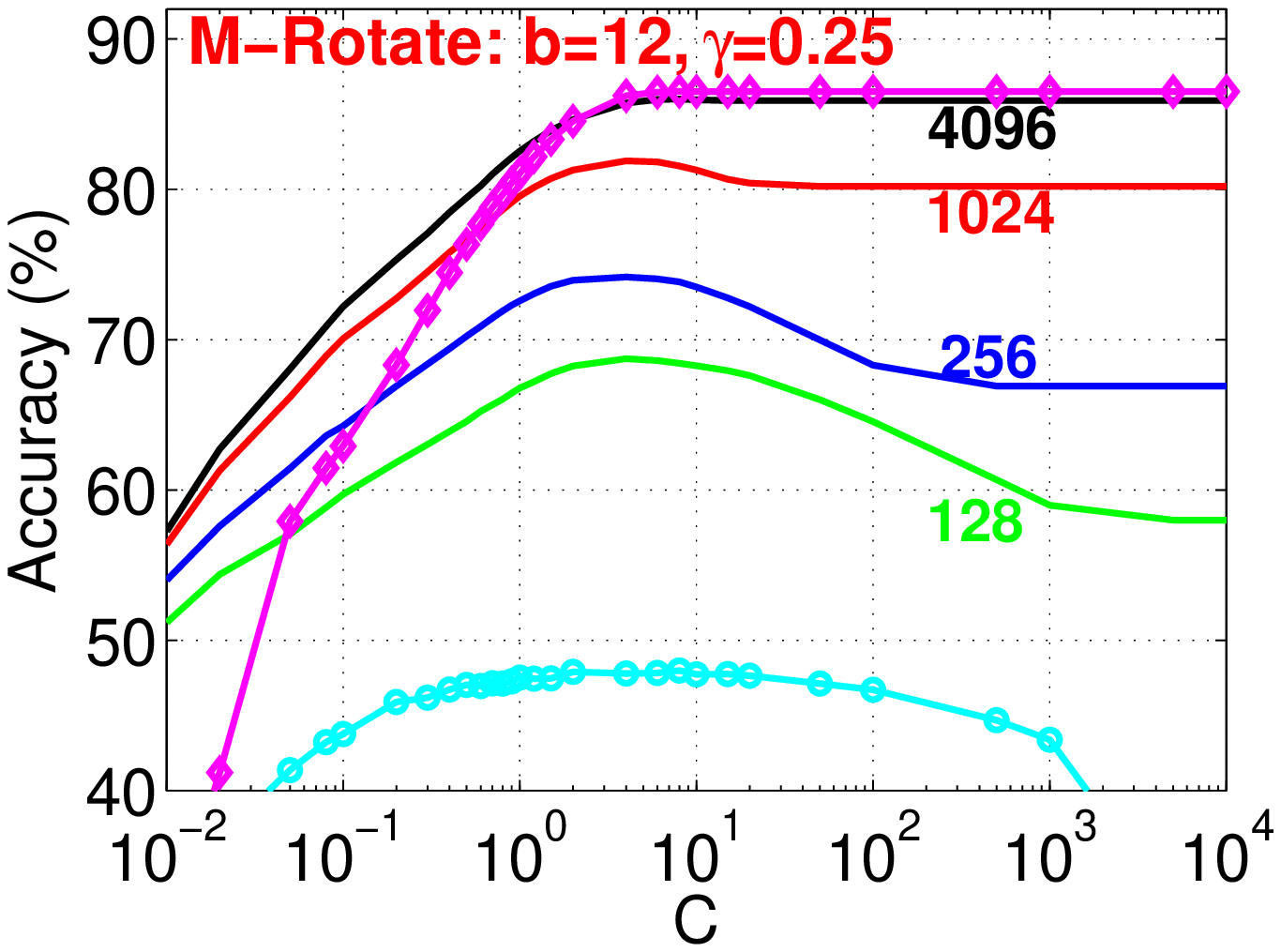}
\includegraphics[width=2.7in]{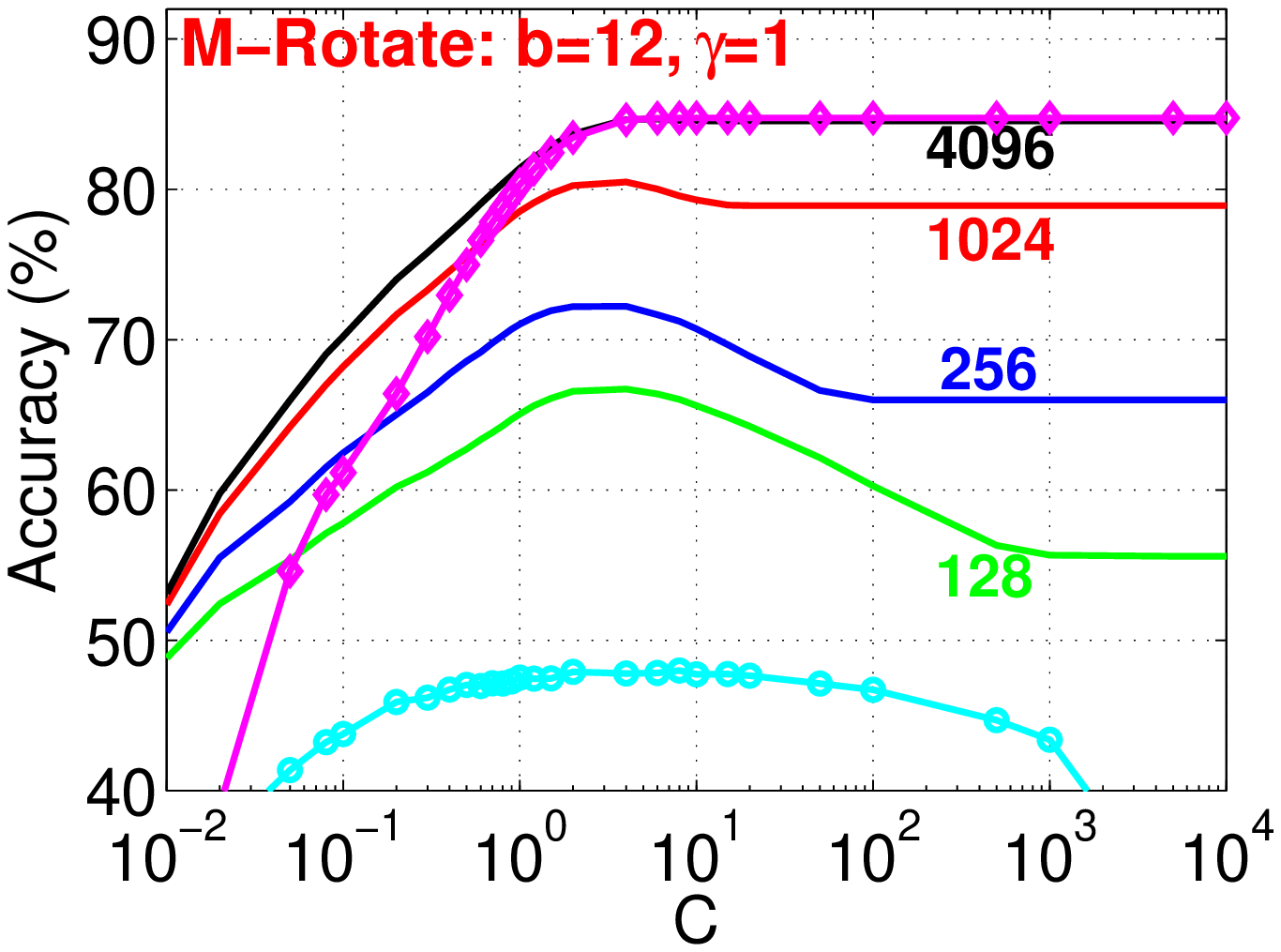}
}

\mbox{
\includegraphics[width=2.7in]{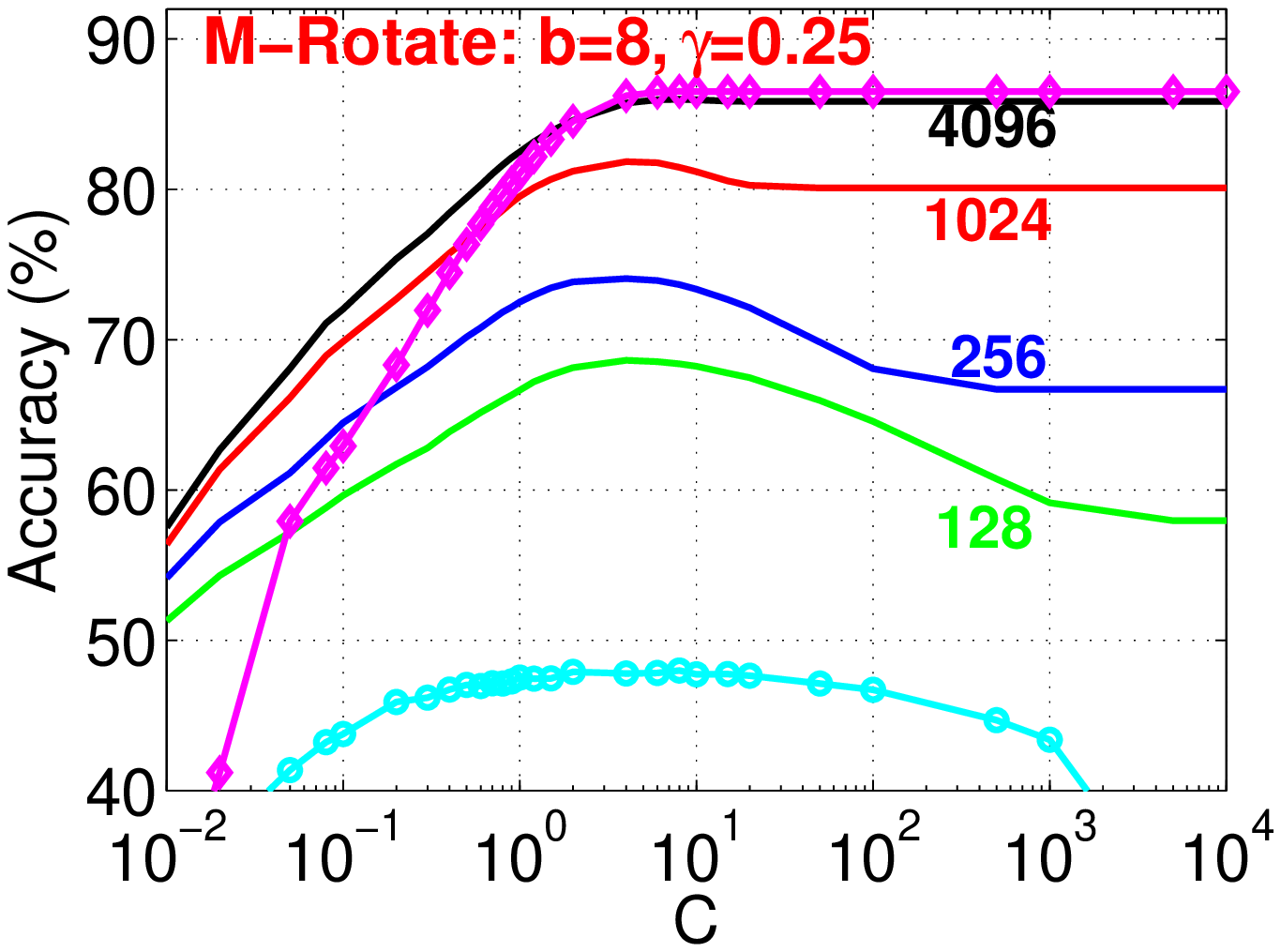}
\includegraphics[width=2.7in]{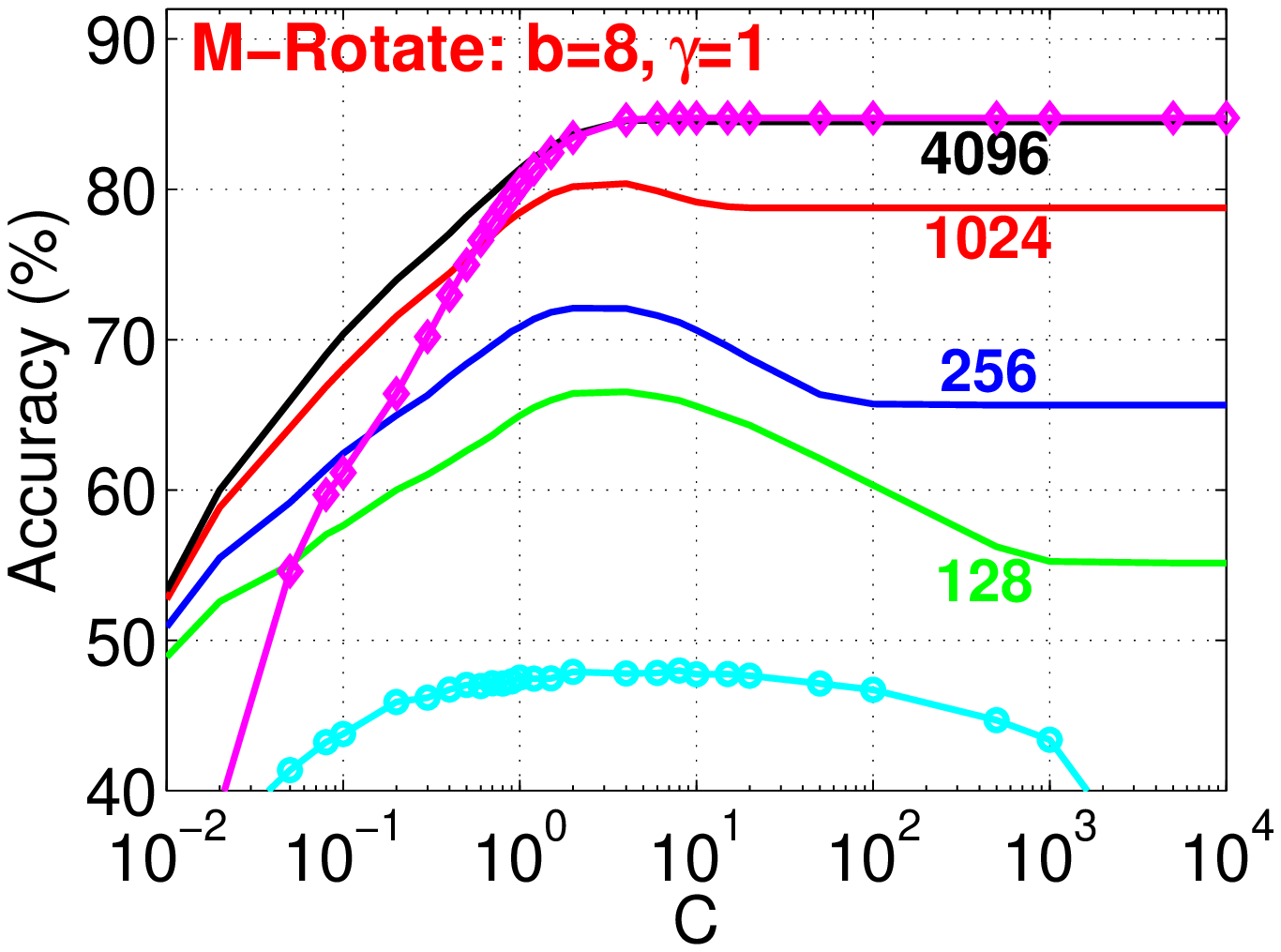}
}

\mbox{
\includegraphics[width=2.7in]{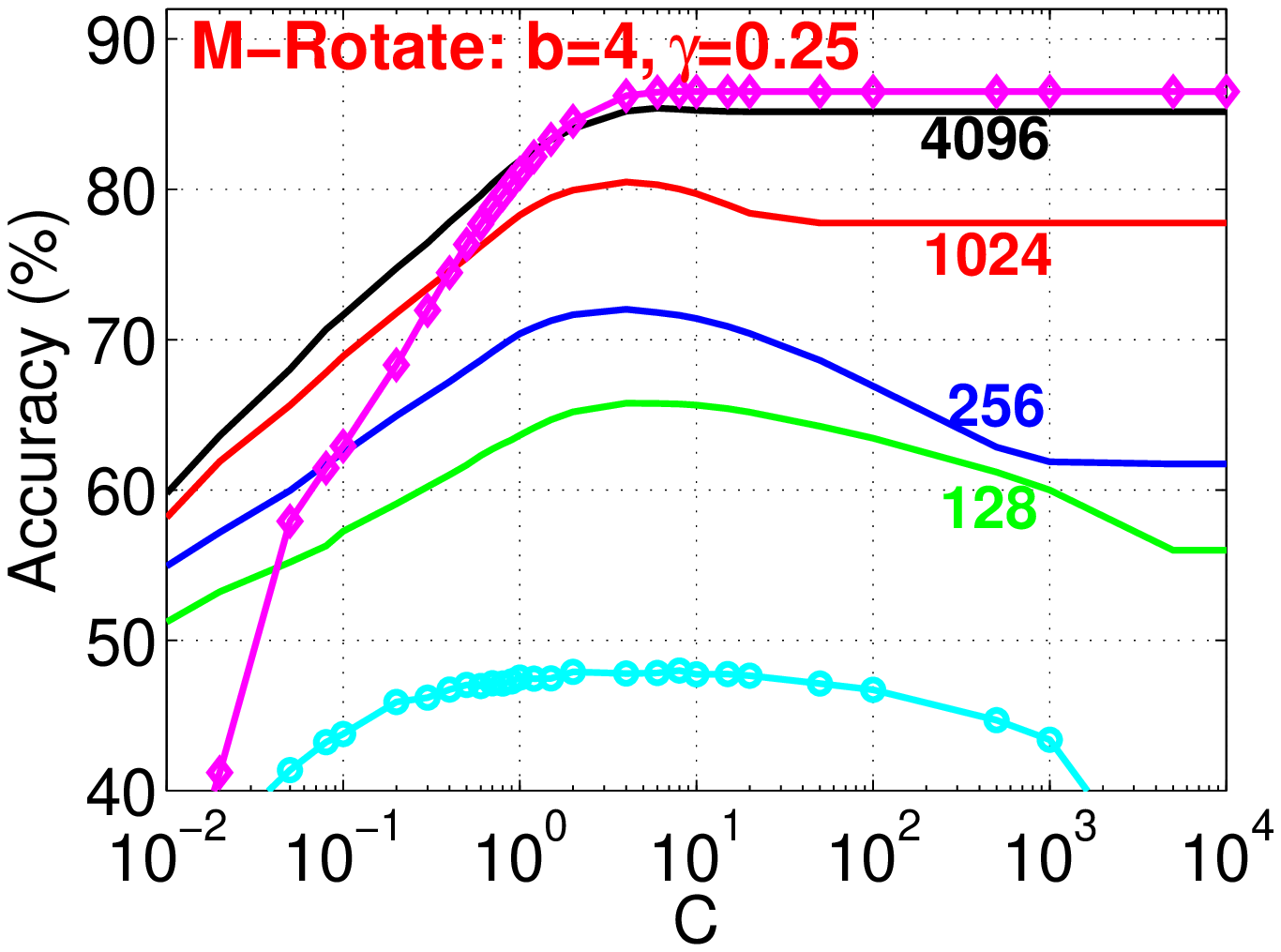}
\includegraphics[width=2.7in]{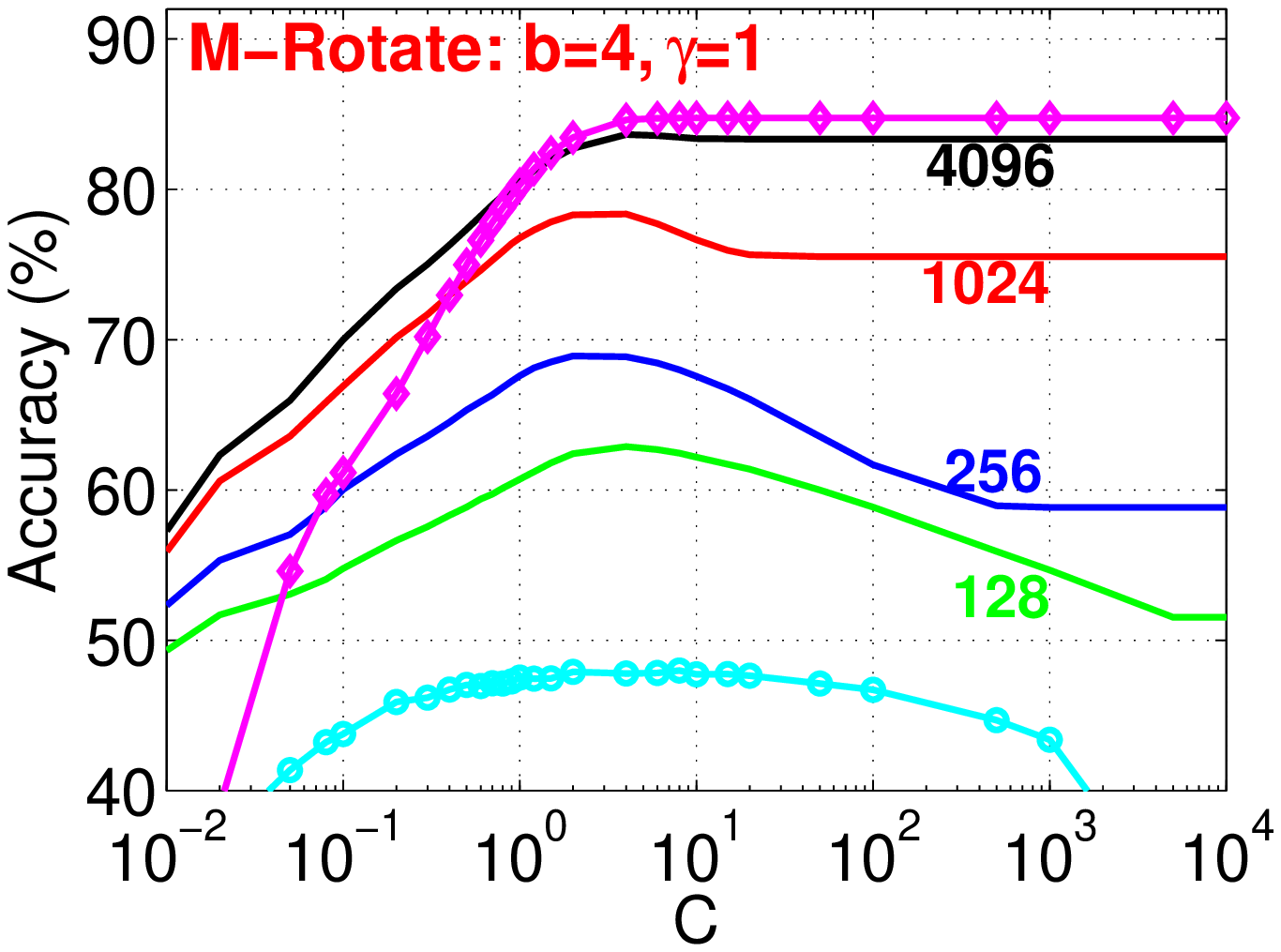}
}

\mbox{
\includegraphics[width=2.7in]{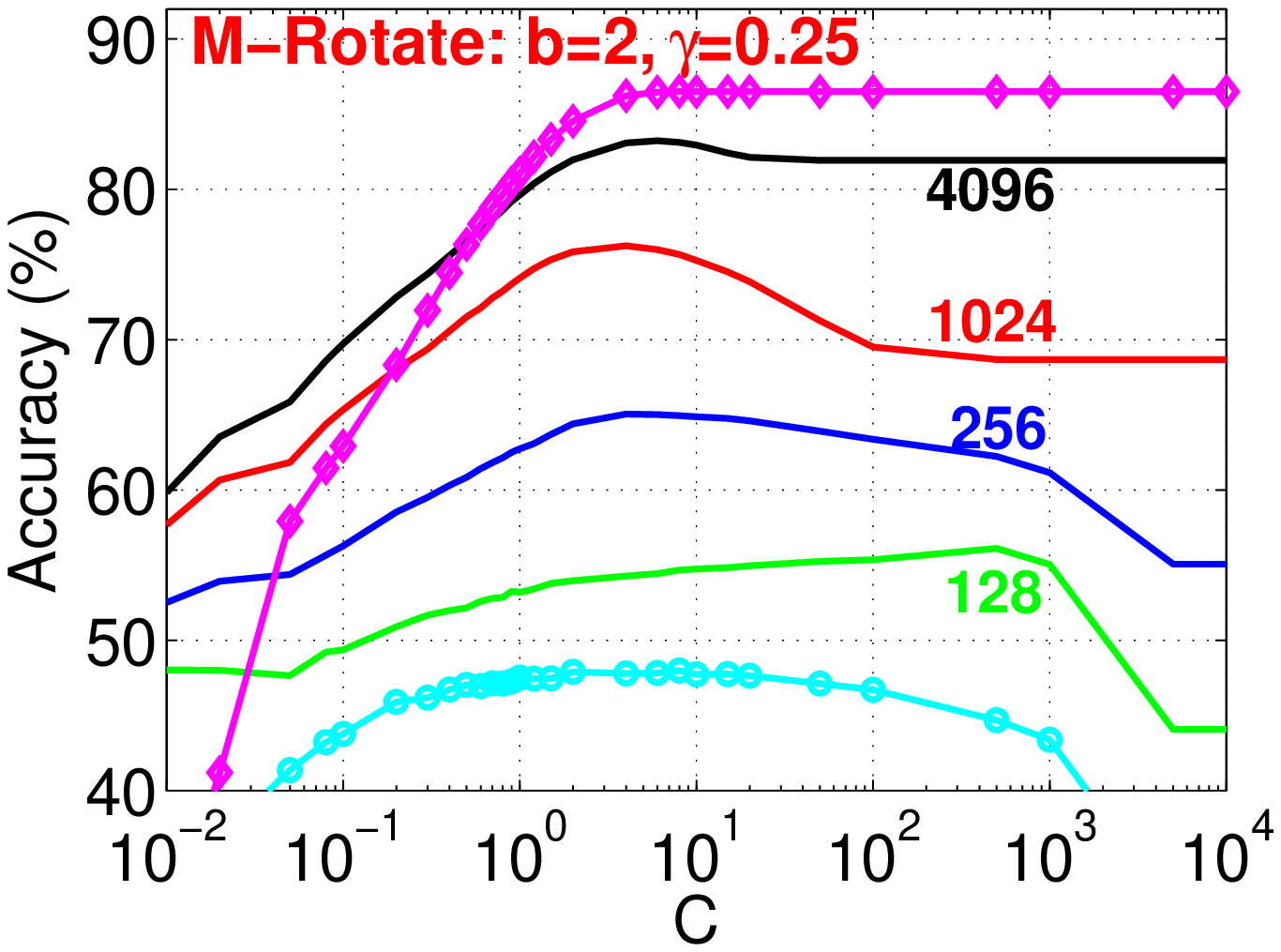}
\includegraphics[width=2.7in]{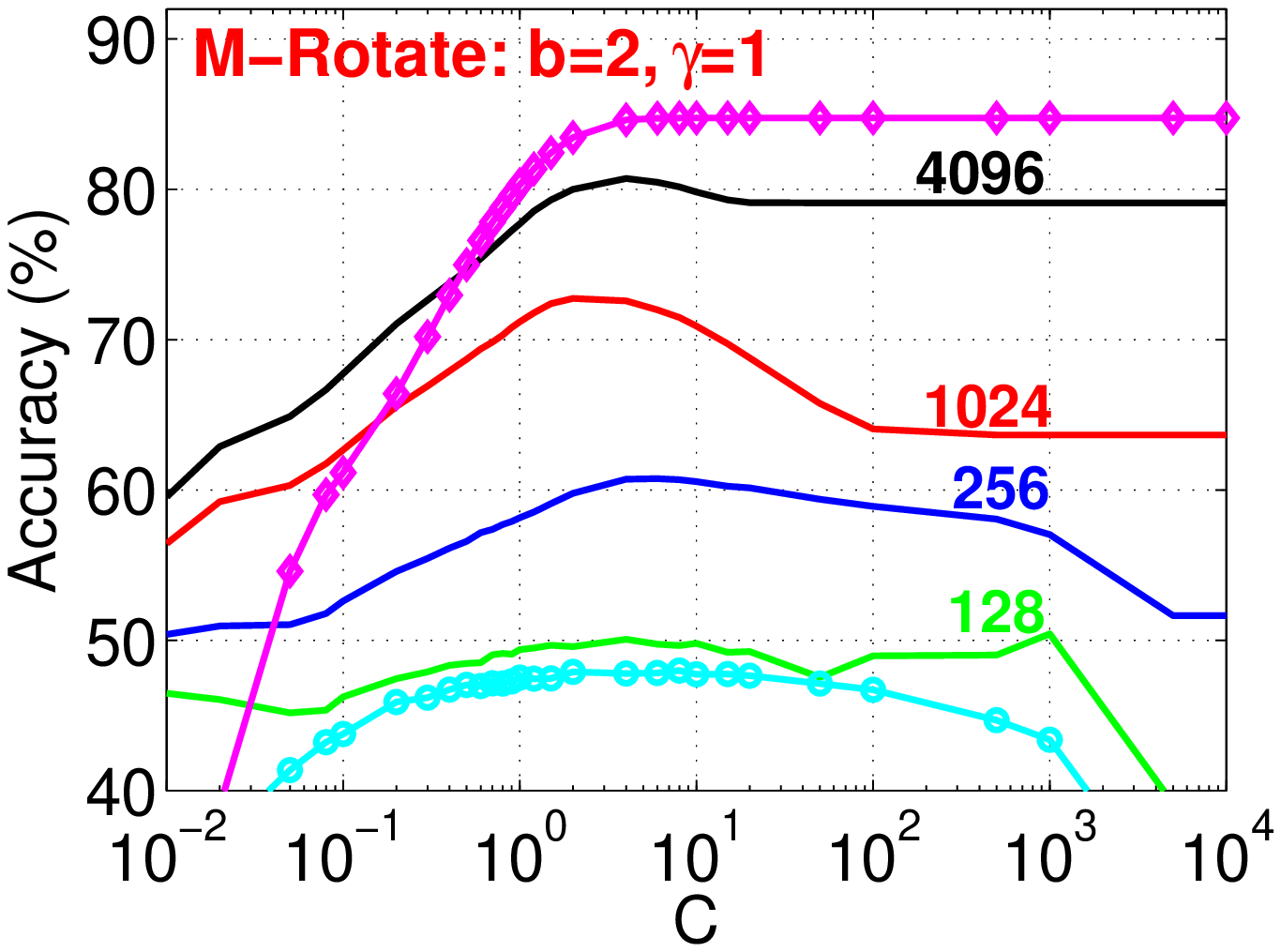}
}

\end{center}
\vspace{-0.3in}
\caption{Test classification accuracies for using linear classifiers combined with hashing in Algorithm~\ref{alg_GCWS} on M-Rotate dataset, for $\gamma = 0.25$ (left panels) and $\gamma = 1$ (right panels), and $b\in\{12, 8, 4, 2\}$. In each panel, the four solid curves correspond to results with $k$ hashes for $k\in\{64, 128, 256,1024\}$. For comparisons, in each panel we also plot the results of linear classifiers on the original data (lower curve) and the results of pGMM kernel SVMs (higher curve), in two marked solid curves. }\label{fig_HashM-Rotate}
\end{figure}

Figure~\ref{fig_HashM-Rotate} presents the experimental results on hashing for  M-Rotate. For this dataset, $\gamma=0.25$ is the best choice (among the range of $\gamma$ values we have searched). Figure~\ref{fig_HashM-Rotate} plots the results for both $\gamma=0.25$ (left panels) and $\gamma=1$ (right panels), for $b\in\{12, 8, 4, 2\}$. Recall here $b$ is the number of bits for representing each hashed value in the ``0-bit CWS'' scheme~\cite{Proc:Li_KDD15}. The results demonstrate that: (i) hashing using $\gamma=0.25$ produces better results than hashing using $\gamma=1$; (ii) It is preferable  to use a fairly large $b$ value, for example, $b\geq 4$ or 8. Using smaller $b$ values (e.g., $b=2$) hurts the accuracy; (iii) With merely a small number of hashes (e.g., $k=128$), the linearized pGMM kernel can significantly outperform the original linear kernel. Note that the original dimensionality is 784. This example illustrates the significant advantage of nonlinear kernel and hashing.\\

\begin{figure}[h!]
\begin{center}

\mbox{
\includegraphics[width=2.7in]{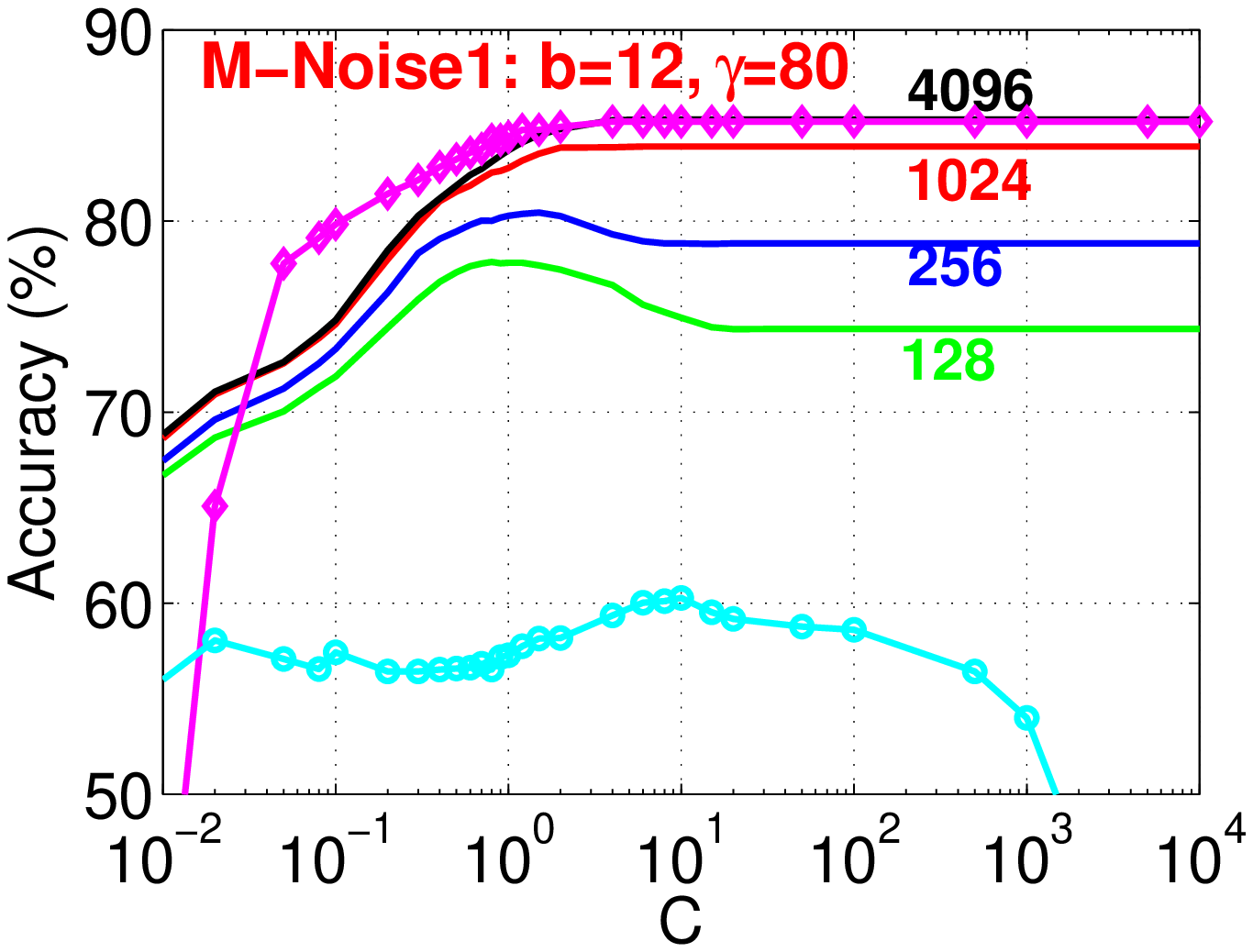}
\includegraphics[width=2.7in]{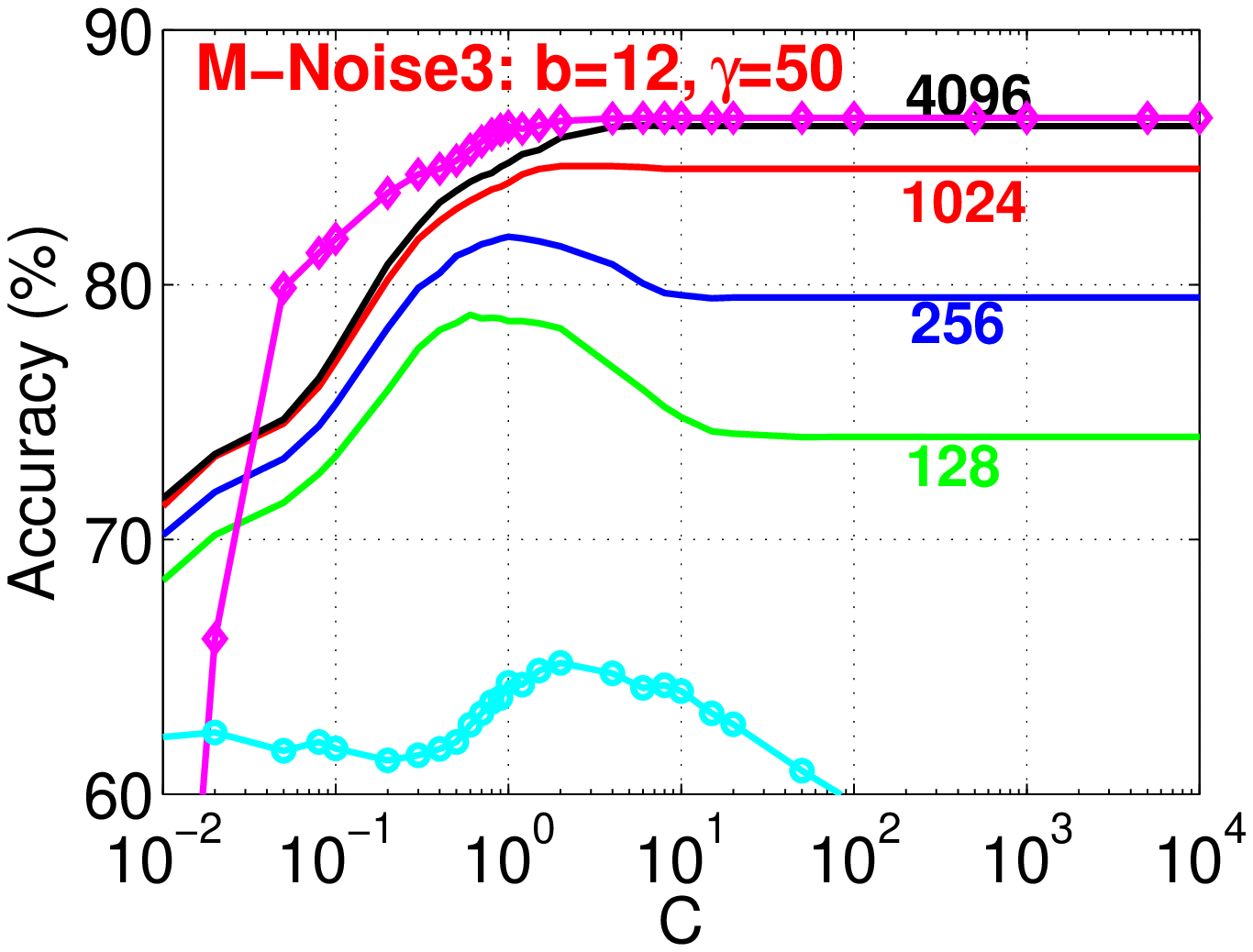}
}

\mbox{
\includegraphics[width=2.7in]{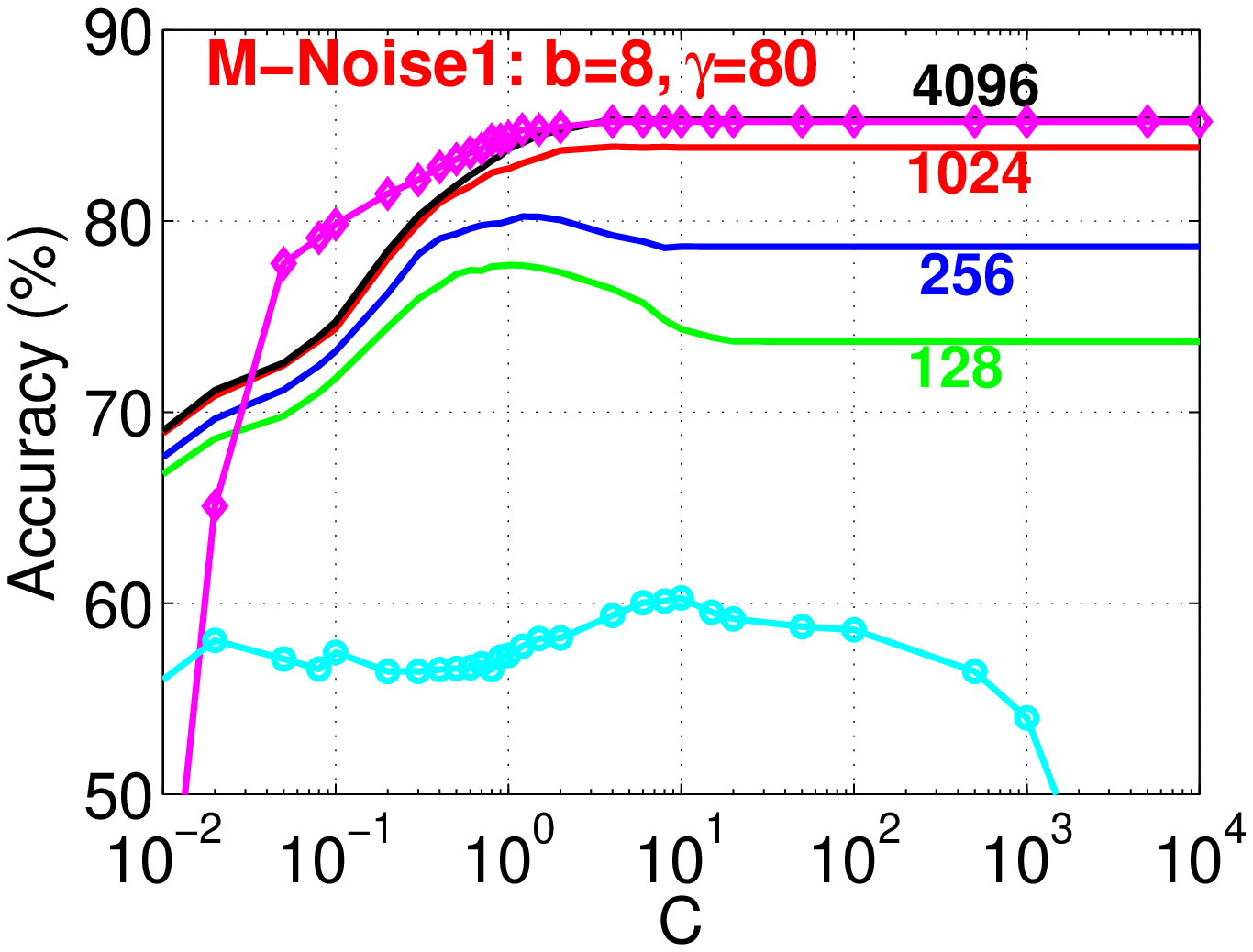}
\includegraphics[width=2.7in]{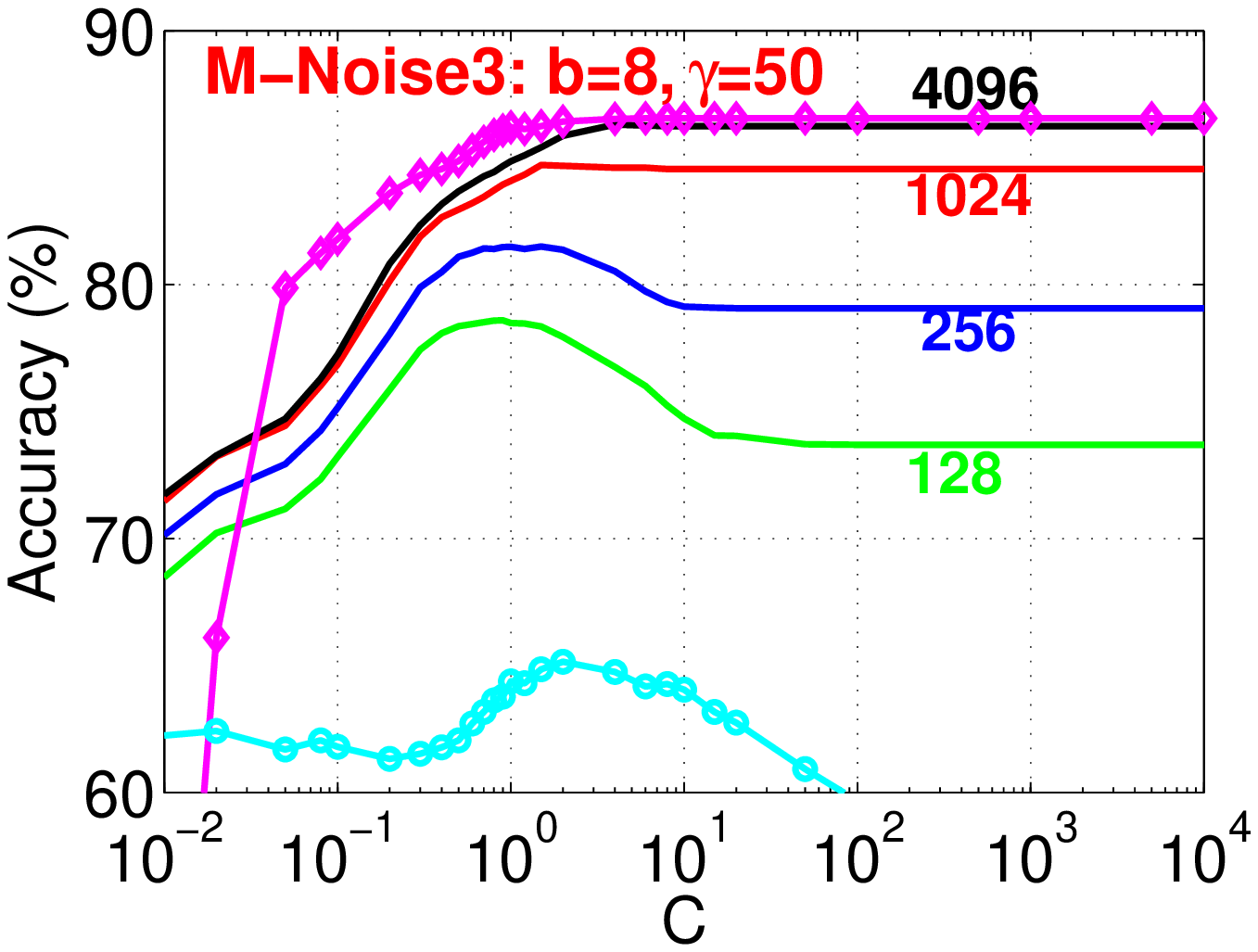}
}

\mbox{
\includegraphics[width=2.7in]{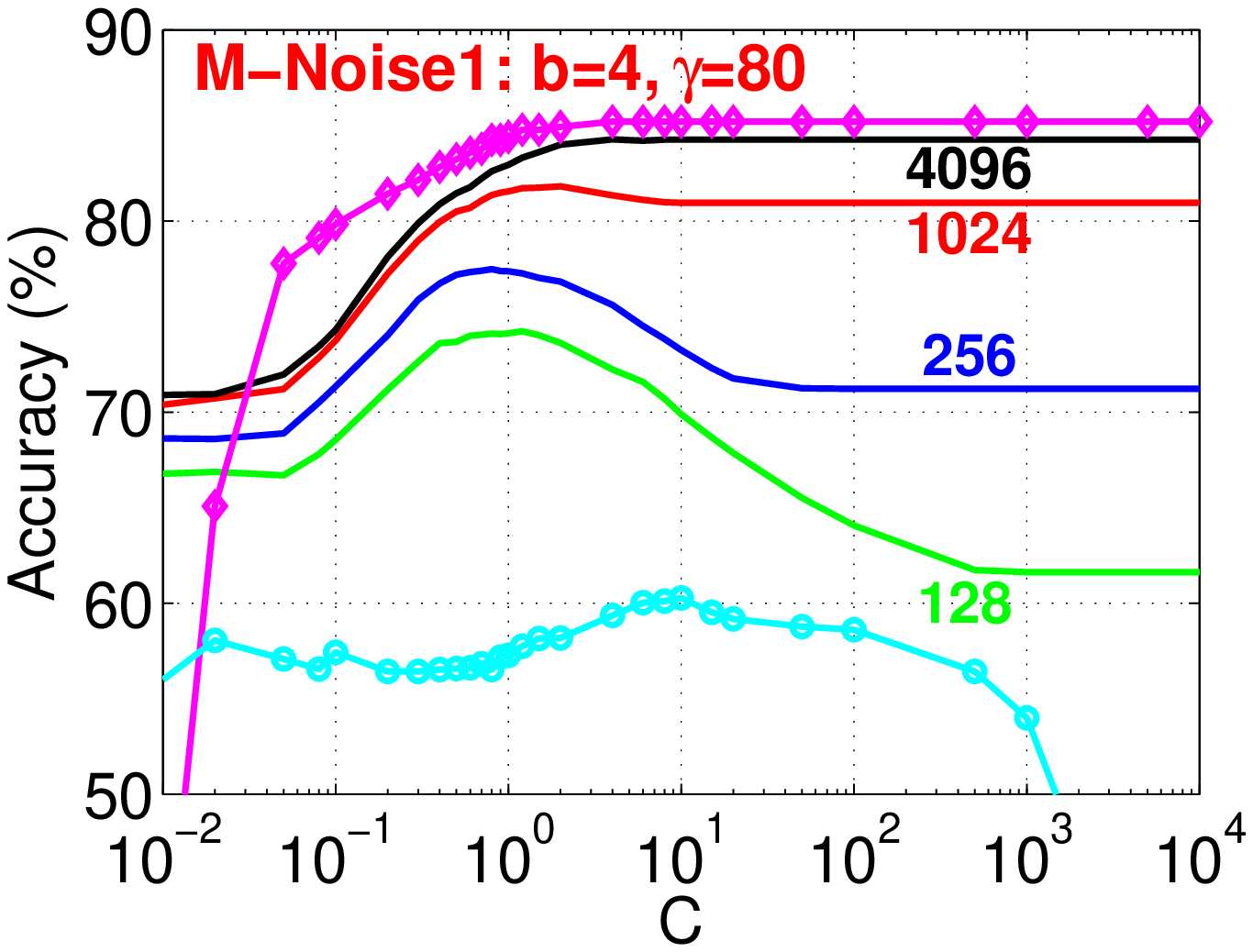}
\includegraphics[width=2.7in]{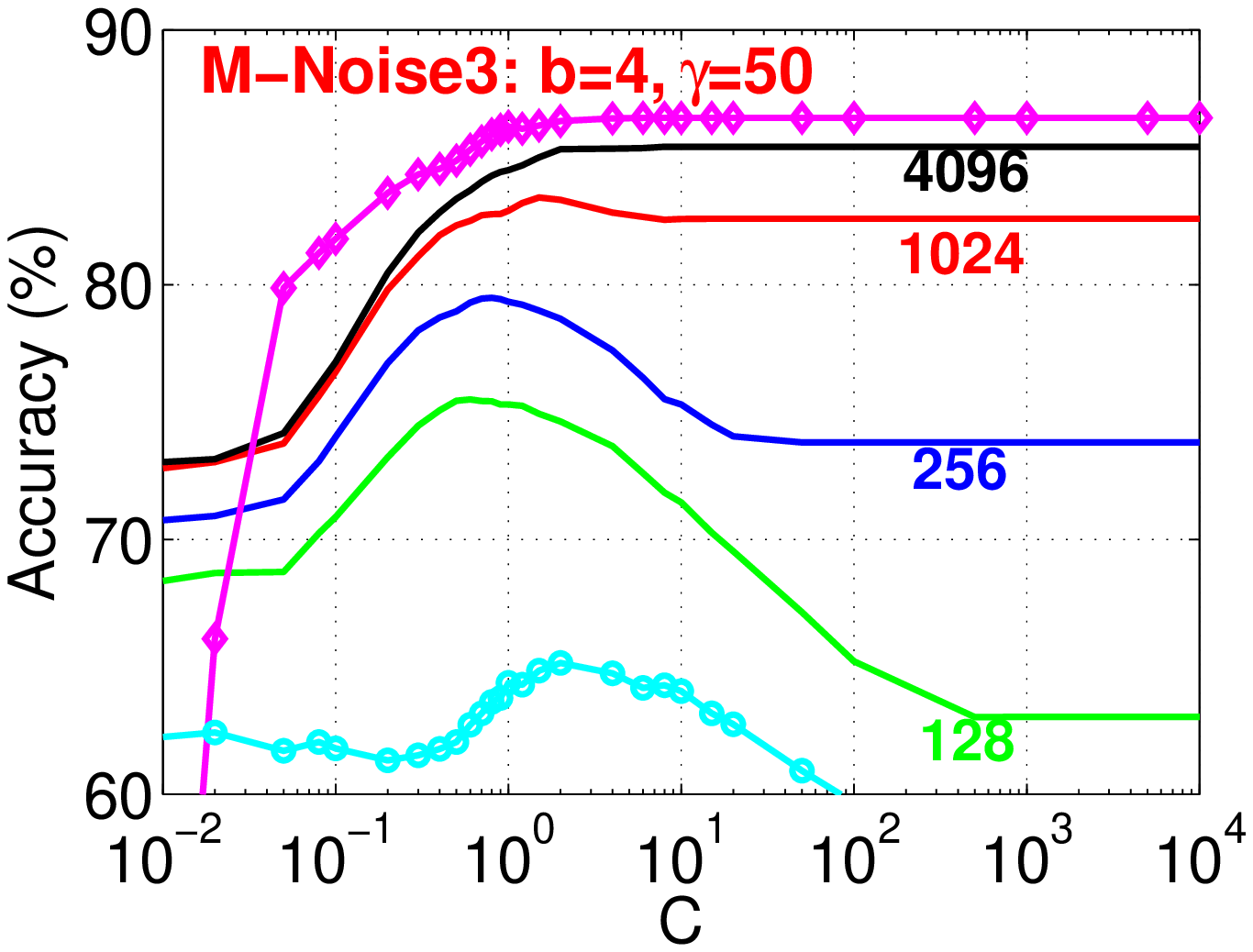}
}

\mbox{
\includegraphics[width=2.7in]{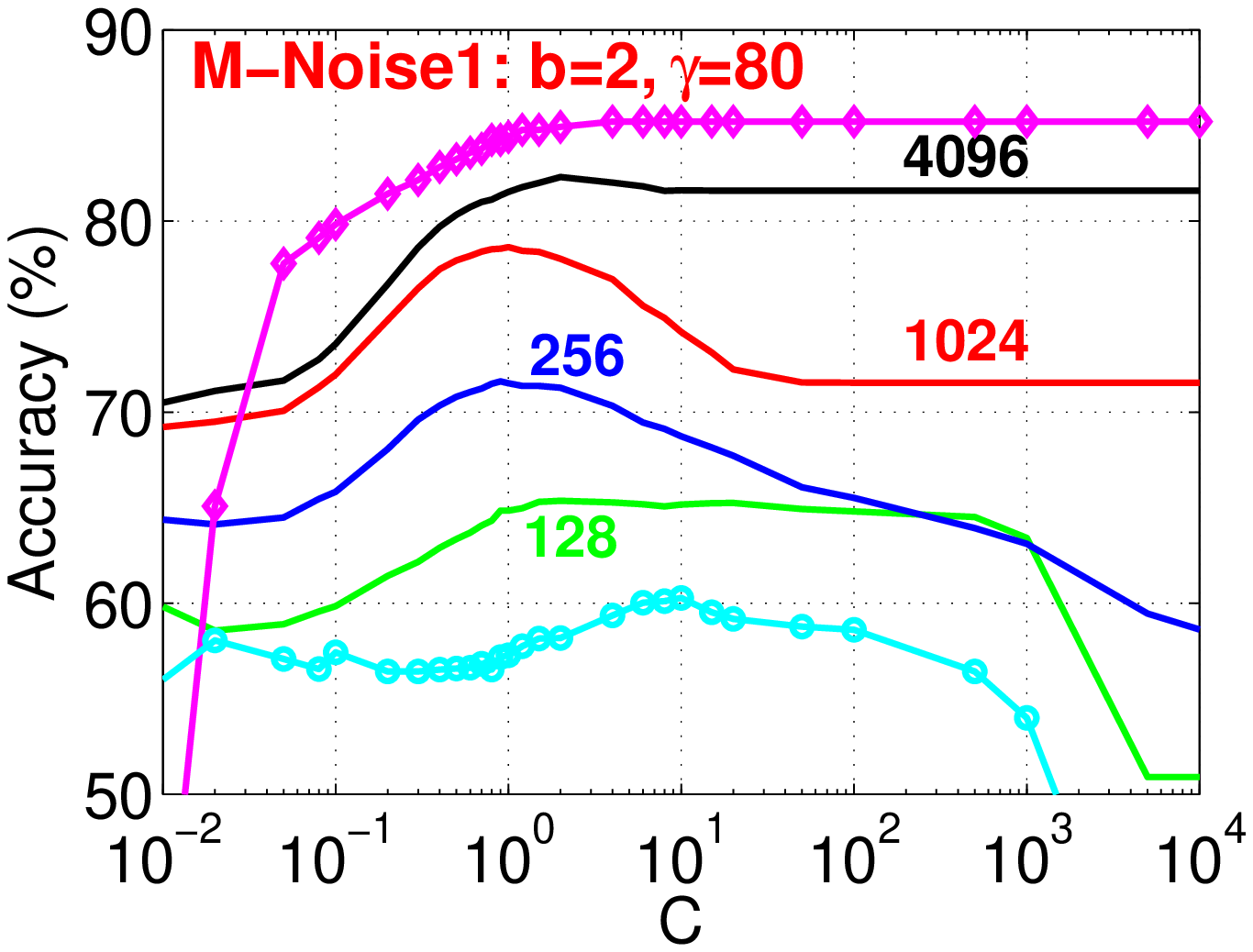}
\includegraphics[width=2.7in]{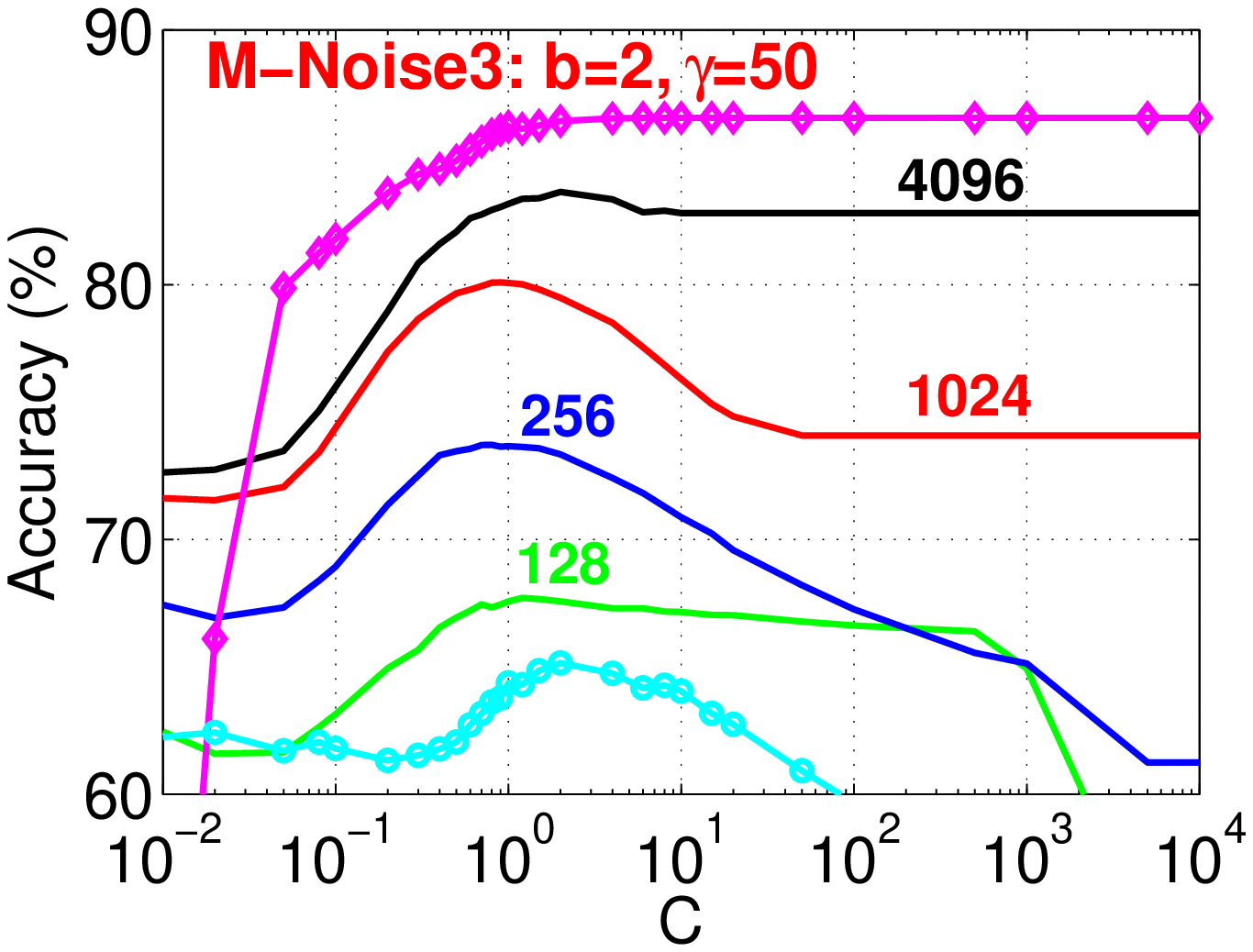}
}

\end{center}
\vspace{-0.3in}
\caption{Test classification accuracies for using linear classifiers combined with hashing in Algorithm~\ref{alg_GCWS} on M-Noise1 dataset (left panels) and M-Noise3 dataset (right panels). In each panel, the four solid curves correspond to results with $k$ hashes for $k\in\{64, 128, 256,1024\}$. For comparisons, in each panel we also plot the results of linear classifiers on the original data (lower curve) and the results of pGMM kernel SVMs (higher curve), in two marked solid curves. }\label{fig_HashM-Noise}
\end{figure}

\begin{figure}[h!]
\begin{center}

\mbox{
\includegraphics[width=2.7in]{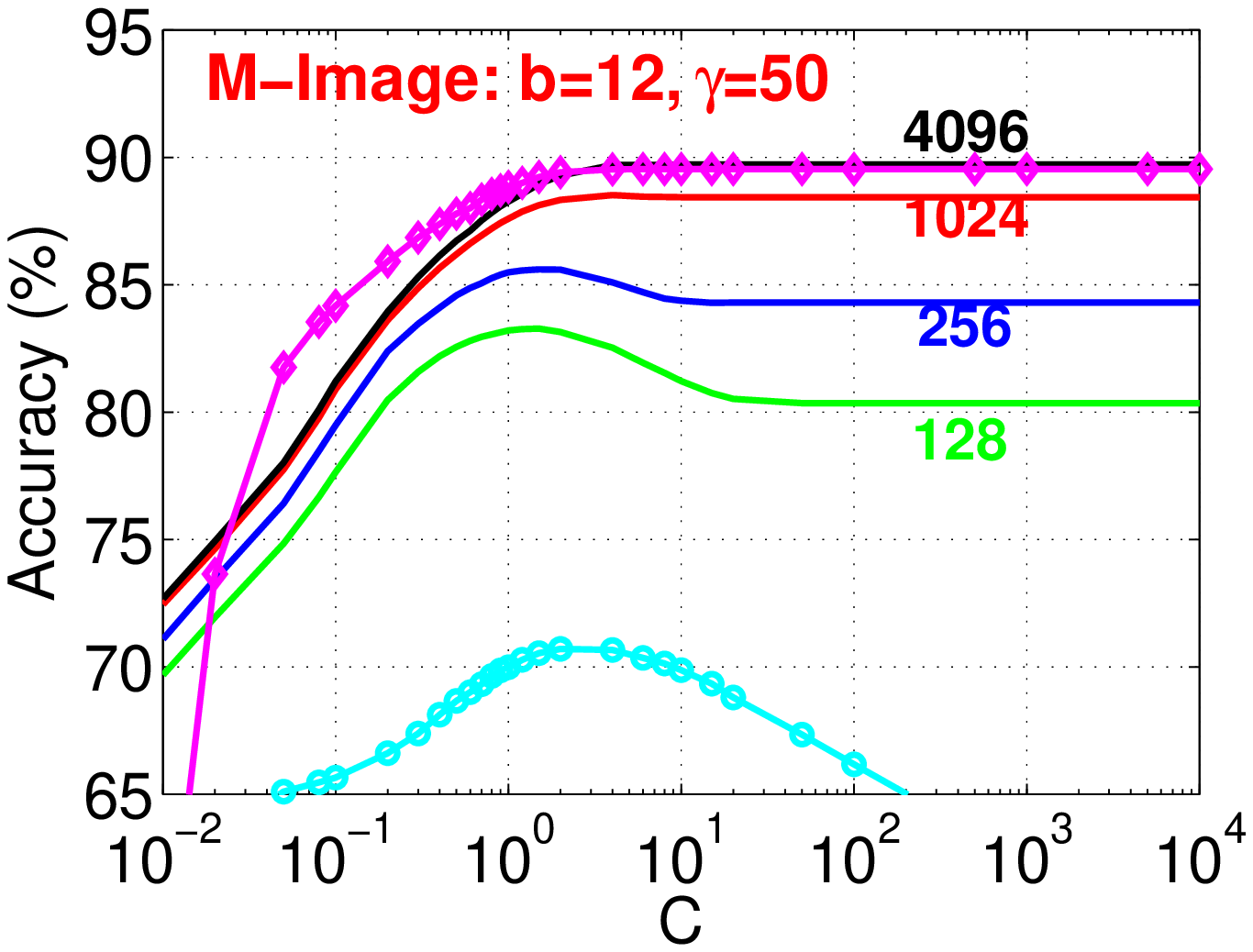}
\includegraphics[width=2.7in]{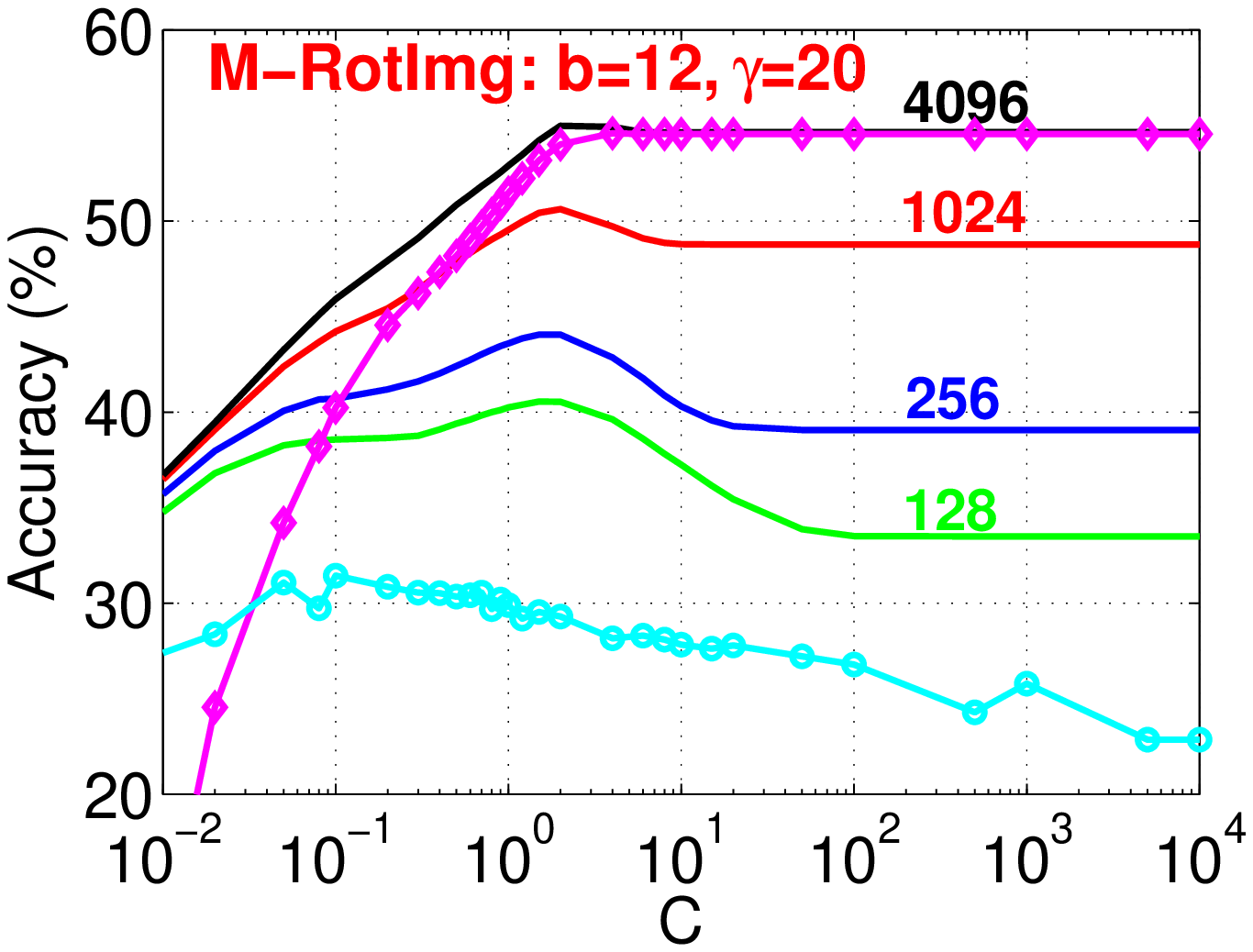}
}

\mbox{
\includegraphics[width=2.7in]{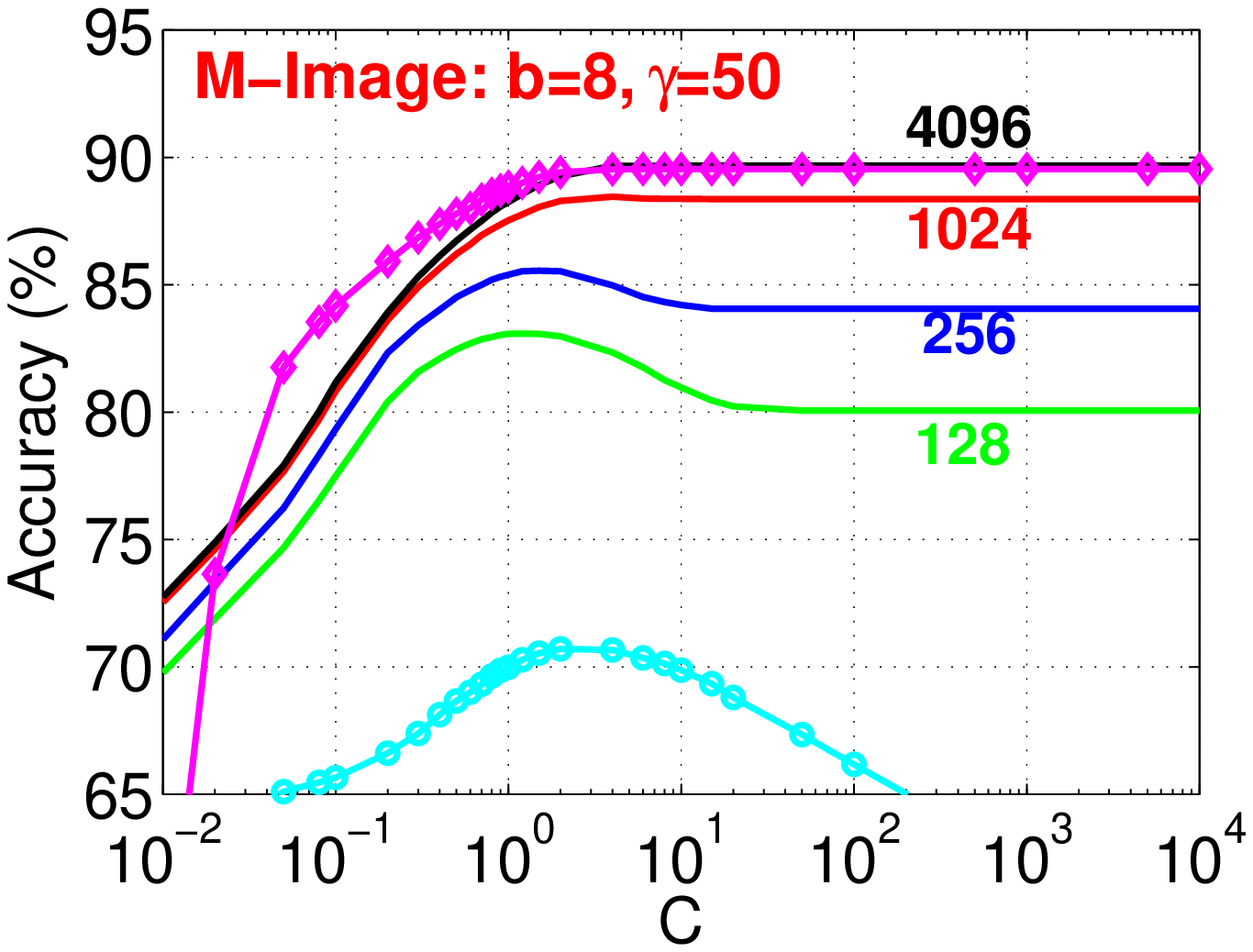}
\includegraphics[width=2.7in]{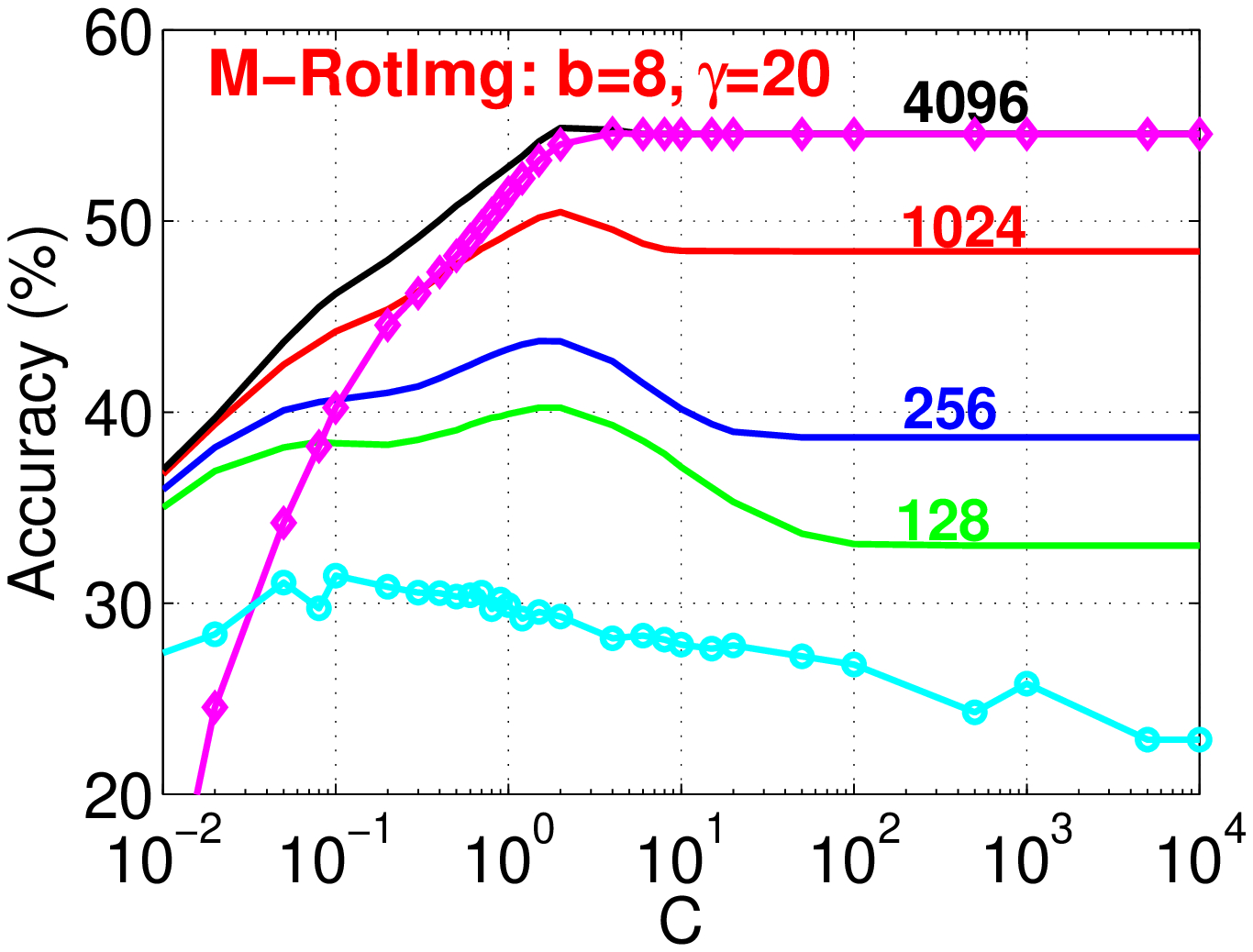}
}

\mbox{
\includegraphics[width=2.7in]{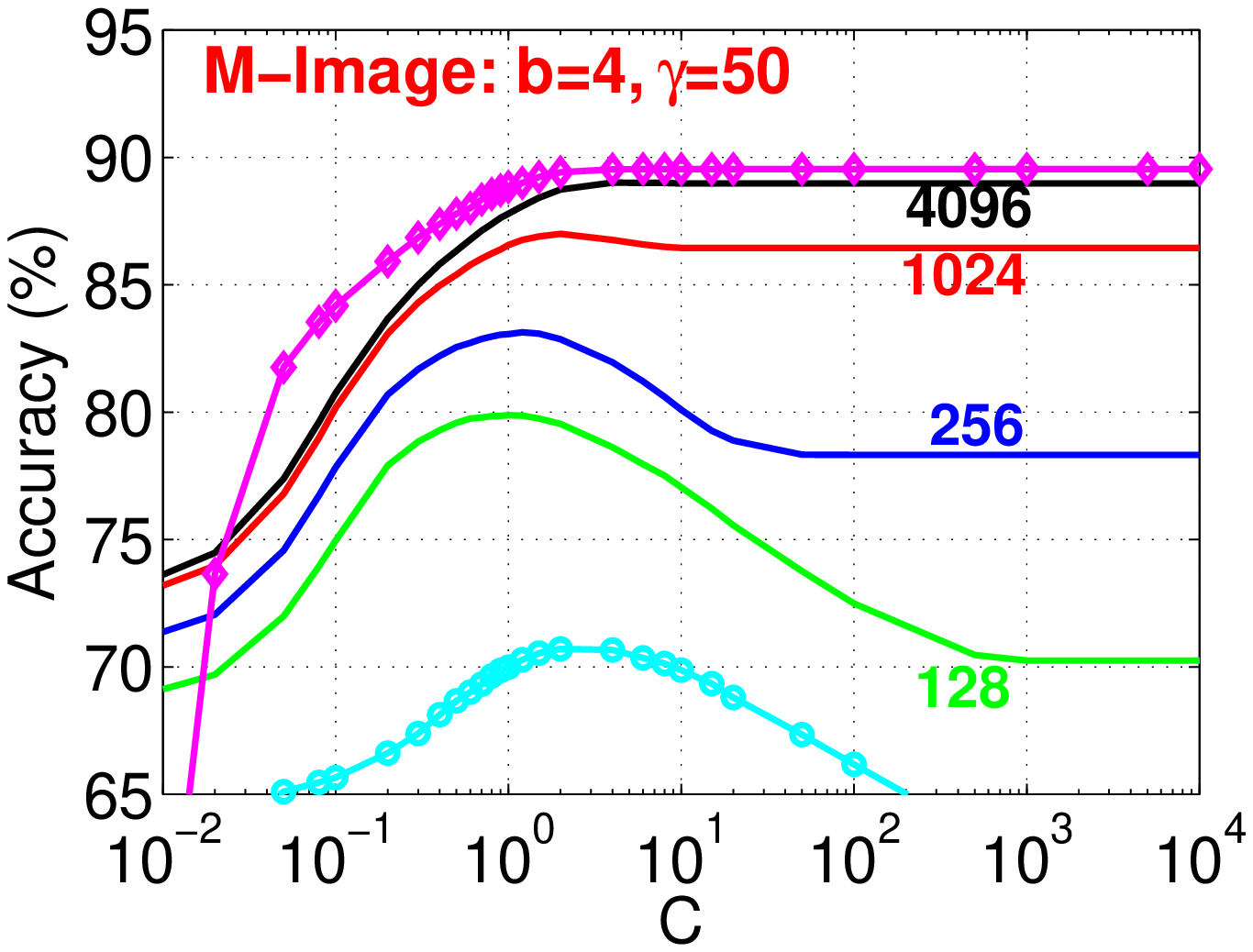}
\includegraphics[width=2.7in]{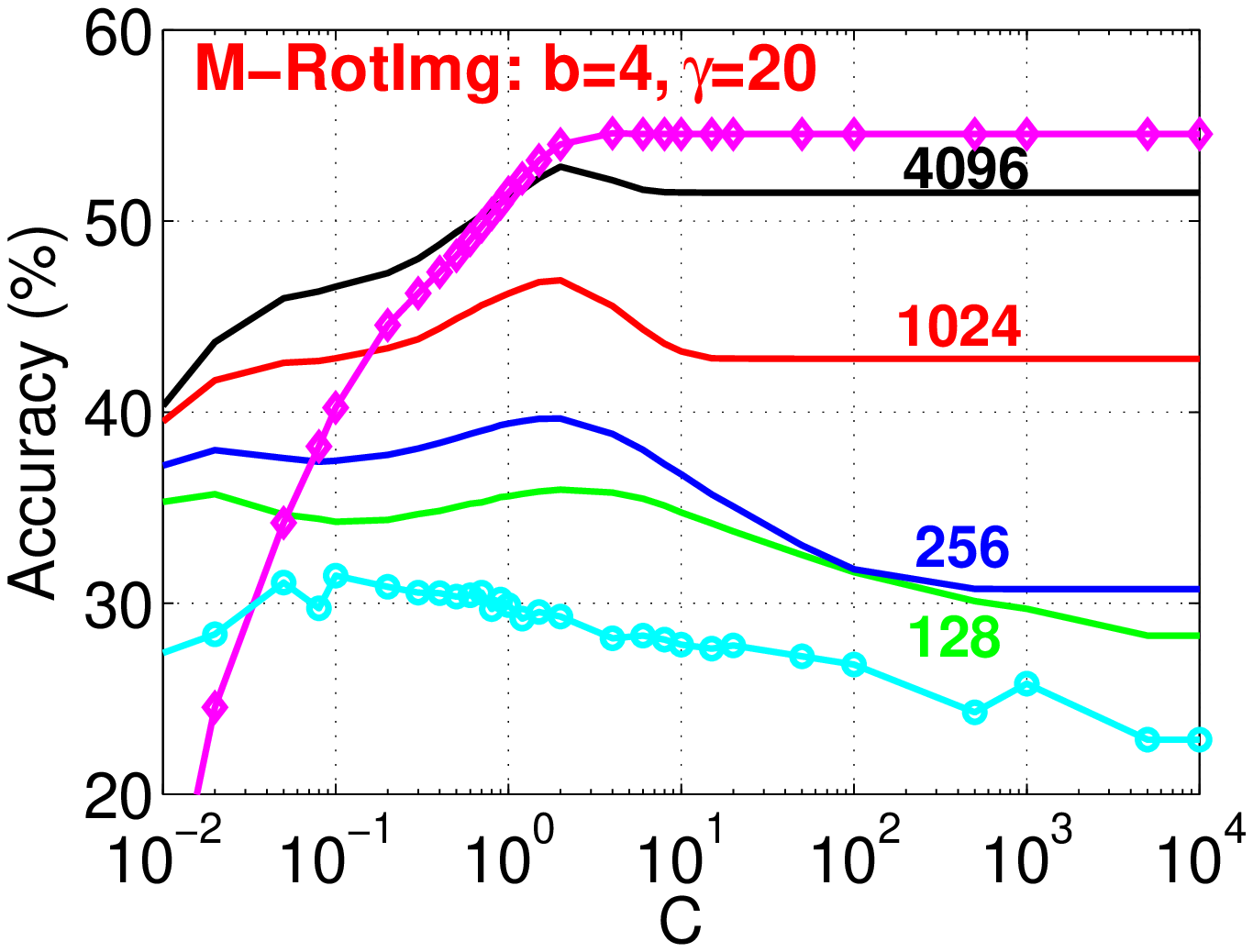}
}

\mbox{
\includegraphics[width=2.7in]{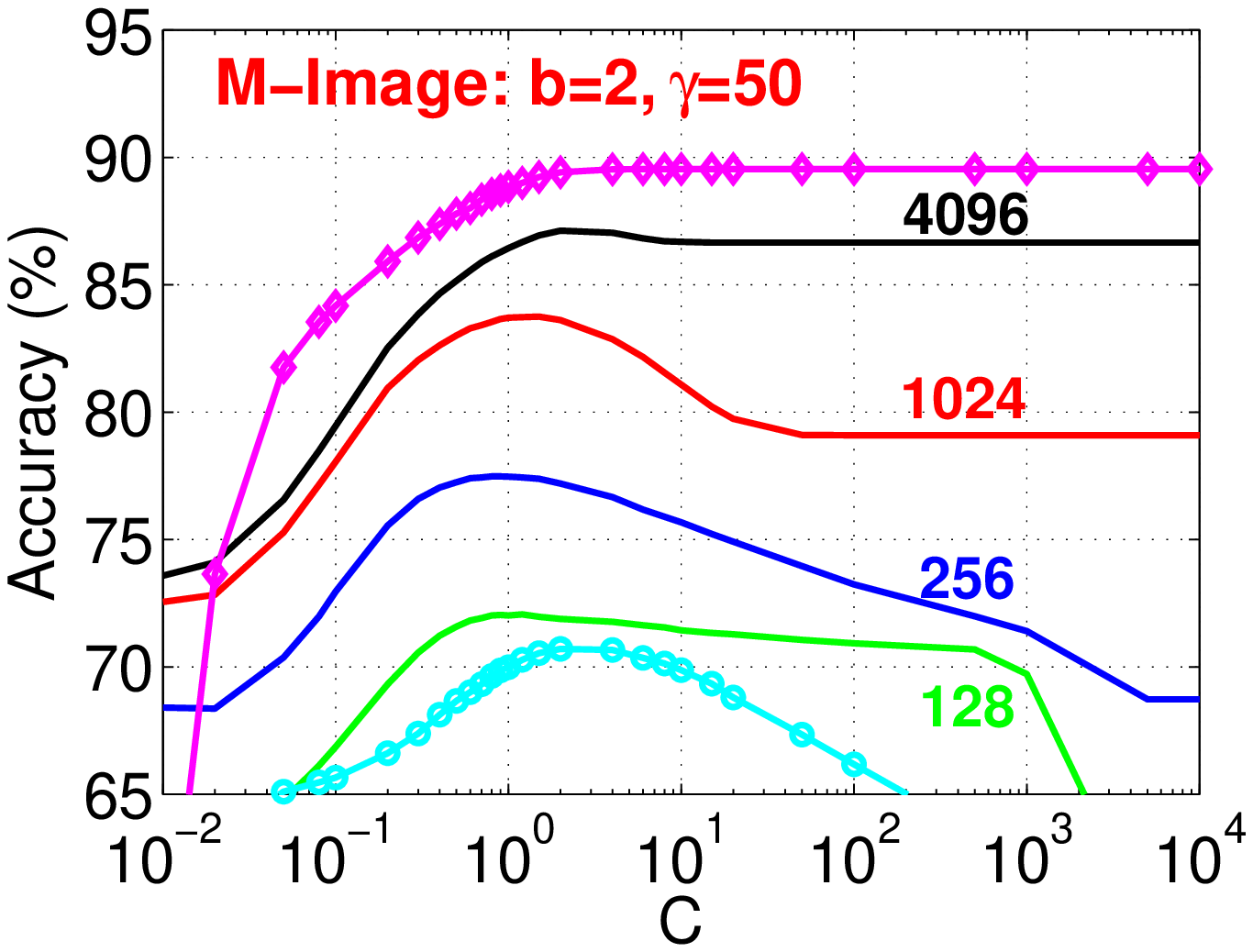}
\includegraphics[width=2.7in]{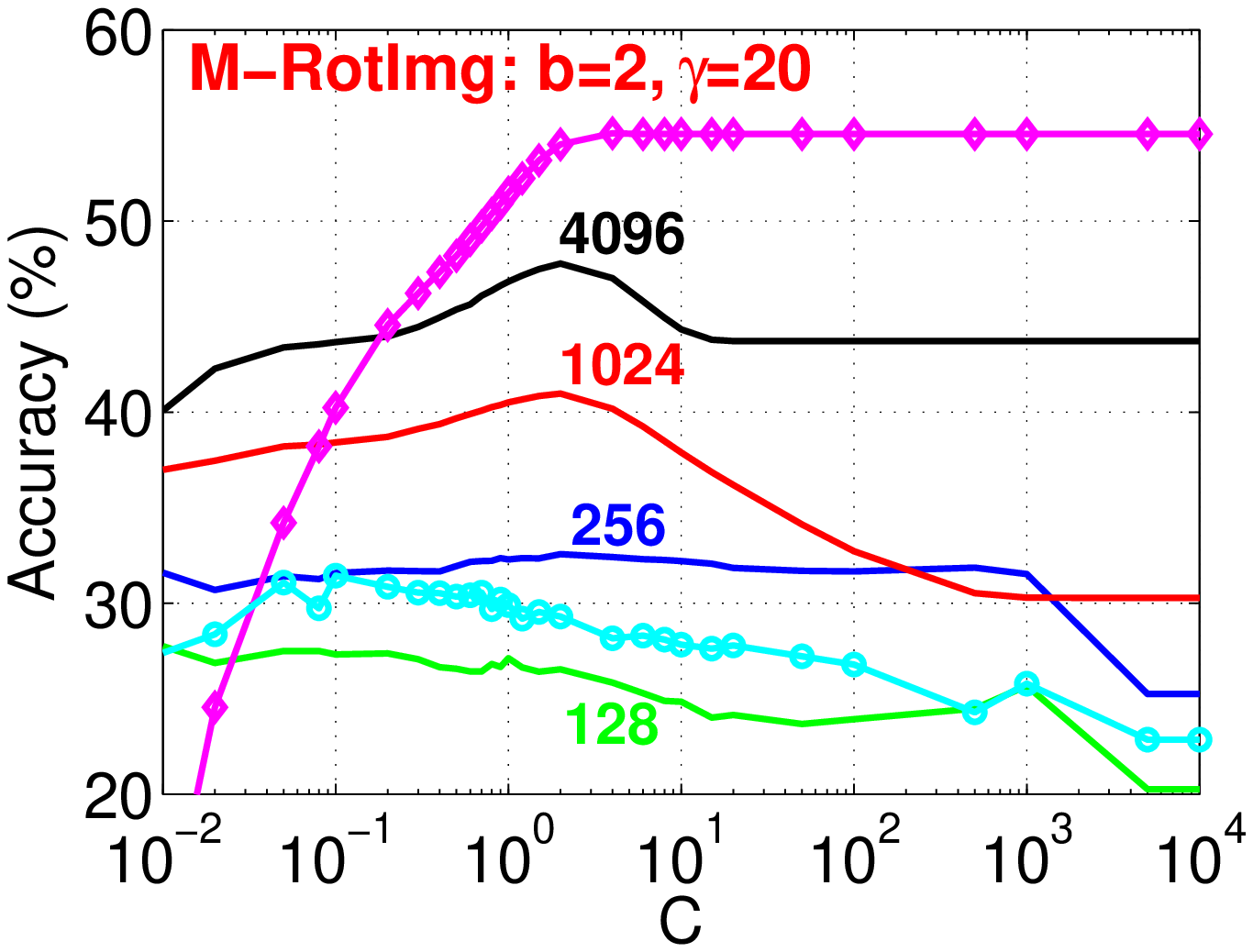}
}

\end{center}
\vspace{-0.3in}
\caption{Test classification accuracies for using linear classifiers combined with hashing in Algorithm~\ref{alg_GCWS} on M-Image dataset (left panels) and M-RotImg dataset (right panels). In each panel, the four solid curves correspond to results with $k$ hashes for $k\in\{64, 128, 256,1024\}$. For comparisons, in each panel we also plot the results of linear classifiers on the original data (lower curve) and the results of pGMM kernel SVMs (higher curve), in two marked solid curves.  }\label{fig_HashM-Img}
\end{figure}

Figure~\ref{fig_HashM-Noise} presents the experimental results on hashing for M-Noise1 dataset and M-Noise3 dataset, respectively on the left panels ($\gamma=80$) and the right panels ($\gamma=50$). Figure~\ref{fig_HashM-Img} presents the experimental results on hashing for M-Image dataset and M-RotImg dataset, respectively on the left panels ($\gamma=50$) and the right panels ($\gamma=20$). These results  deliver similar information as the results in Figure~\ref{fig_HashM-Rotate}, confirming the significant advantages of the pGMM kernel and hashing.\\

\begin{figure}[h!]
\begin{center}
\mbox{
\includegraphics[width=2.3in]{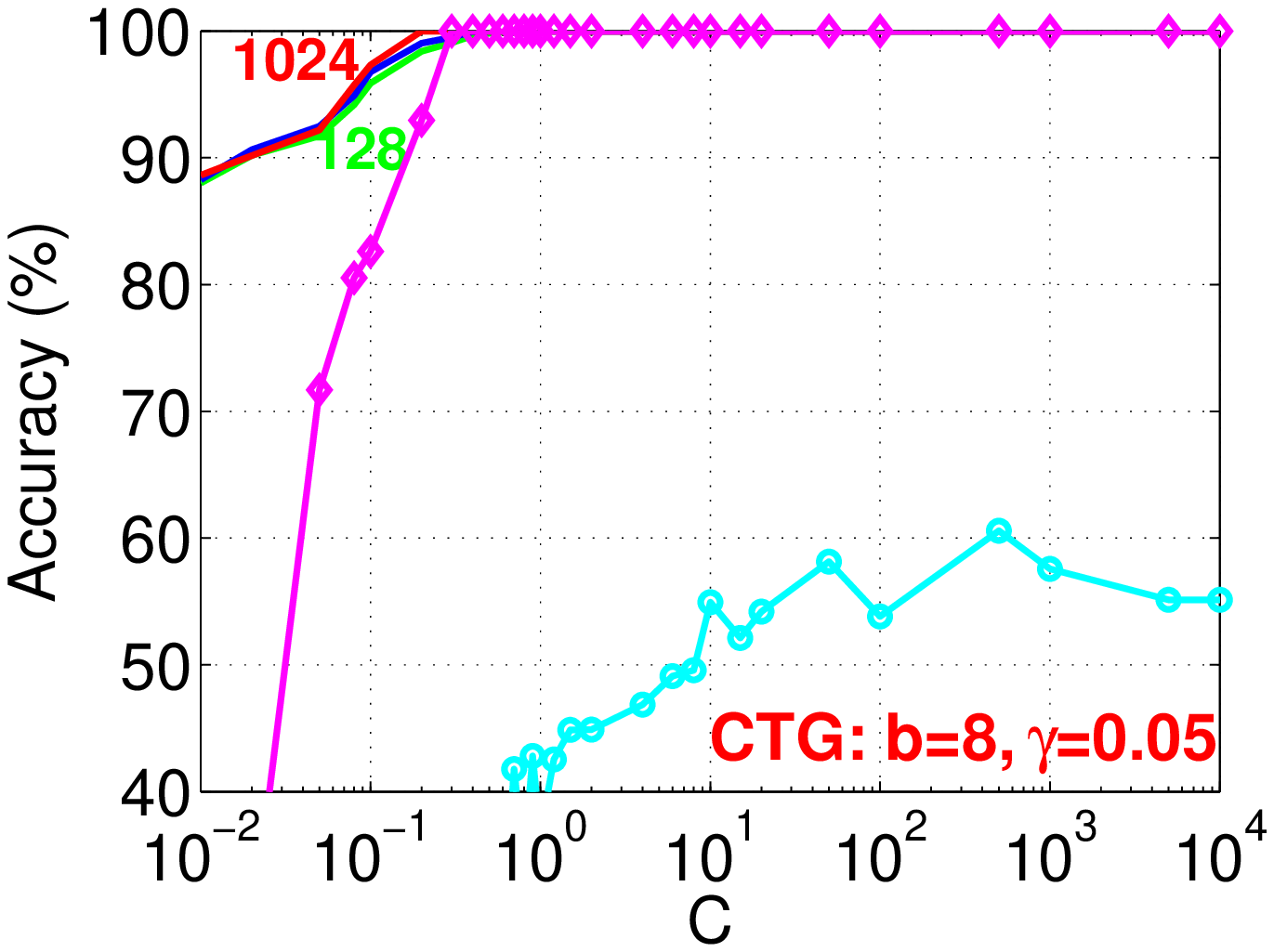}\hspace{-0.14in}
\includegraphics[width=2.3in]{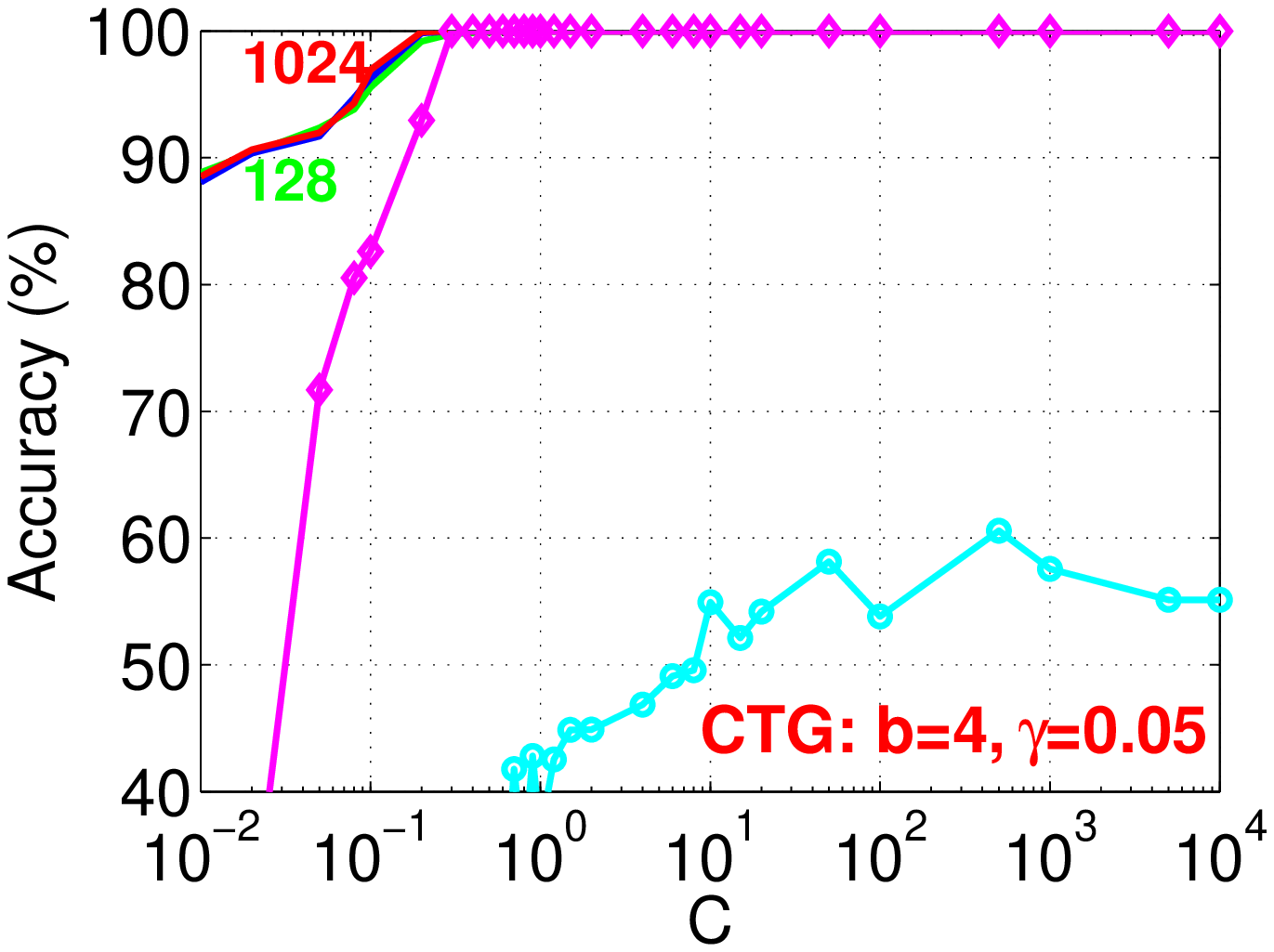}\hspace{-0.14in}
\includegraphics[width=2.3in]{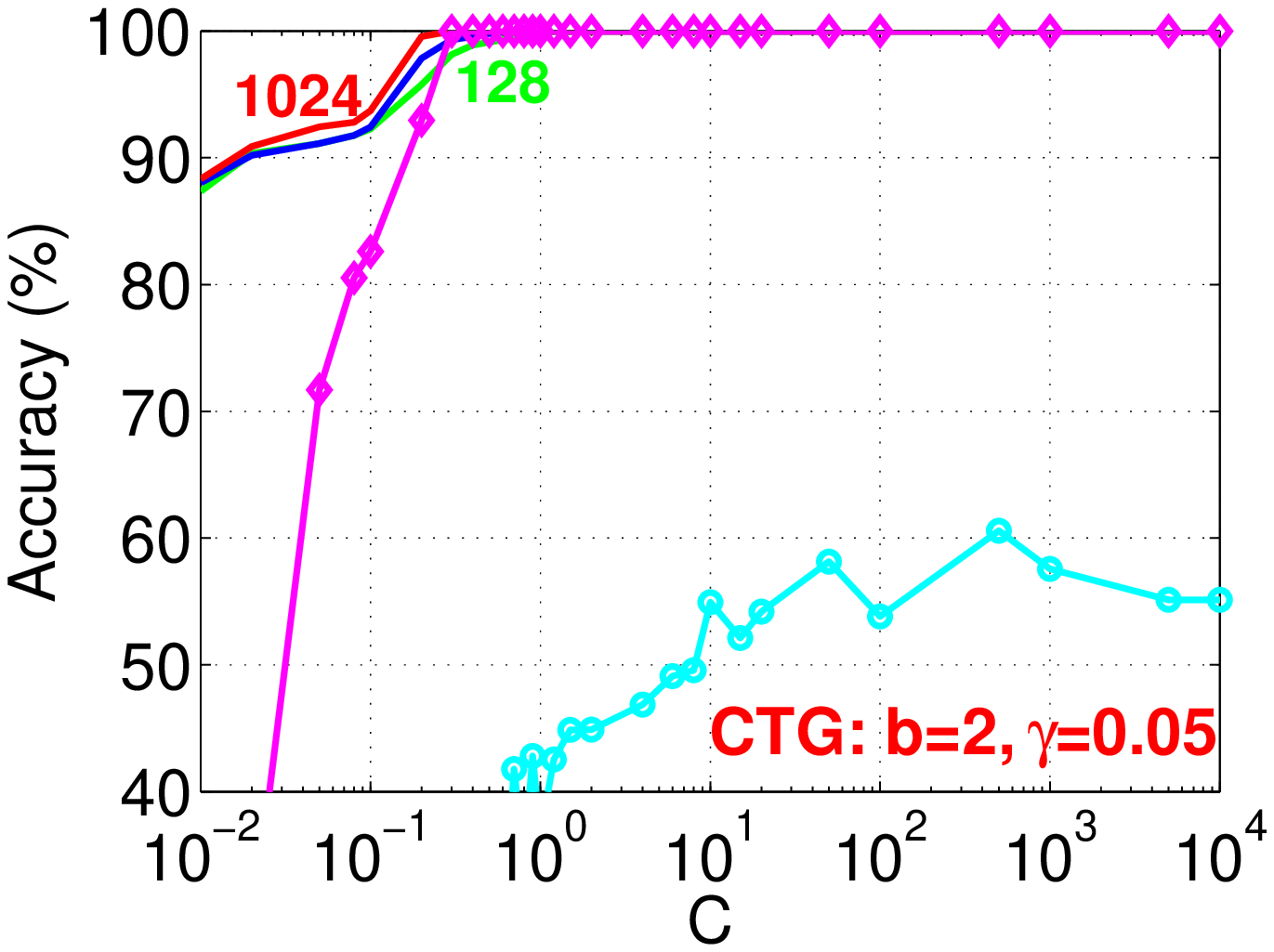}\hspace{-0.14in}
}

\mbox{
\includegraphics[width=2.3in]{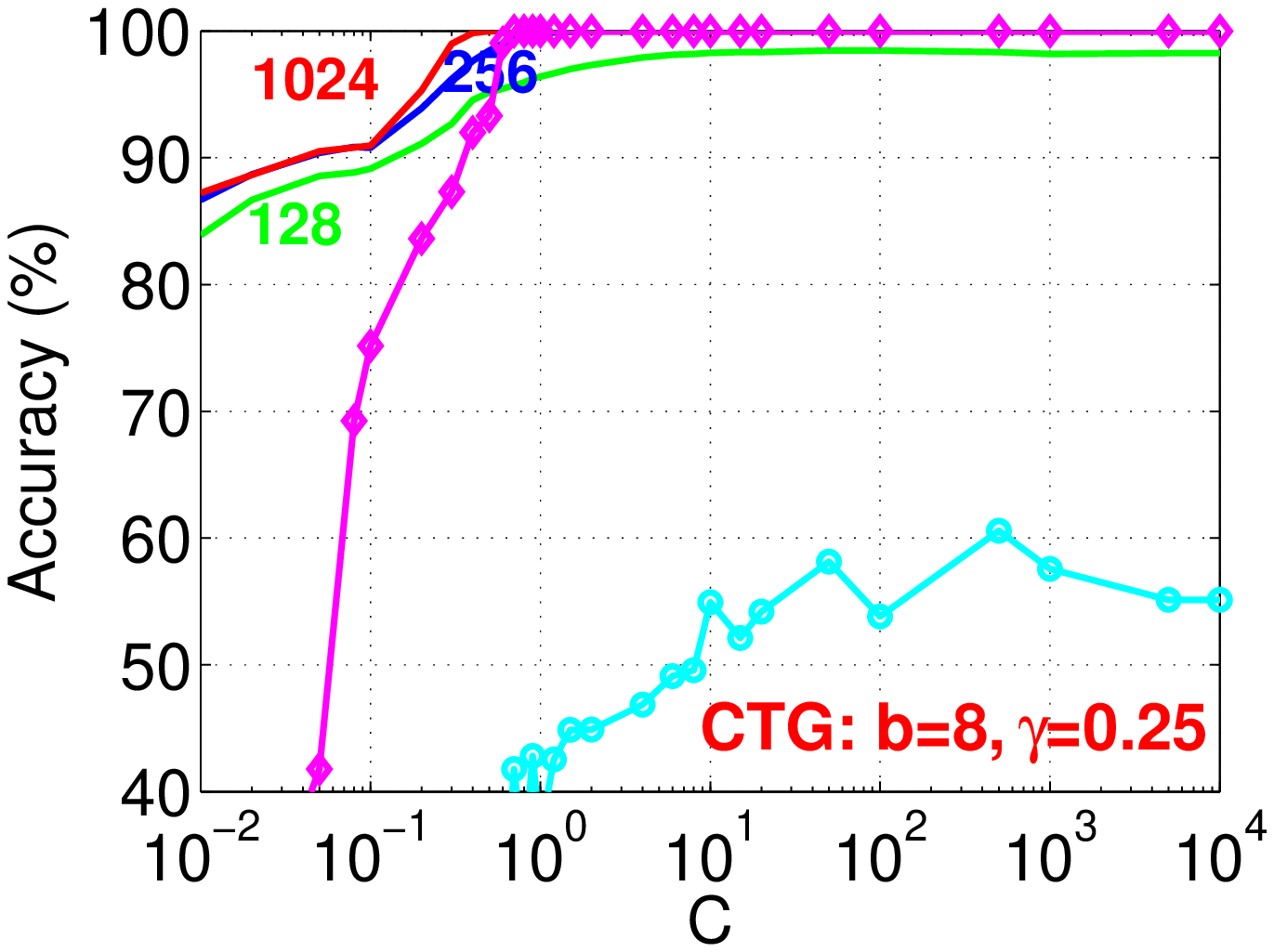}\hspace{-0.14in}
\includegraphics[width=2.3in]{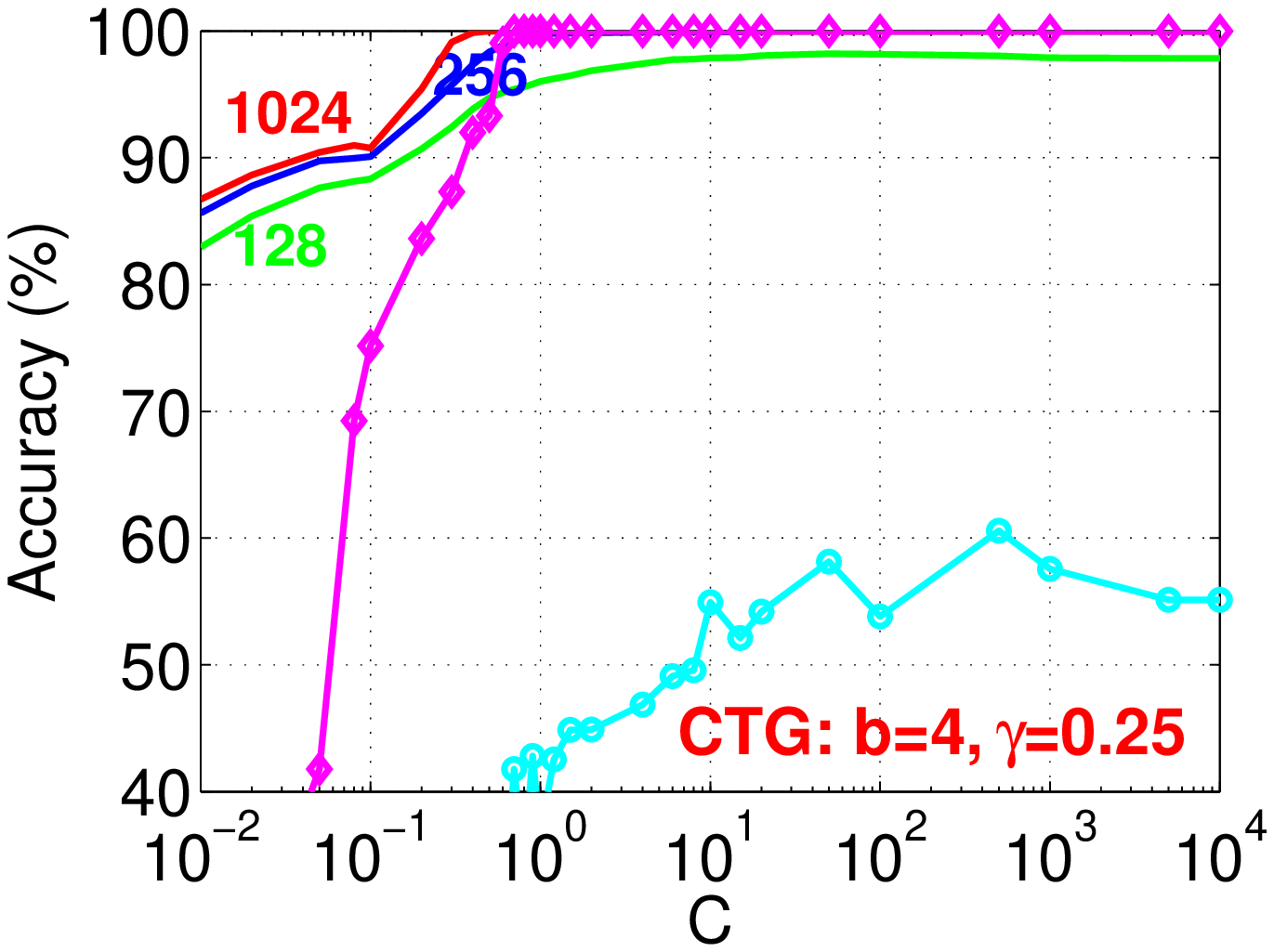}\hspace{-0.14in}
\includegraphics[width=2.3in]{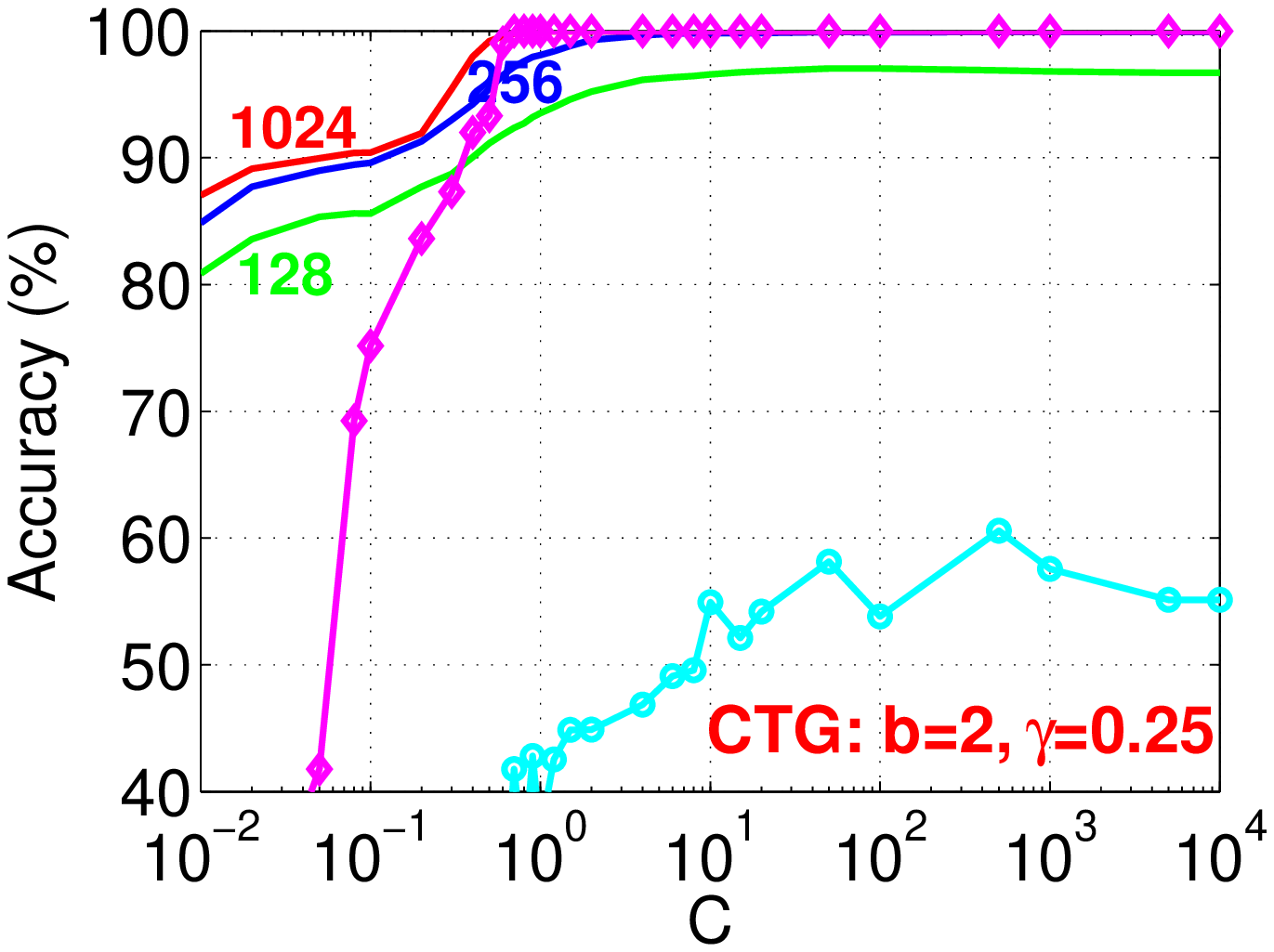}\hspace{-0.14in}
}

\mbox{
\includegraphics[width=2.3in]{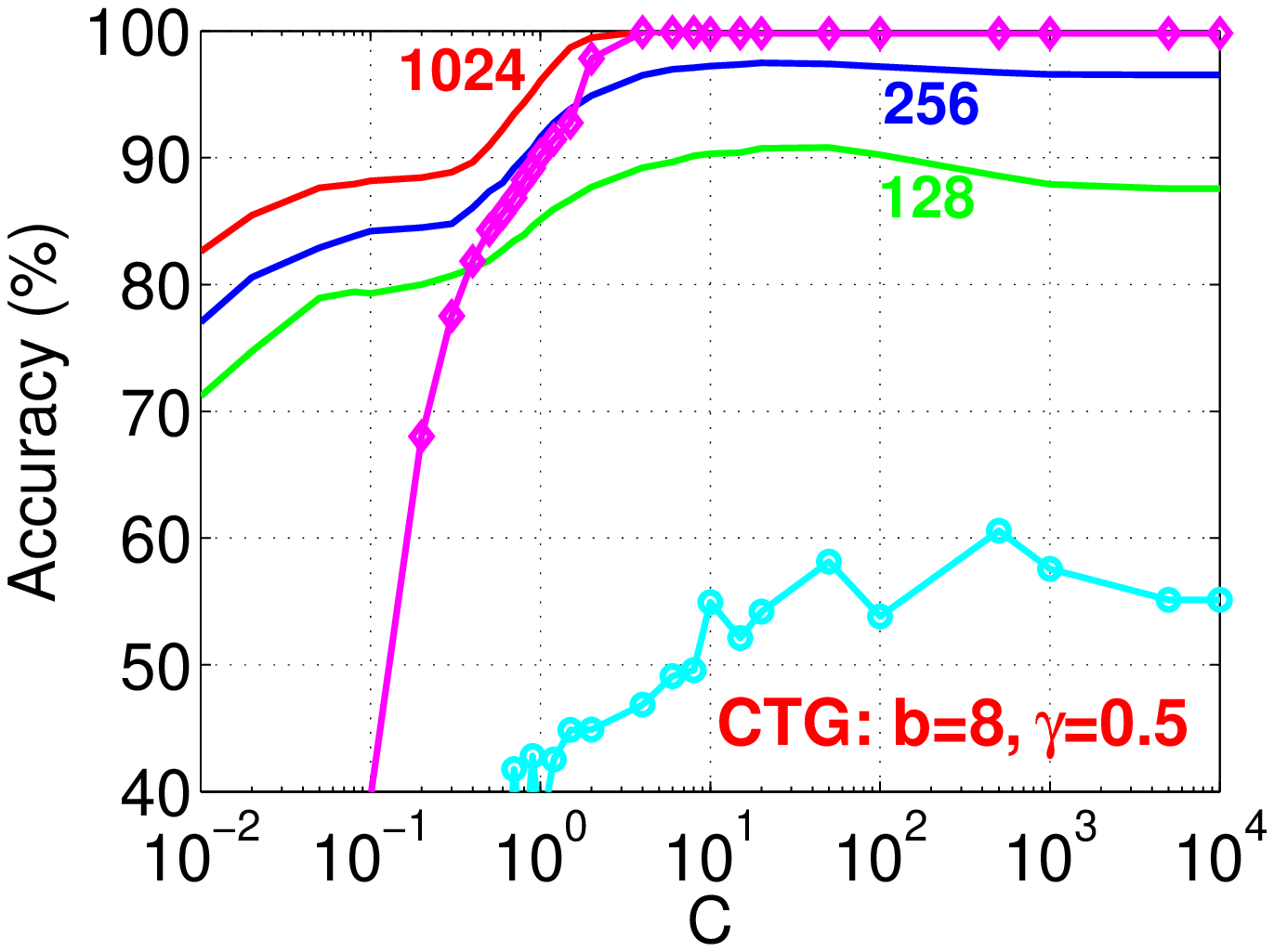}\hspace{-0.14in}
\includegraphics[width=2.3in]{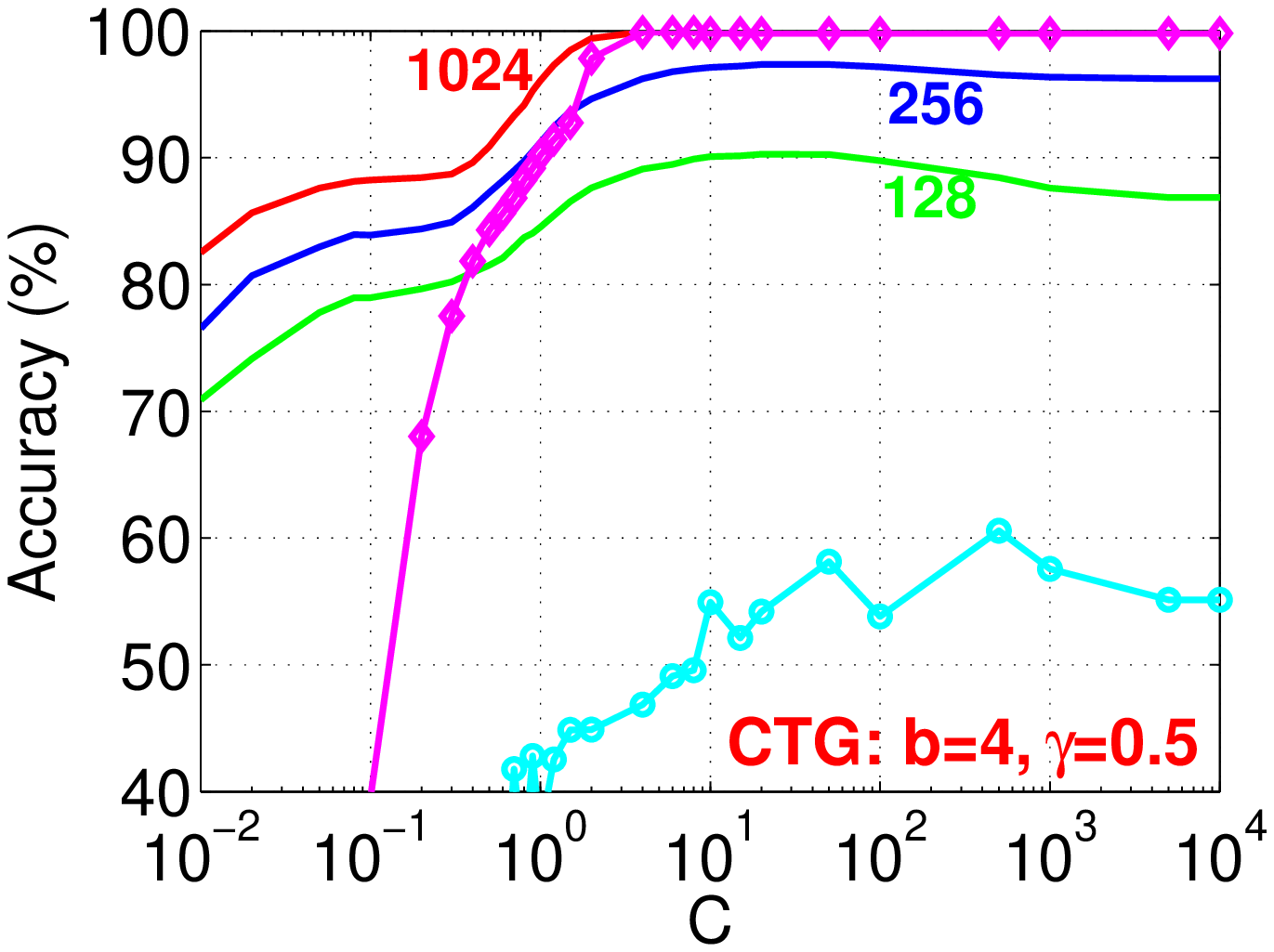}\hspace{-0.14in}
\includegraphics[width=2.3in]{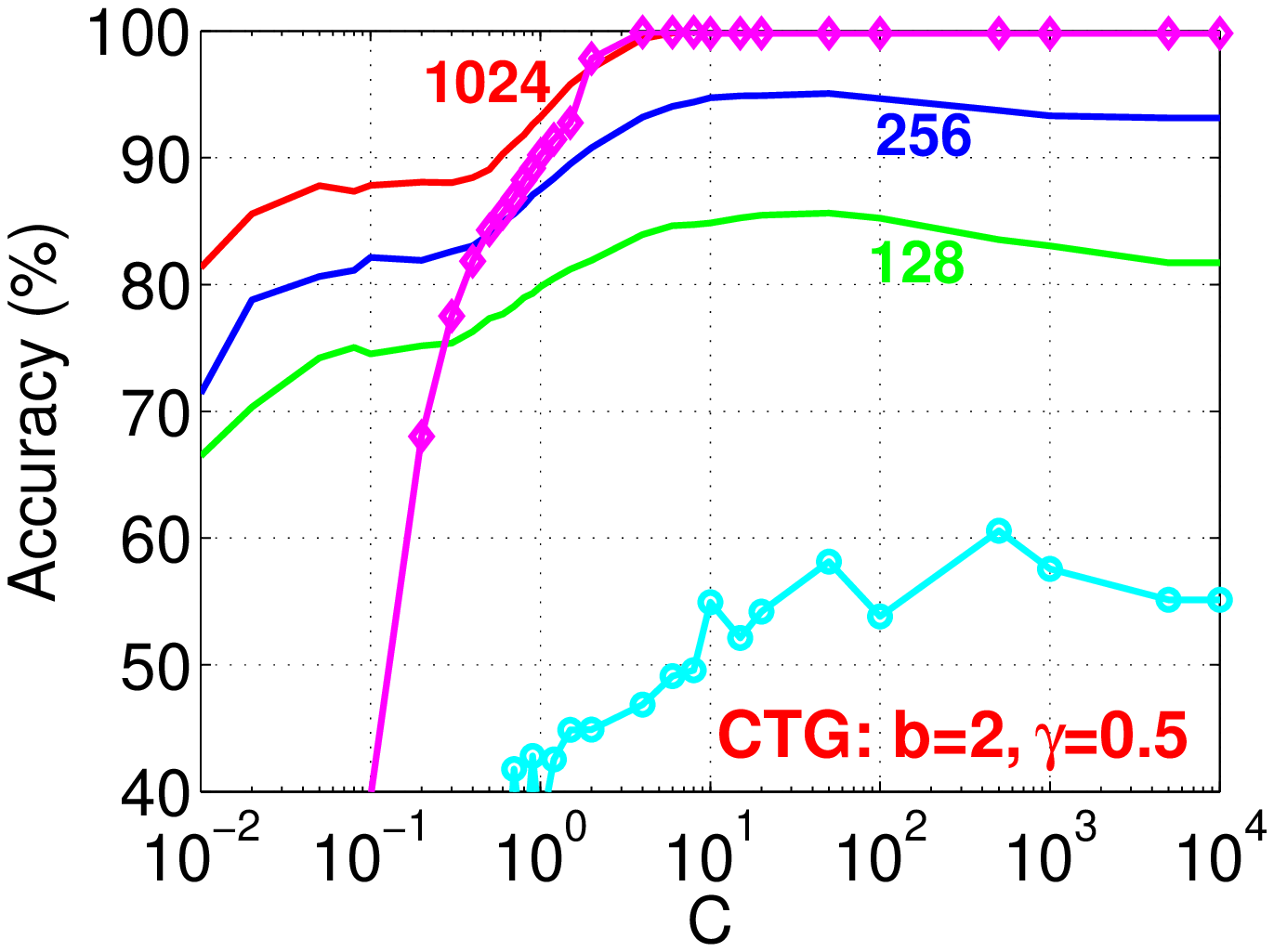}\hspace{-0.14in}
}

\mbox{
\includegraphics[width=2.3in]{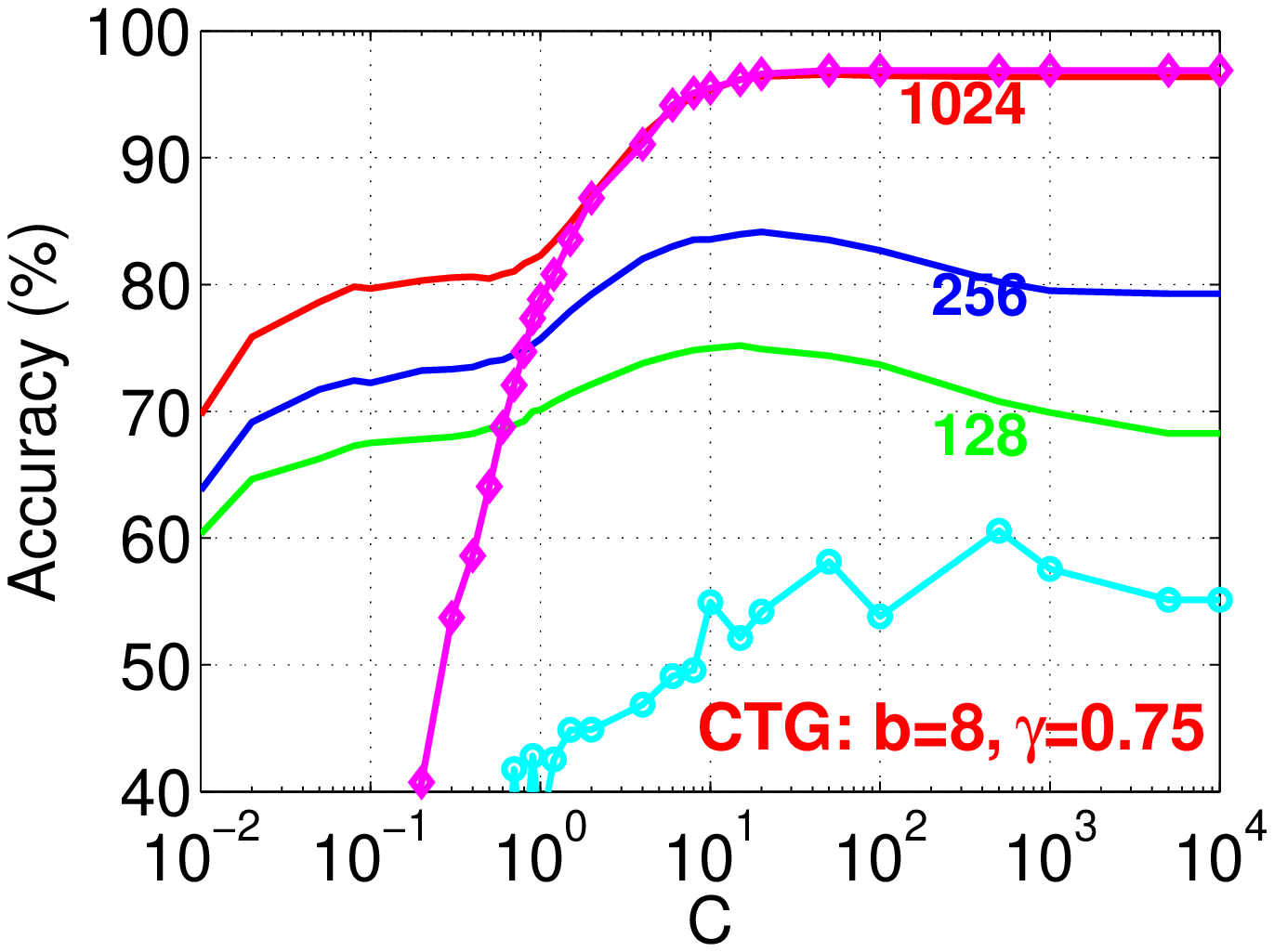}\hspace{-0.14in}
\includegraphics[width=2.3in]{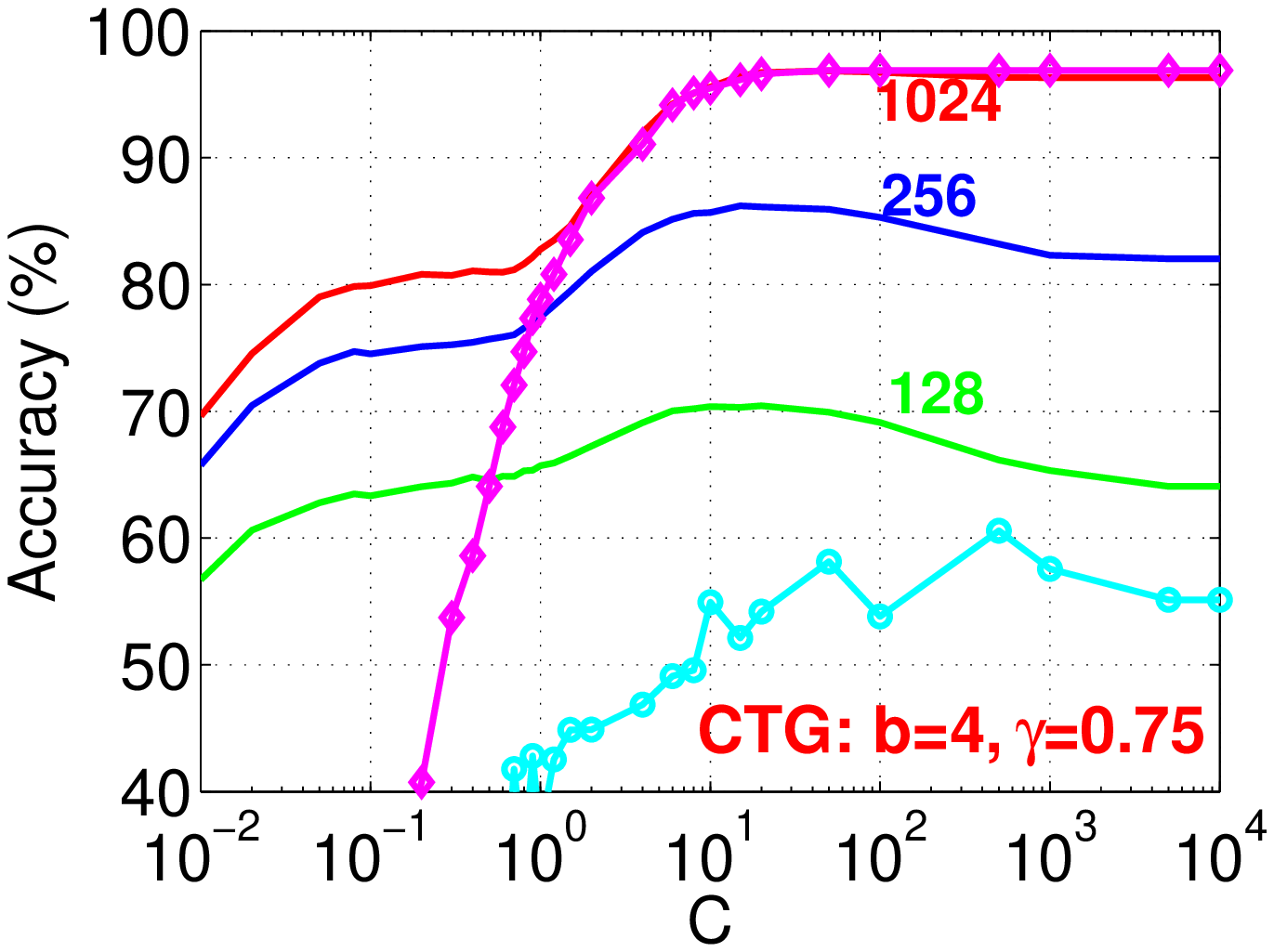}\hspace{-0.14in}
\includegraphics[width=2.3in]{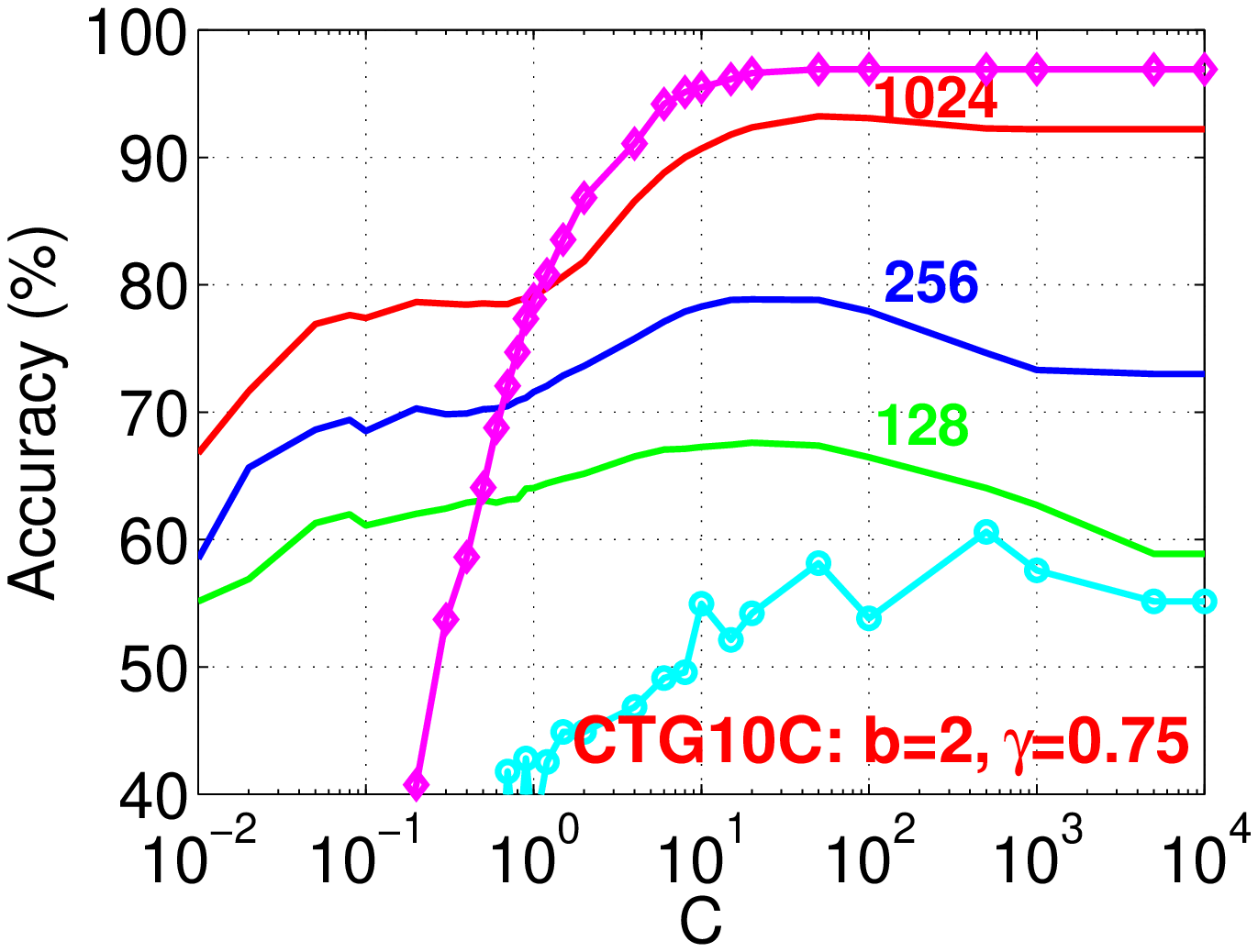}\hspace{-0.14in}
}

\end{center}
\vspace{-0.3in}
\caption{Test classification accuracies for using linear classifiers combined with hashing in Algorithm~\ref{alg_GCWS} on CTG dataset, for $\gamma\in\{0.05, 0.25, 0.5, 0.75\}$ to visualize the trend that, for this dataset, the accuracy decreases with increasing $\gamma$. Three columns presents results for $b=8, 4, 2$, respectively.  }\label{fig_HashCTG10C}
\end{figure}

\begin{figure}[h!]
\begin{center}

\mbox{
\includegraphics[width=2.3in]{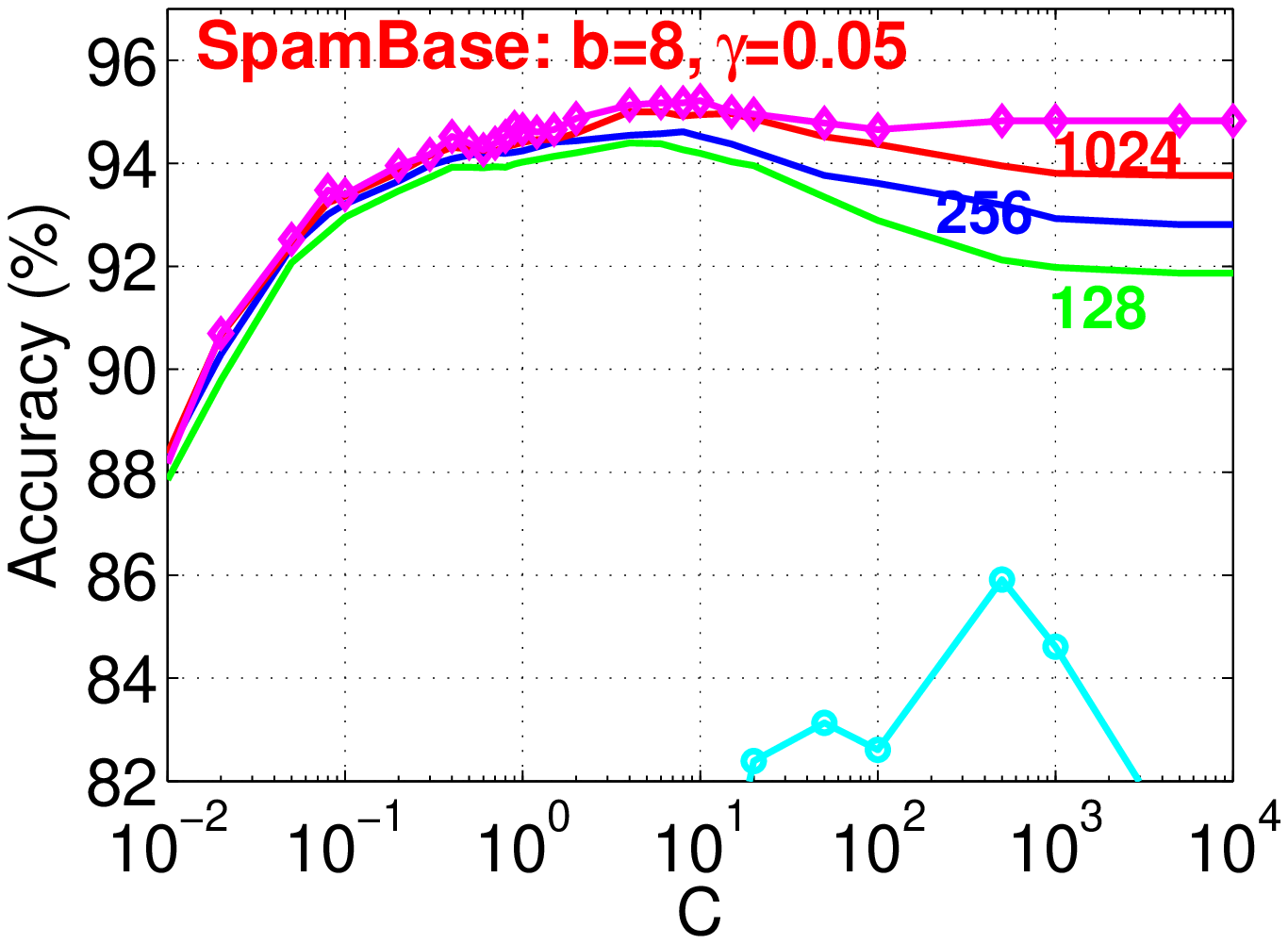}\hspace{-0.14in}
\includegraphics[width=2.3in]{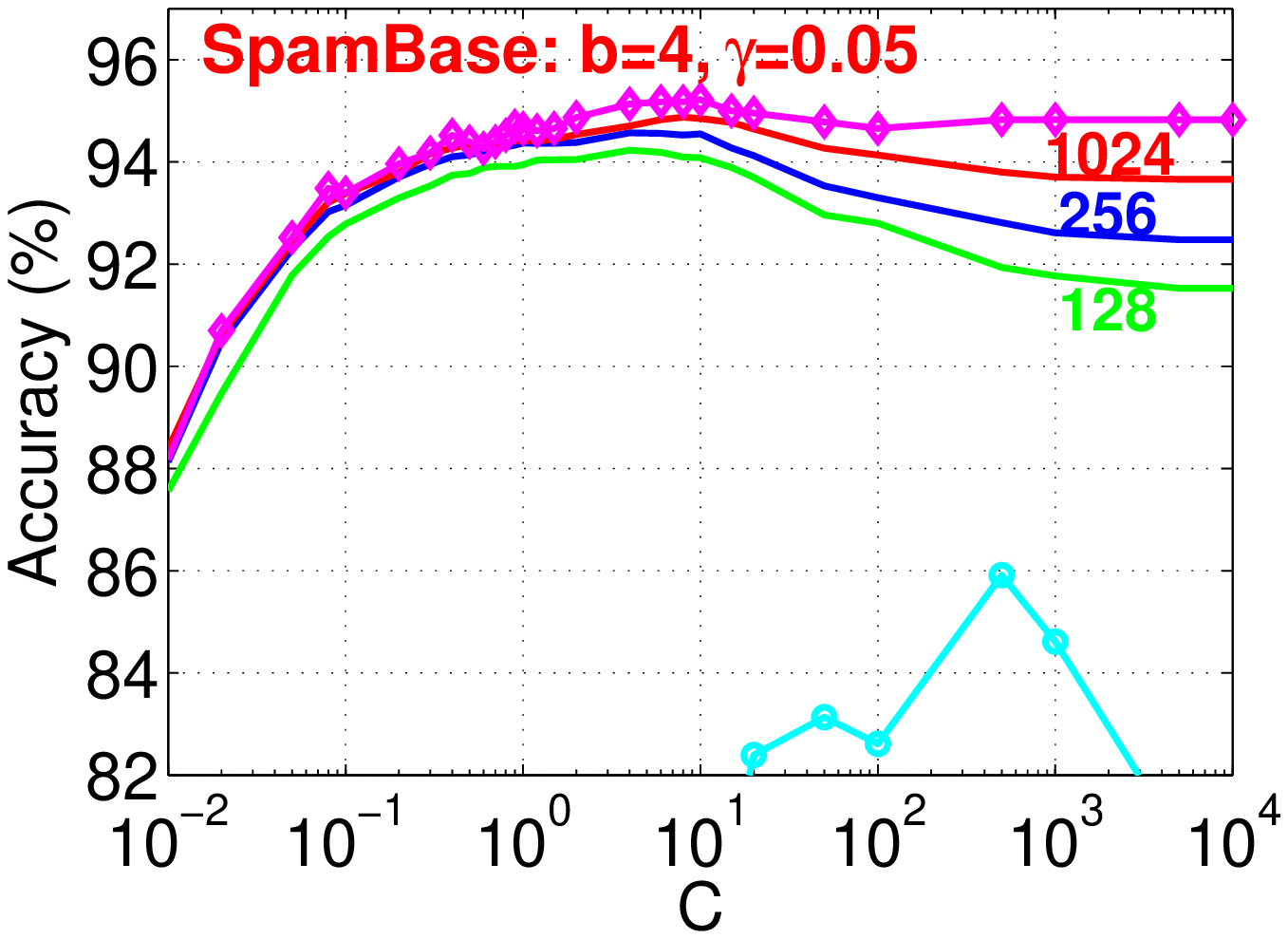}\hspace{-0.14in}
\includegraphics[width=2.3in]{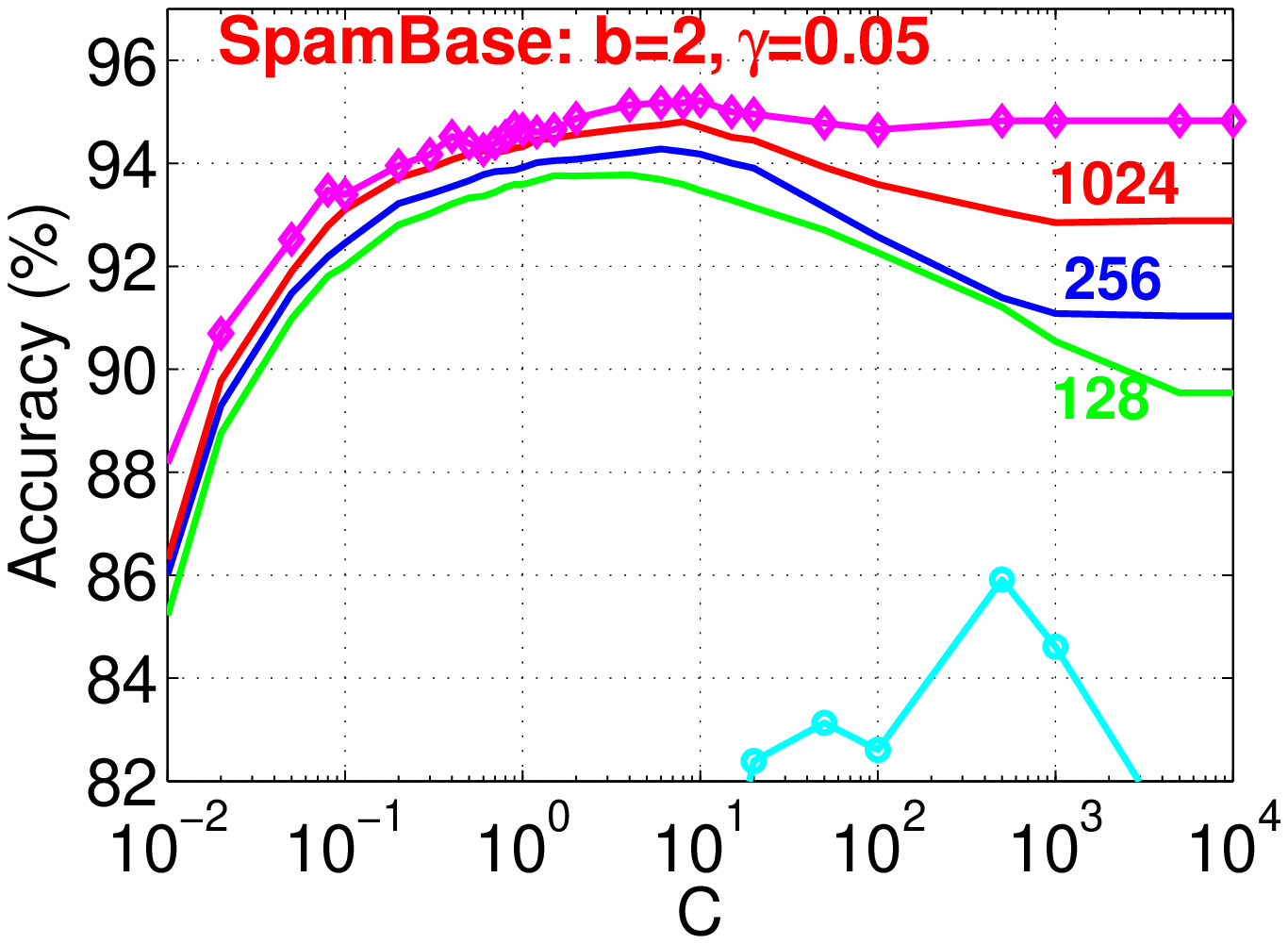}\hspace{-0.14in}
}

\mbox{
\includegraphics[width=2.3in]{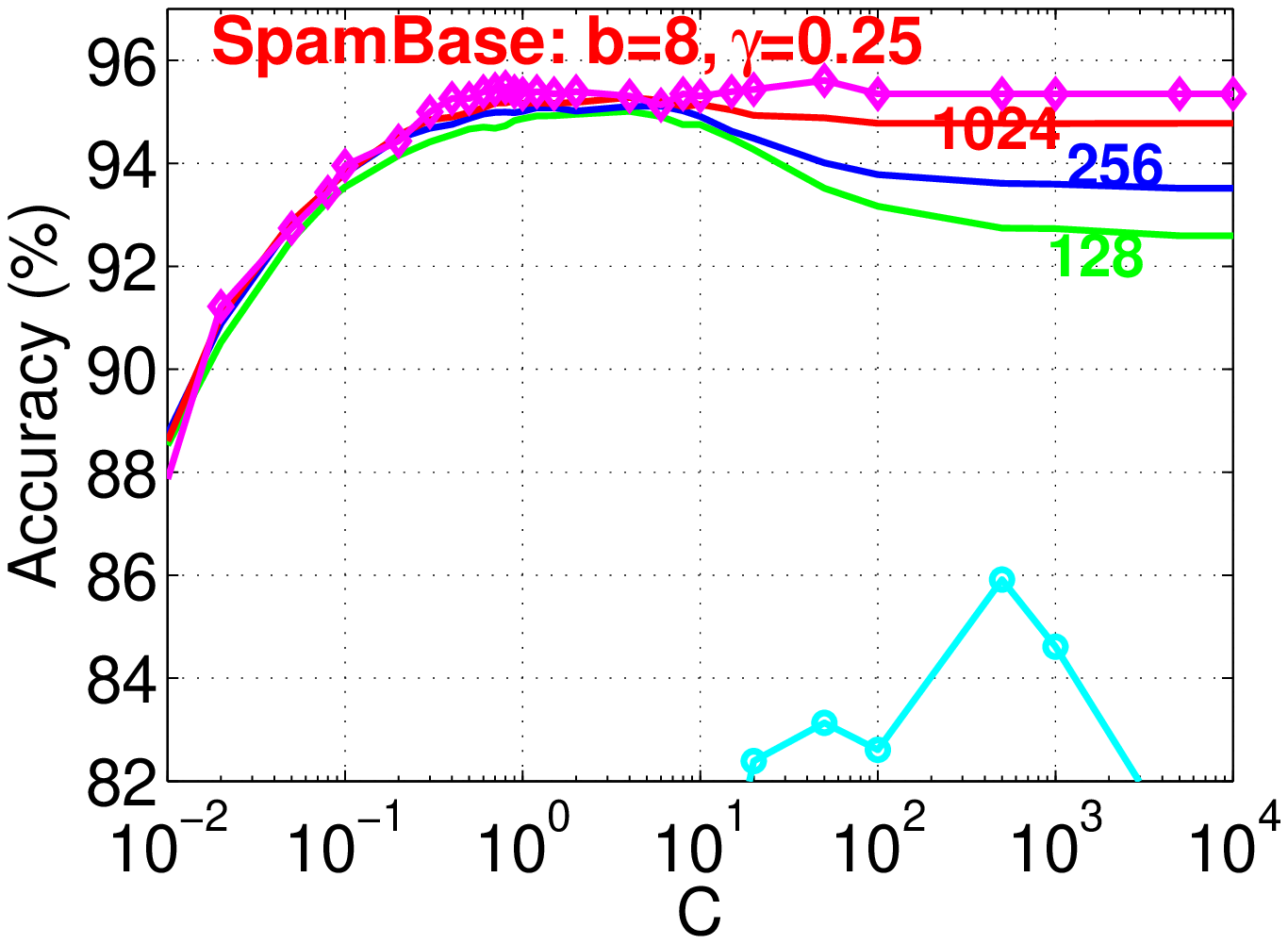}\hspace{-0.14in}
\includegraphics[width=2.3in]{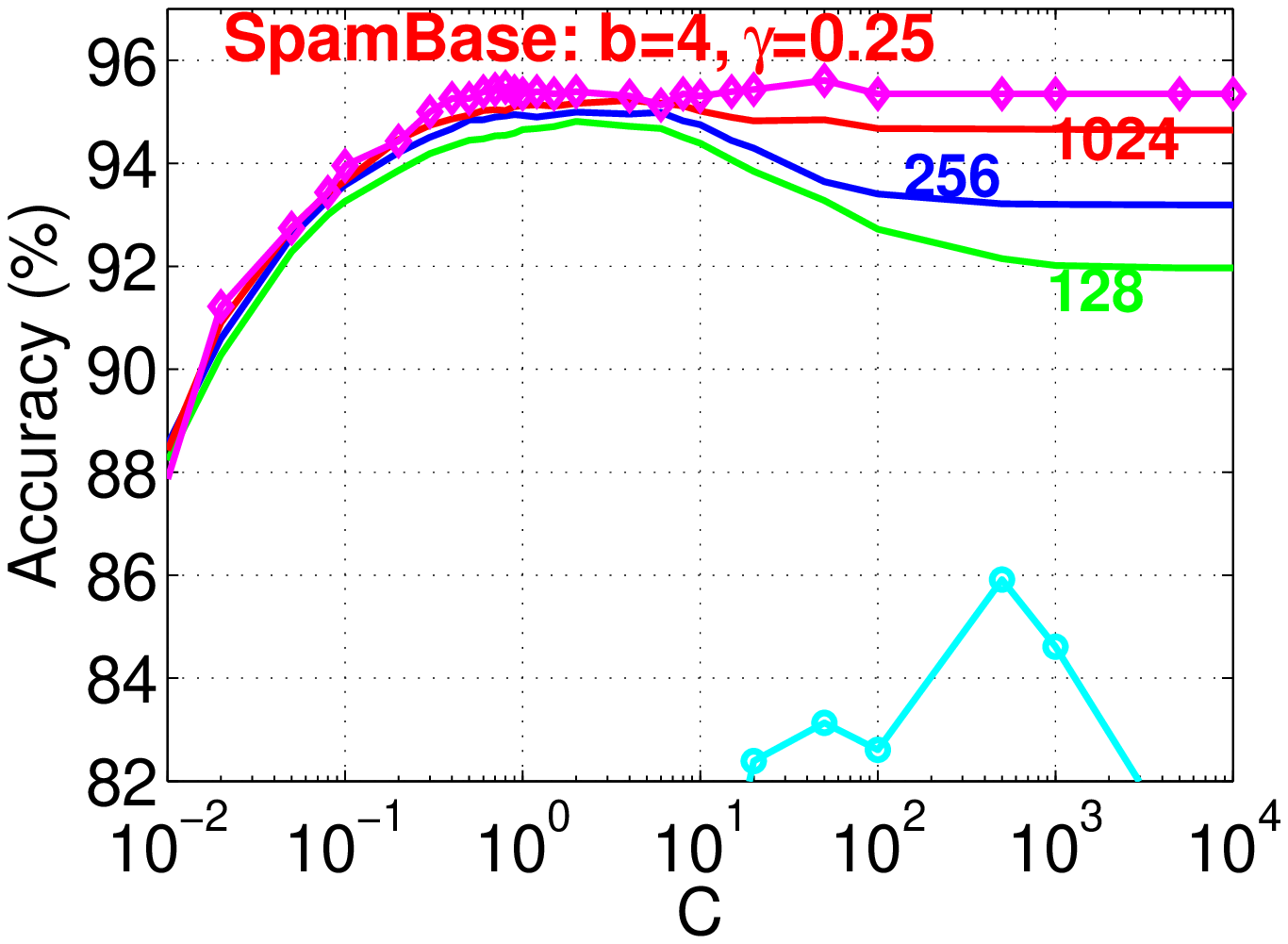}\hspace{-0.14in}
\includegraphics[width=2.3in]{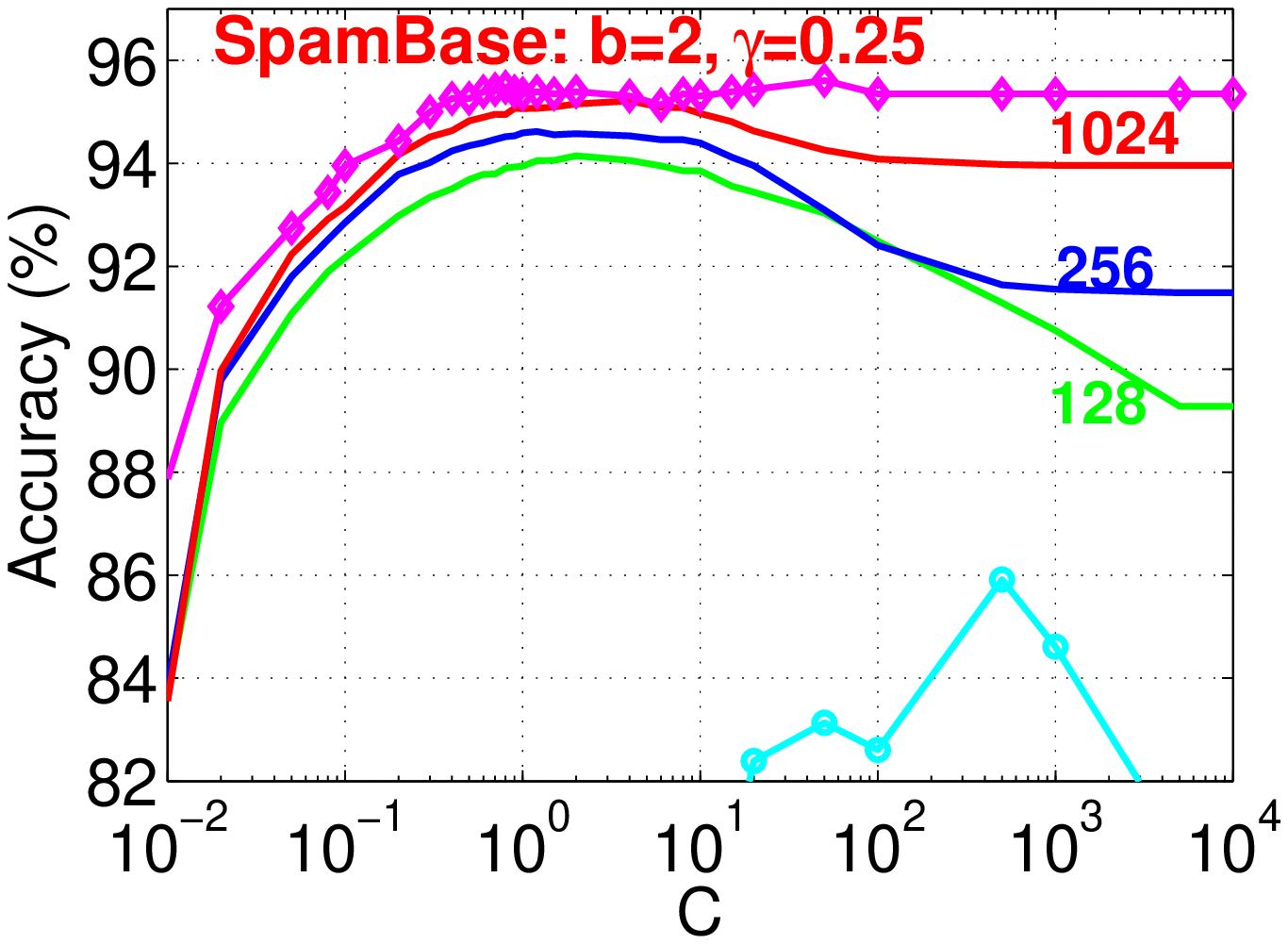}\hspace{-0.14in}
}

\mbox{
\includegraphics[width=2.3in]{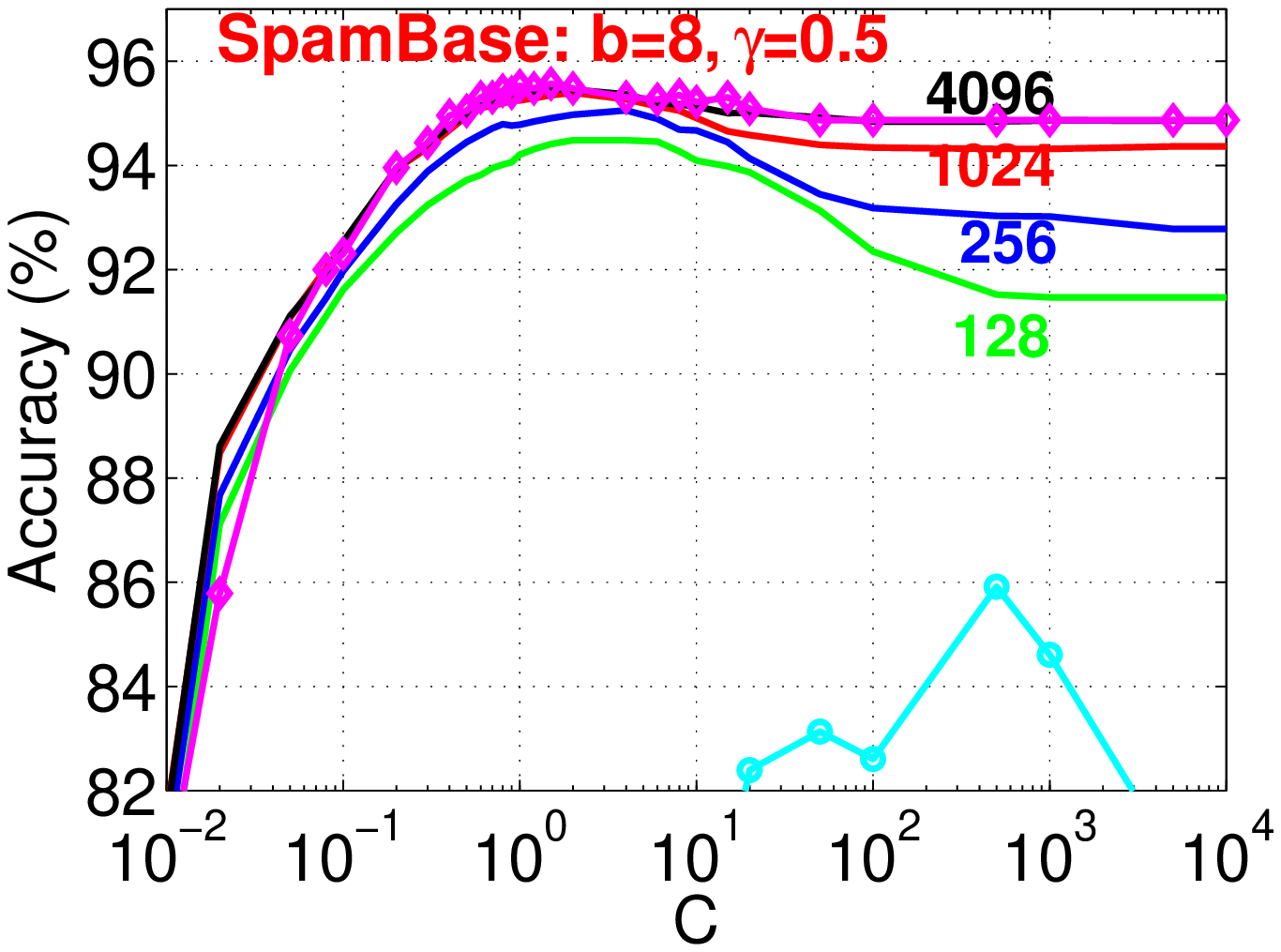}\hspace{-0.14in}
\includegraphics[width=2.3in]{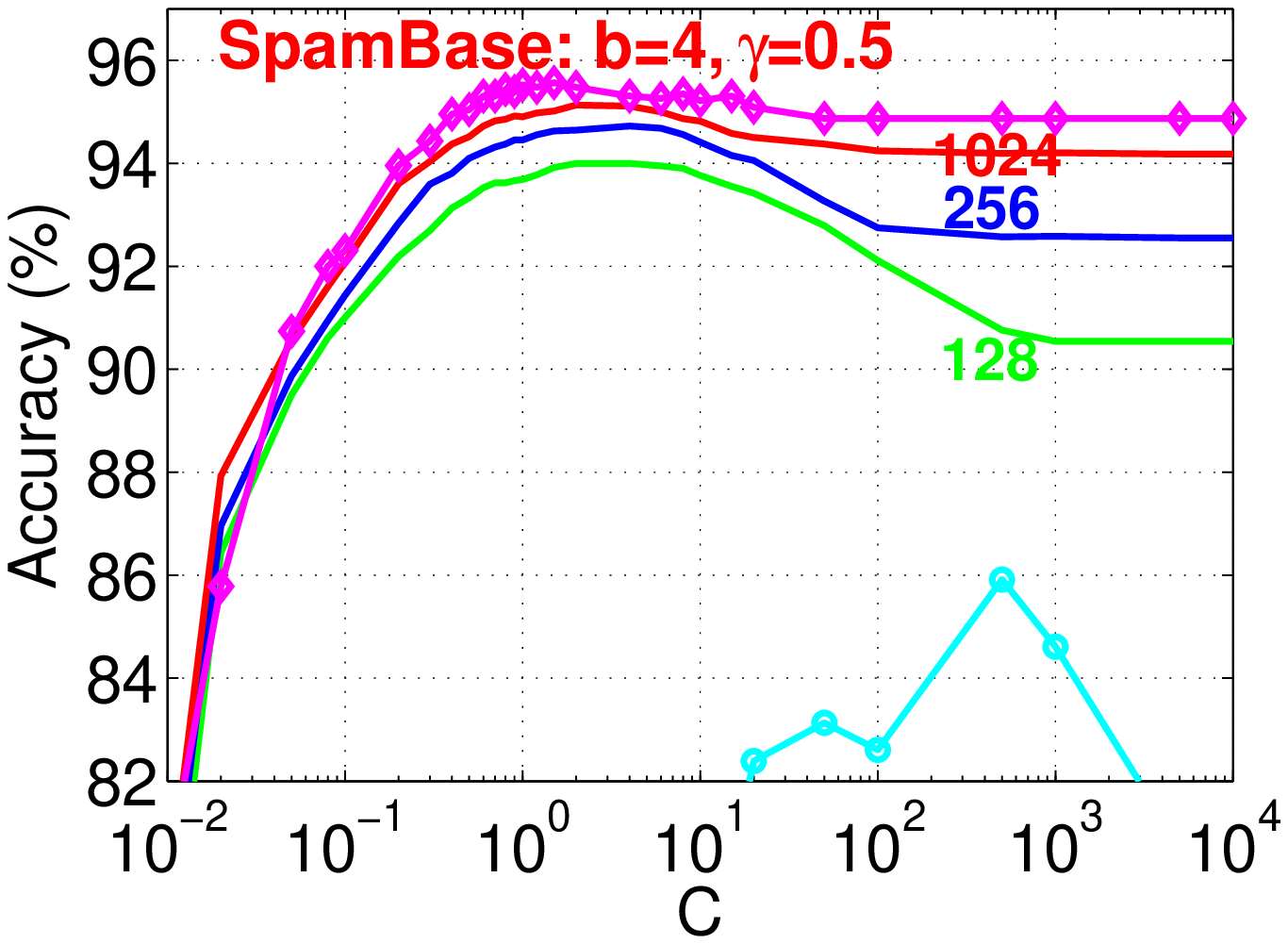}\hspace{-0.14in}
\includegraphics[width=2.3in]{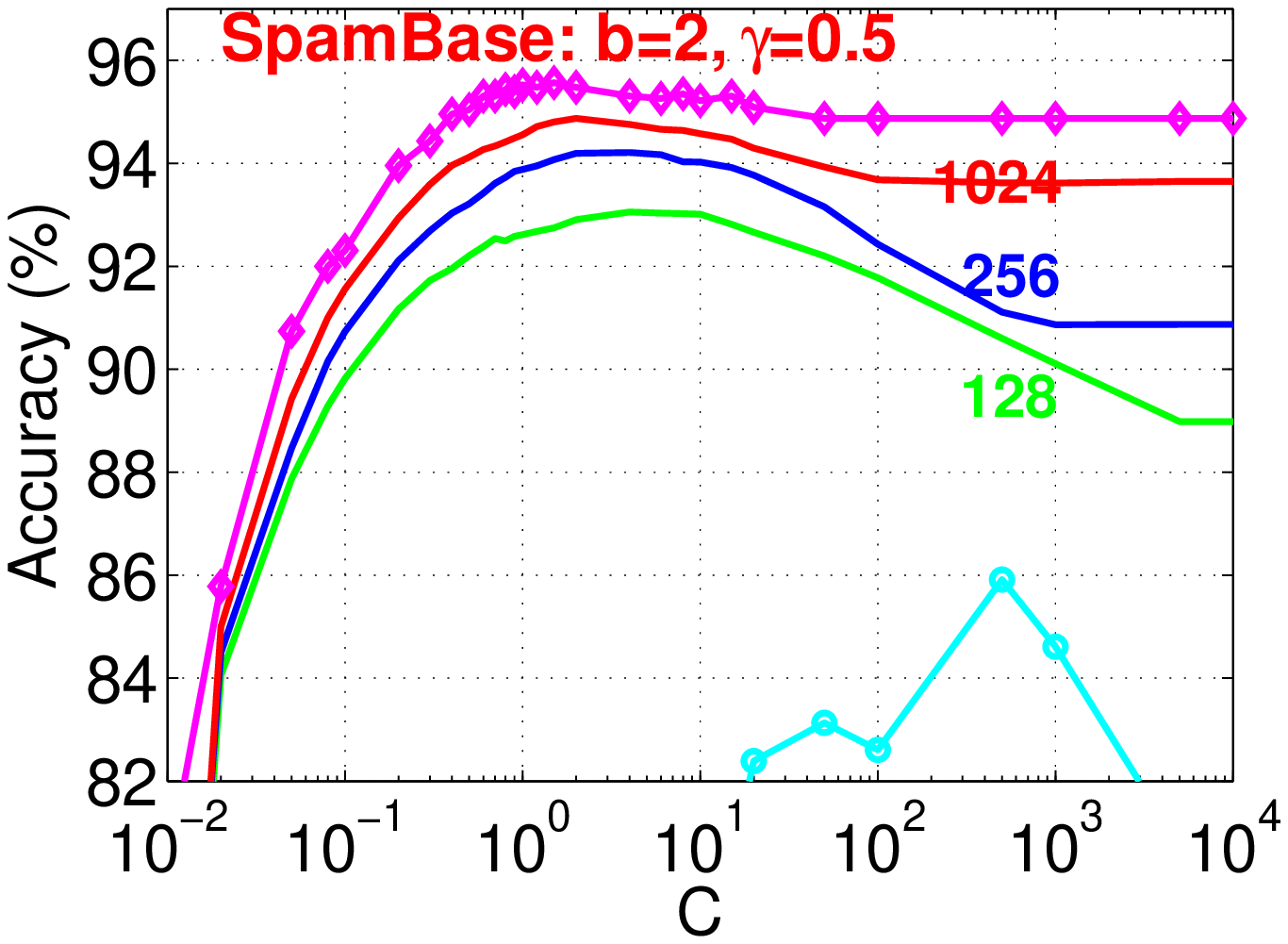}\hspace{-0.14in}
}

\mbox{
\includegraphics[width=2.3in]{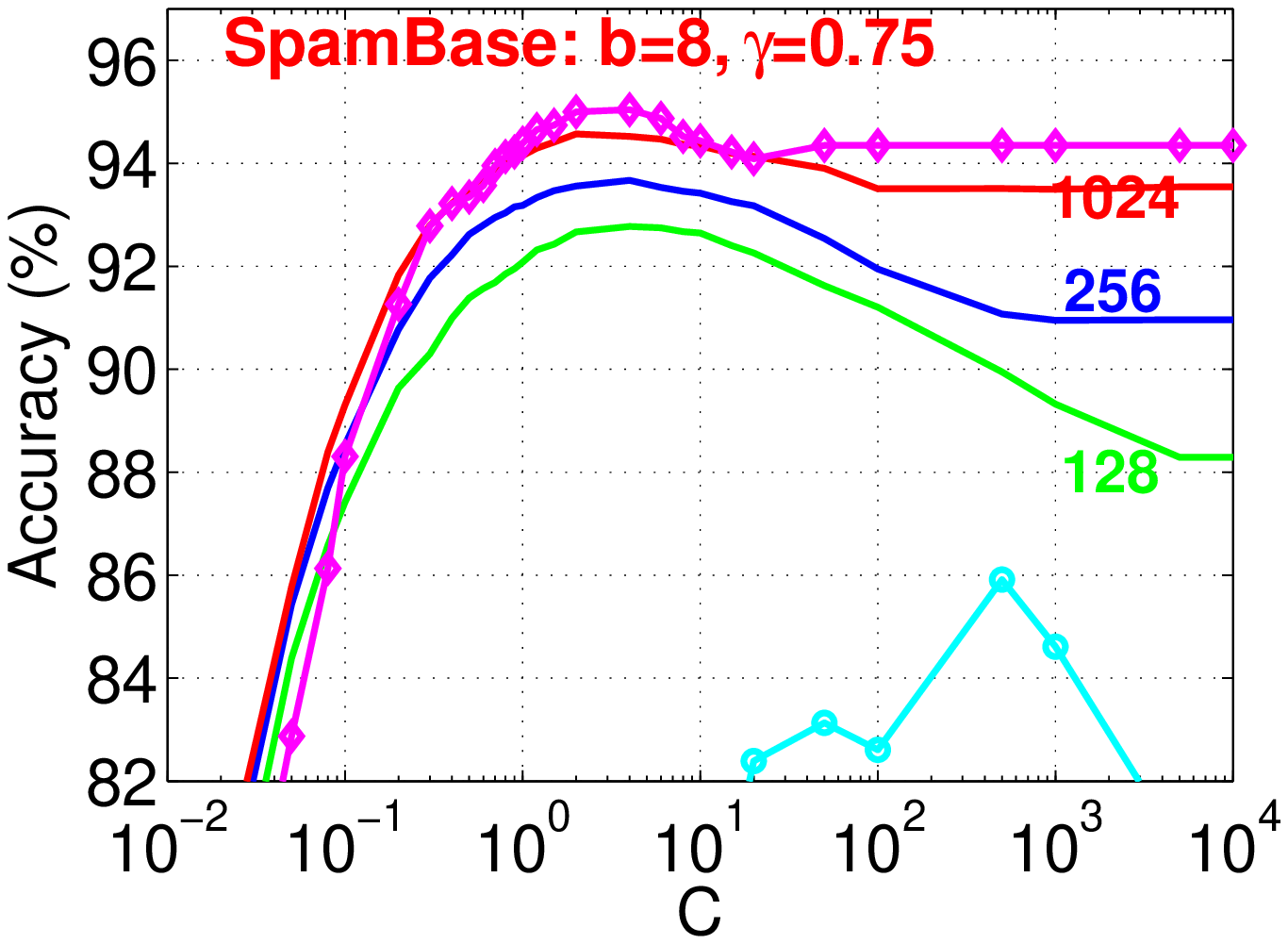}\hspace{-0.14in}
\includegraphics[width=2.3in]{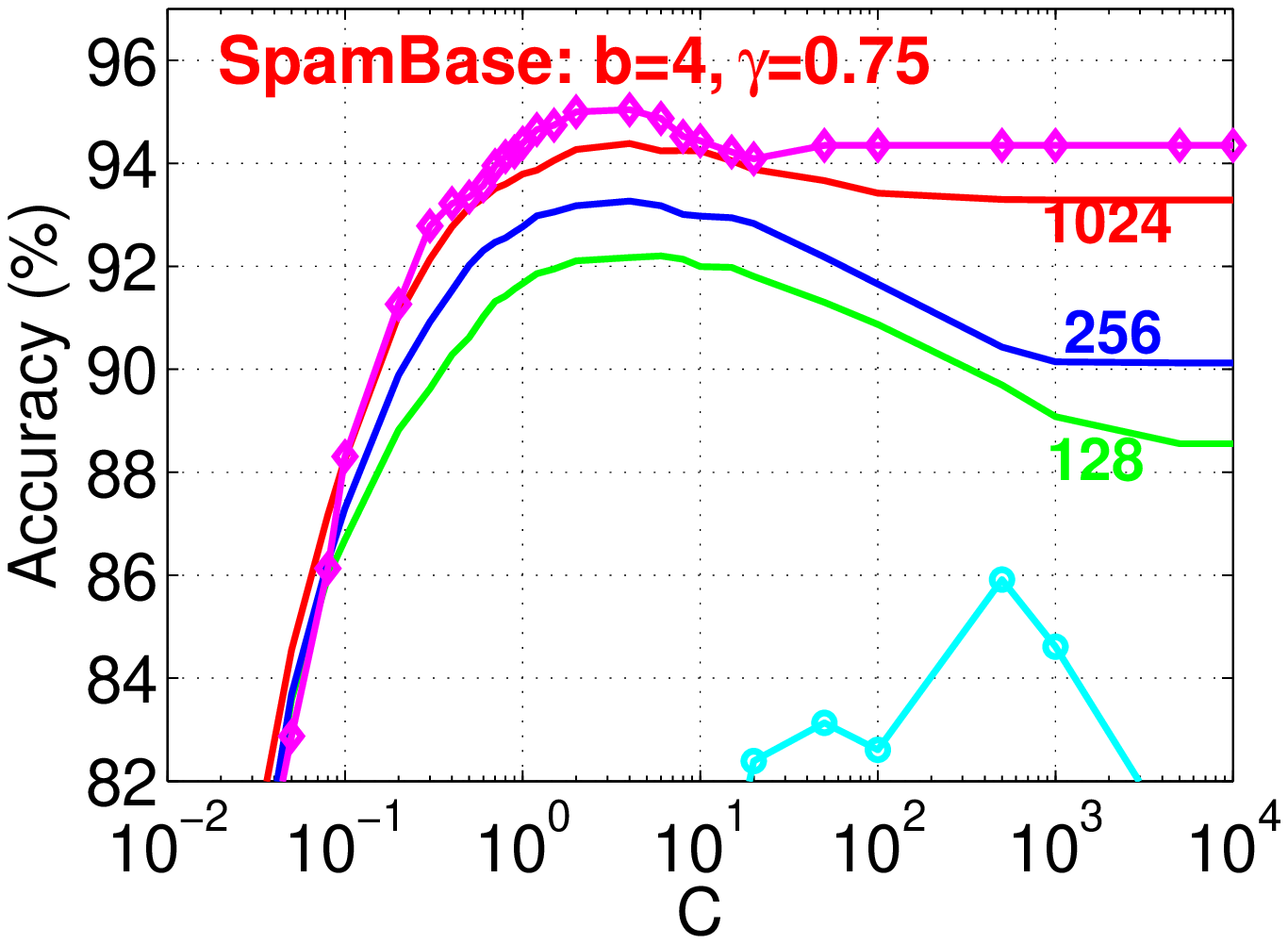}\hspace{-0.14in}
\includegraphics[width=2.3in]{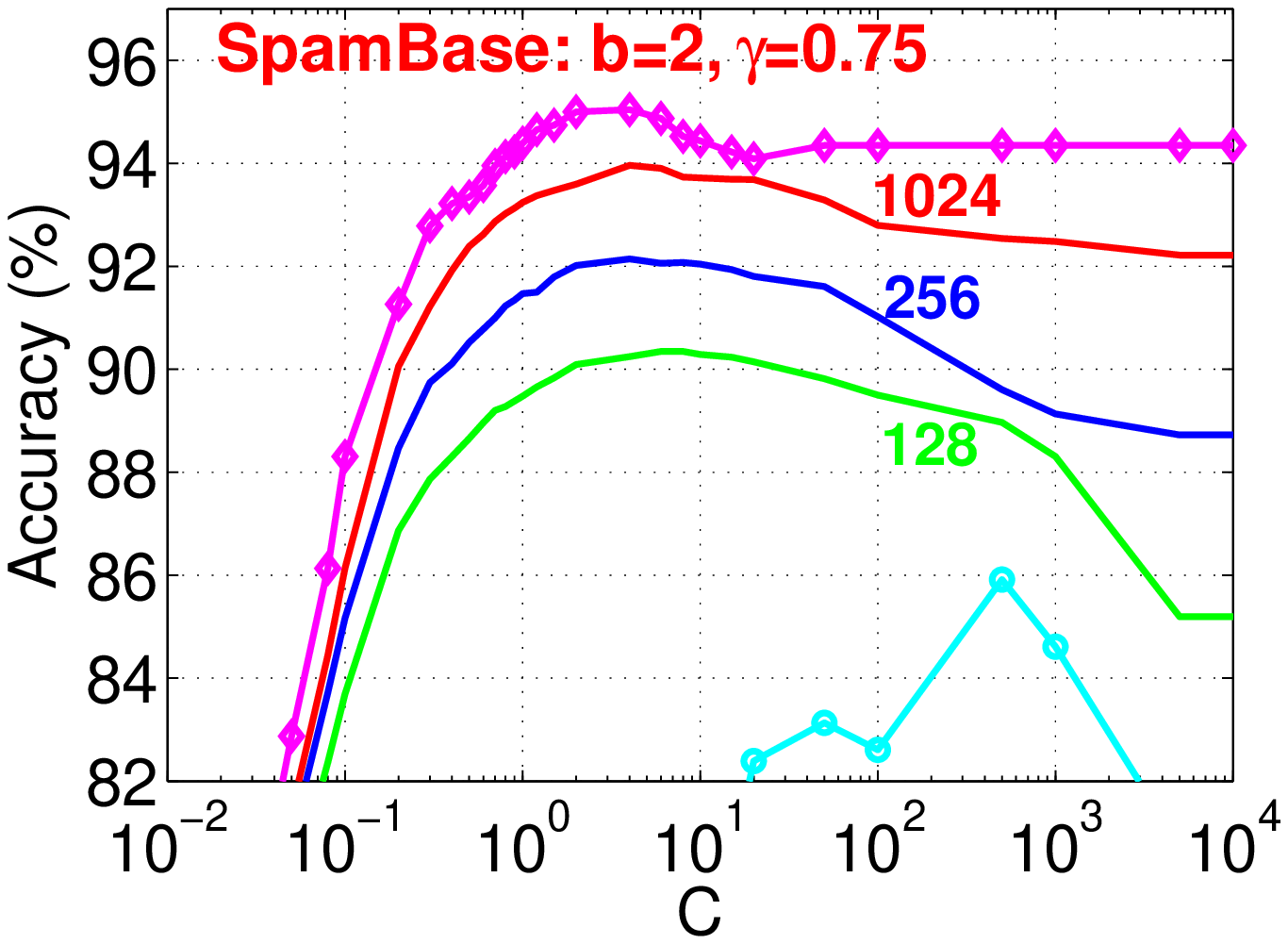}\hspace{-0.14in}
}

\end{center}
\vspace{-0.3in}
\caption{Test classification accuracies for using linear classifiers combined with hashing in Algorithm~\ref{alg_GCWS} on SpamBase dataset, for $\gamma\in\{0.05, 0.25, 0.5, 0.75\}$ (from top to bottom) to visualize the trend that, for this dataset, the accuracy decreases with increasing $\gamma$. Three columns presents results for $b=8, 4, 2$, respectively (from left to right). }\label{fig_HashSpamBase}
\end{figure}

Figure~\ref{fig_HashCTG10C} (for CTG dataset) and Figure~\ref{fig_HashSpamBase} (for SpamBase dataset) are somewhat different from the previous figures. For both datasets, using $\gamma=0.05$ achieves the best accuracy. We plot the results for $\gamma=0.05, 0.25, 0.5, 0.75$, and $b=8, 4, 2$, to visualize the trend.

\section{Conclusion}

It is  commonly believed that deep learning algorithms and tree methods can produce the state-of-the-art results in many statistical machine learning tasks. In 2010, ~\cite{Proc:ABC_UAI10} reported a set of surprising experiments on the datasets used by the deep learning community~\cite{Proc:Larochelle_ICML07}, to show that tree methods can outperform deep nets on a majority (but not all) of those datasets and the improvements can be substantial on a good portion of datasets. \cite{Proc:ABC_UAI10} introduced several ideas including the second-order tree-split formula and the new derivatives for multi-class logistic loss function. Nevertheless, tree methods are slow and their model sizes are typically large.

\vspace{0.1in}

\noindent In machine learning practice with massive data, it is desirable to develop algorithms which run almost as efficient as linear methods (such as linear logistic regression or linear SVM) and achieve similar accuracies as nonlinear methods. In this study, the tunable linearized GMM kernels are promising tools for achieving those goals. Our extensive experiments on the same datasets used for testing tree methods and deep nets demonstrate that tunable GMM kernels and their linearized versions through hashing can achieve comparable accuracies as trees. In general, the state-of-the-art boosted tree method called ``abc-robust-logitboost'' typically achieves better accuracies than the proposed tunable GMM kernels. Also, on some datasets, deep learning methods or RBF kernel SVM outperform tunable GMM kernels. Therefore, there is still  room for future improvements.

\vspace{0.1in}

\noindent In this study, we focus on testing tunable GMM kernels and their linearized versions using classification tasks. It is clear that  these techniques basically generate new data representations and hence can be applied to a wide variety of statistical learning tasks including clustering and regression. Due to the discrete name of the hashed values, the techniques naturally can also be used for building hash tables for fast near neighbor search.

\vspace{0.1in}

\noindent The current version of this paper is mainly a technical note for supporting the recent work on ``The Linearized GMM Kernels and Normalized Random Fourier Features''~\cite{Report:Li_GMM16}.

\newpage\clearpage
{
\bibliographystyle{abbrv}
\bibliography{../bib/mybibfile}

}

\end{document}